\documentclass[10pt]{article}
\usepackage[margin=1in]{geometry}
\usepackage[utf8]{inputenc}
\usepackage[T1]{fontenc}
\usepackage[hidelinks]{hyperref}
\usepackage{url}
\usepackage{booktabs}
\usepackage{amsfonts}
\usepackage{nicefrac}
\usepackage{microtype}
\usepackage{graphicx}
\usepackage{bm}
\usepackage{bbm}
\usepackage{siunitx}
\usepackage{amsmath,amsfonts}
\usepackage[numbers,sort&compress]{natbib}
\usepackage[labelfont=bf]{caption}
\usepackage{cleveref}
\usepackage{makecell}
\usepackage[maxfloats=256]{morefloats}
\DeclareMathOperator{\E}{\mathbb{E}}

\title{
    Net benefit, calibration, threshold selection, and training objectives for algorithmic fairness in healthcare
}

\author{
    Stephen R. Pfohl$^{*,1}$, 
    Yizhe Xu$^{1}$,
    Agata Foryciarz$^{2}$, \\
    Nikolaos Ignatiadis$^{3}$,
    Julian Genkins$^{1}$,
    Nigam H. Shah$^{1}$
}

\date{
    $^1$Stanford Center for Biomedical Informatics Research, Stanford University, Stanford, California 94305, USA \\
    $^2$Department of Computer Science, Stanford University, Stanford, California 94305, USA \\
    $^3$Department of Statistics, Stanford University, Stanford, California 94305, USA \\
    *Correspondence to: \texttt{spfohl@stanford.edu}
}

\begin{document}

\maketitle

\begin{abstract}
    A growing body of work uses the paradigm of algorithmic fairness to frame the development of techniques to anticipate and proactively mitigate the introduction or exacerbation of health inequities that may follow from the use of model-guided decision-making.
    We evaluate the interplay between measures of model performance, fairness, and the expected utility of decision-making to offer practical recommendations for the operationalization of algorithmic fairness principles for the development and evaluation of predictive models in healthcare.
    We conduct an empirical case-study via development of models to estimate the ten-year risk of atherosclerotic cardiovascular disease to inform statin initiation in accordance with clinical practice guidelines.
    We demonstrate that approaches that incorporate fairness considerations into the model training objective typically do not improve model performance or confer greater net benefit for any of the studied patient populations compared to the use of standard learning paradigms followed by threshold selection concordant with patient preferences, evidence of intervention effectiveness, and model calibration.
    These results hold when the measured outcomes are not subject to differential measurement error across patient populations and threshold selection is unconstrained, regardless of whether differences in model performance metrics, such as in true and false positive error rates, are present.
    In closing, we argue for focusing model development efforts on developing calibrated models that predict outcomes well for all patient populations while emphasizing that such efforts are complementary to transparent reporting, participatory design, and reasoning about the impact of model-informed interventions in context. 
\end{abstract}

\section{Introduction}
The use of machine learning to guide clinical decision-making and resource allocation can introduce or perpetuate inequities in care access and quality, ultimately contributing to health disparities \cite{obermeyer2019dissecting,Vyas2020}. 
Aiming to detect and mitigate such harms, recent works leverage the \textit{algorithmic fairness} paradigm \cite{barocas-hardt-narayanan} to define evaluation criteria and model development procedures that quantify and constrain the magnitude of statistical differences in model behavior or performance across patient subgroups \cite{Rajkomar2018,chen2019can,coley2021racial,Seyyed-Kalantari2020,park2021comparison,barda2021addressing,Pfohl2019AIES,zink2020fair,banerjee2021reading,seyyed2021underdiagnosis}.
Within this paradigm, numerous criteria, metrics, and algorithms have been proposed, and both major and minor incompatibilities and trade-offs among them have been identified \cite{Pfohl2021a,foryciarz2021ascvd,Kleinberg2016,Chouldechova2017,Corbett-Davies:2017:ADM:3097983.3098095,Liu2019}.

The purpose of this work is to synthesize, contextualize, and validate underappreciated limitations of the algorithmic fairness paradigm to contribute to the development of best practices for appropriately operationalizing algorithmic fairness principles in healthcare \cite{Gichoya2021operationalize}.
We do so in a setting where observational data stored in an electronic health records or claims database is used to fit a patient-level predictive model for a clinical outcome where the score output by the model informs the allocation of a clinical intervention, typically through comparison of the score to a decision threshold.
For our analysis, we assume that the observed outcomes are not subject to unobserved differential measurement error across patient subgroups \cite{obermeyer2019dissecting,jacobs2021measurement}, that the choice of decision threshold used to allocate a clinical intervention on the basis of the output of a predictive model is not constrained by resource or operational constraints \cite{jung2021framework}, and that the values embedded in the data collection and problem formulation processes are transparently reported and reflect those of the patient populations affected by the use of the model \cite{chen2020ethical,benjamin2019assessing,Passi2019,sendak2020presenting,gebru2018datasheets,Mitchell2019}.

For the development of predictive models to inform clinical decision-making, we argue for aiming to maximize the expected utility that the model-informed intervention confers to each patient subgroup of interest.
The notion of expected utility that we consider depends on the values and preferences of affected stakeholders and can be quantified in terms of the expected costs or utilities associated with false positive and false negative errors in binary classification settings or in terms of the expected benefits and harms of the intervention conditioned on risk in more general settings \cite{Sox2013,Vickers2016}.
We hypothesize that, in practice, model development strategies that nominally promote fairness, by constraining for parity in model performance metrics across subgroups or by maximizing worst-case model performance over subgroups, do not confer greater expected utility for any patient subgroup than the approach of identifying a set of calibrated models that predict the outcome well for each subgroup, followed by threshold selection reflecting the contextual assessment of the benefits and harms of the intervention.
The key observations motivating this hypothesis are detailed in section \ref{sec:background} and largely follow directly from related work \cite{Corbett-Davies:2017:ADM:3097983.3098095,bakalar2021fairness,foryciarz2021ascvd,Liu2019,Kleinberg2016,simoiu2017problem,Sox2013,Wynants2019,vickers2006decision,vickers2007method,Vickers2016,Pfohl2021a,pfohl_dro}.

We evaluate our hypothesis through a case study of estimators of the risk of atherosclerotic cardiovascular disease (ASCVD) within ten years to inform the initiation of cholesterol-lowering statin therapy \cite{goff20142013,Stone2014,Grundy2019,arnett2019,Lloyd-Jones2019}.
We conduct experiments to assess which model development strategies confer maximal expected utility for subgroups defined in terms of race, ethnicity, sex, or co-morbidities (type 1 and type 2 diabetes, chronic kidney disease (CKD), or rheumatoid arthritis (RA)). 
We compare pooled and stratified unconstrained empirical risk minimization (ERM) to regularized fairness objectives and distributionally robust optimization (DRO) objectives that aim to minimize differences in or improve the worst-case area under the receiver operating characteristic curve (AUC) or log-loss across subgroups.
We further conduct an analysis to investigate the impact of constraints on differences in true and false positive rates.
To evaluate the utility that the model confers, we use the notion of \textit{net benefit} \cite{vickers2006decision,Vickers2016} to define normalized expected utility measures that parameterize the relative value of the harms and benefits of statin initiation on the basis of decision thresholds recommended by clinical practice guidelines.
To evaluate net benefit in this setting, we adopt the assumption that the intervention induces constant relative risk reduction (section \ref{sec:net_benefit_risk_reduction}).
Furthermore, we use an inverse probability of censoring weighting (IPCW) approach to extend each of the training objectives and evaluation metrics used to account for censoring in ten-year ASCVD outcomes.
\section{Background and problem formulation} \label{sec:background}
\subsection{Supervised learning for binary outcomes} \label{sec:preliminaries}
Here, we introduce the formal notation and key assumptions used throughout the work.
Let $X \in \mathcal{X} = \mathbb{R}^m$ be a variable designating a vector of covariates and $Y \in \mathcal{Y} = \{0, 1\}$ be a binary indicator of an outcome.
We consider data that may be partitioned on the basis of a discrete indicator of a categorical attribute $A \in \mathcal{A} = \{A_k\}_{k=1}^{K}$ with $K$ categories. 
In some cases, $A$ may correspond to an attribute that describes partitions of the population, where the value of $A=A_k$ refers to a specific partition defined by the attribute. 
Examples of attributes used to partition the population include demographic attributes (\textit{e.g.} race, ethnicity, gender, sex, age subgroup) or strata defined by complex clinical phenotypes or comorbidity profiles.
We use the shorthand $\mathcal{D}_{A_k}$, when referring to the subset of the data $\mathcal{D}$ corresponding to the subgroup $A_k$.

The objective of supervised learning with binary outcomes is to use data $\mathcal{D} = \{(x_i, y_i, a_i)\}_{i = 1}^N \sim P(X, Y, A)$ to learn a function $f_{\theta} \in \mathcal{F}: \mathbb{R}^{m} \rightarrow [0,1]$ parameterized by $\theta$.
The function $f_{\theta}$ can be considered to be a risk estimator that, when optimal, estimates $\E[Y \mid X] = P(Y = 1 \mid X)$. 
We designate the random variable resulting from the application of the model $f_{\theta}$ to $X$ to be given by $S$, such that $S=f_{\theta}(X)$.
Given $S$, a predictor $\hat{Y}$ may be derived by comparing $S$ to a threshold $\tau_y \in [0, 1]$ to produce binary predictions $\hat{Y}(X) = \mathbbm{1}[f_{\theta}(X) \geq \tau_y] \in \{0, 1\}$. 

The \textit{calibration curve} $c: [0,1] \rightarrow [0, 1]$ is defined as a function that describes the expected value of $Y$ given $S$, such that $c(s) = E[Y \mid S=s]=P(Y=1 \mid S=s)$.
A model is said to be calibrated if $c(s)=s$ for all $s$.
The calibration curve can be used to assess the extent to which a model over or underestimates the risk of the outcome $Y$.
For instance, if $c(s')>s'$ then the observed event rate for the set of patients with scores of $s'$ is greater than $s'$, implying that the model underestimates risk for patients with scores of $s'$. 
Analogously, $c(s')<s'$ implies overestimation of risk for patients with scores of $s'$.

\subsection{Algorithmic fairness criteria}

Assessments of algorithmic fairness rely on \textit{fairness criteria}, \textit{i.e.} statistical properties reflecting moral or normative judgements as to the principles that constitute fairness.
A broad class of fairness criteria can be described in terms of \textit{metric parity} ($g_j(\cdot) \perp A$), which requires that one or more metrics $g_j: \mathcal{F} \times (\mathcal{X}, \mathcal{Y}) \rightarrow \mathbb{R}^+$ be equal across the subgroups defined by $A$.
Common instantiations of metric parity include \textit{equalized odds} ($\hat{Y} \perp A \mid Y$ or $S \perp A \mid Y$) \cite{Hardt2016}, which requires both the true positive rates and the false positive rates to be equal across subgroups, \textit{demographic parity} ($\hat{Y} \perp A$ or $S \perp A$) \cite{calders2009building}, which requires the rate at which patients are classified as belonging to the positive class is equal across subgroups, \textit{predictive parity} ($Y \perp A \mid \hat{Y}=1$) \cite{Chouldechova2017}, which requires parity in the positive predictive values, as well as criteria defined over other performance metrics \cite{Celis2018,cotter2019optimization}, including the AUC \cite{beutel2019fairness,narasimhan2020pairwise} or the average log-loss or empirical risk \cite{Williamson2019}.
Another important class of fairness criteria is defined over the calibration curve.
Within that class, we focus on the \textit{sufficiency} condition ($Y \perp A \mid S$) \cite{Liu2019,barocas-hardt-narayanan}, which requires the calibration curves for each subgroup be equal, and the \textit{group calibration} condition ($\E[Y \mid S=s, A]=s$) \cite{Kleinberg2016,pleiss2017fairness}, which requires the model to be calibrated for each subgroup.

\subsection{Assessing the utility and net benefit of decision-making at a threshold} \label{sec:utility}
To contextualize the presentation of algorithmic fairness, we present a utility-theoretic perspective on clinical decision-making.
For this framing, we consider a decision rule that implies intervention allocation on the basis of a binary predictor $\hat{Y}(X)=\mathbbm{1}[f_{\theta}(X) \geq \tau_y]$.

We define
\begin{equation}
    U_{\mathrm{cond}}(s) = U_{\mathrm{cond}}^1(s) - U_{\mathrm{cond}}^0(s)
\end{equation}
as the \textit{conditional} expected utility of the decision rule, where $U_{\mathrm{cond}}^1(s)$ designates the expected utility associated with treating patients whose predicted scores $S=f_{\theta}(X)$ are $s$, and $U_{\mathrm{cond}}^0(s)$ is the expected utility of \textit{not} treating patients whose scores are $s$.
We define the \textit{aggregate} expected utility $ U_{\textrm{agg}}(\tau_y)$ of the decision to be the average utility over the population, given that the intervention is allocated for all patients with scores at or above the threshold $\tau_y$:
\begin{equation} \label{eq:agg_utility_expectation}
    U_{\textrm{agg}}(\tau_y) = \E[U_{\mathrm{cond}}^1 \mid S \geq \tau_y] P(S \geq \tau_y) + \E[U_{\mathrm{cond}}^0 \mid S < \tau_y] P(S < \tau_y).
\end{equation}

The optimal decision rule for a fixed predictive model is one where the intervention is allocated to patients with scores for which $U_{\mathrm{cond}}(s) > 0$ and not allocated to those for which $U_{\mathrm{cond}}(s) < 0$.
If $U_{\mathrm{cond}}(s)$ is strictly monotonically increasing in $s$ and has a root in $[0,1]$ then the optimal threshold $\tau_y^*$ is given by the point at which $U_{\mathrm{cond}}(s=\tau_y^*)=0$.
When $U_{\mathrm{cond}}(s)$ is strictly monotonic but has no root in $[0,1]$, then either the treat-all ($\tau_y=0$) or treat-none ($\tau_y=1$) strategies is optimal.

In some cases, $U_{\mathrm{cond}}$ can be written as a simple function of the calibration curve.
For example, if the costs and benefits of decision-making can be written as fixed expected costs or utilities of true positive ($u_{\mathrm{TP}}$), false positive ($u_{\mathrm{FP}}$), true negative ($u_{\mathrm{TN}}$), and false negative ($u_{\mathrm{FN}}$) classification, then 
\begin{align}
        U_{\mathrm{cond}}(s)  = (u_{\mathrm{TP}} - u_{\mathrm{FN}})c(s) + (u_{\mathrm{FP}} - u_{\mathrm{TN}})(1-c(s))
\end{align}
and the optimal threshold is given by \cite{Sox2013,Corbett-Davies:2017:ADM:3097983.3098095}
\begin{equation} \label{eq:optimal_threshold}
    \tau_y^* = c^{-1}\Big(\frac{u_{\mathrm{TN}} - u_{\mathrm{FP}}}{u_{\mathrm{TN}} - u_{\mathrm{FP}} + u_{\mathrm{TP}} - u_{\mathrm{FN}}} \Big).
\end{equation}

It follows that when a model is calibrated, the optimal threshold is given by $\tau_y' = \frac{u_{\mathrm{TN}} - u_{\mathrm{FP}}}{u_{\mathrm{TN}} - u_{\mathrm{FP}} + u_{\mathrm{TP}} - u_{\mathrm{FN}}}$. When the model is miscalibrated, but the calibration curve is strictly monotonic, the optimal threshold is given the point at which the calibration curve intersects $\tau_y'$.
Furthermore, given the relationship between the $c(s)$ and $U_{\mathrm{cond}}$, monotonicity in the calibration curve implies monotonicity in the conditional utility, and setting a threshold on the basis of the calibration curve can be interpreted as setting a threshold on $U_{\mathrm{cond}}$.

To assess the expected utility of the decision rule over a population, it is typically not necessary to evaluate $U_{\textrm{agg}}(\tau_y)$ with equation (\ref{eq:agg_utility_expectation}).
Instead, a chosen decision threshold can be used to parameterize the \textit{net benefit} \cite{vickers2006decision,Vickers2016} of the decision rule under the assumption that the chosen threshold is optimal, for a calibrated model, based on the values of the decision maker and the effectiveness of the intervention.
The net benefit under the assumption of fixed costs or utilities of classification errors is given by \cite{vickers2006decision,Vickers2016}
\begin{equation}
    \mathrm{NB}(\tau_y; \tau_y^*) = P(S \geq \tau_y \mid Y = 1) P(Y=1) - P(S \geq \tau_y \mid Y = 0) P(Y=0) \frac{\tau_y^*}{1-\tau_y^*},
\end{equation}
where $\tau_y$ is the evaluated decision threshold and $\tau_y^*$ parameterizes the net benefit. 
This metric is fundamental to \textit{decision curve analysis} \cite{vickers2006decision,Vickers2016}, as a decision curve is the curve that results from evaluating net benefit for a range of thresholds for which $\tau_y=\tau_y^*$.
Both the net benefit and $U_{\mathrm{agg}}$ are maximized at the threshold that results from the application of equation (\ref{eq:optimal_threshold}) when the assumptions outlined above are met.

We introduce the notion of the \textit{calibrated net benefit} (cNB) to assess the net benefit under the assumption that the decision threshold used is adjusted on the basis of observed miscalibration.
If $c(s)$ is the calibration curve, then the calibrated net benefit evaluated at a threshold $\tau_y$ is given by the net benefit evaluated at a threshold $\tau_{\mathrm{c}}=c^{-1}(\tau_y)$ on the score $S$.
The calibrated net benefit under the assumption of fixed classification costs is given by
\begin{equation}
    \mathrm{cNB}(\tau_y; \tau_y^*) = P(S \geq c^{-1}(\tau_y) \mid Y = 1) P(Y=1) - P(S \geq c^{-1}(\tau_y) \mid Y = 0) P(Y=0) \frac{\tau_y^*}{1-\tau_y^*}.
\end{equation}

\subsection{Implications for algorithmic fairness} \label{sec:implications}
A key consequence of the analysis presented thus far is that, subject to the assumptions detailed in section \ref{sec:utility}, the optimal threshold rule applied to a predictive model that outputs a continuous-valued risk score is based directly on the calibration characteristics of the model and the assumed expected costs or utilities of classification errors that encapsulate the effectiveness of the intervention and the preferences for downstream benefits and harms.
As has been argued in related work \cite{Corbett-Davies:2017:ADM:3097983.3098095,corbett2018measure,bakalar2021fairness,foryciarz2021ascvd}, it follows that if the model is calibrated for each subgroup, the decision threshold that maximizes expected utility and net benefit for each subgroup is the same when the expected utilities associated with each classification error do not change across subgroups.
We verify this claim in simulation in supplementary section \ref{fig:simulation} (Supplementary Figure \ref{fig:simulation}).
Furthermore, in this case, sufficiency implies that the use of a consistent threshold on the risk score for all subgroups corresponds to the use of a consistent threshold on the conditional utility $U_{\textrm{cond}}$ across subgroups, corresponding to an intuitive notion of \textit{fairness} even in the case that the chosen decision threshold is not necessarily optimal \cite{Corbett-Davies:2017:ADM:3097983.3098095,bakalar2021fairness}.
However, we note that this can still be a misleading notion of fairness given that it does not account for heterogeneity in the outcome not accounted for by the model under consideration \cite{Corbett-Davies:2017:ADM:3097983.3098095}.

As is described in prior work \cite{Liu2019,Kleinberg2016,Chouldechova2017,Corbett-Davies:2017:ADM:3097983.3098095,simoiu2017problem}, one should expect models that minimize the empirical risk for the population overall, with respect to a data distribution containing features $X$ that encode $A$, to satisfy fairness criteria defined in terms of the calibration curve, including sufficiency and group calibration, and to violate equalized odds, demographic parity, and predictive parity when such models exhibit differences in the distribution of the risk score $S$ or in the prevalence or incidence of the outcome $Y$.
As discussed in \citet{Liu2019}, the use of ERM with a sufficiently large training dataset drawn from the target population is consistent with learning a model satisfying that criteria, implying that such models are expected to be calibrated overall and for each patient subgroup, but are expected to have non-trivial differences in true and false positive error rates, as well as in other performance metrics, when the incidence of the outcome or distribution of risk varies over those subgroups \cite{Liu2019,Kleinberg2016,simoiu2017problem}.

Consequently, approaches undertaken to constrain the model training objective \cite{Agarwal2018,pmlr-v54-zafar17a,cotter2019training,cotter2019optimization,Celis2018,Pfohl2021a} to minimize violation of fairness criteria such as equalized odds or demographic parity typically reduce utility through some combination of explicit threshold adjustment \cite{Hardt2016} towards a threshold unrelated to the one selected on the basis of preference solicitation in the context of the intervention, induced miscalibration that analogously implies decision-making at a threshold unrelated to the utility-maximizing one \cite{foryciarz2021ascvd}, or reduction in model fit \cite{Pfohl2021a}.
Given the relationship between the calibration curve and the conditional utility described in section \ref{sec:utility}, induced miscalibration that results in sufficiency violation implies that the use of a consistent threshold on the score across subgroups results in the use of different thresholds on $U_{\mathrm{cond}}$ across subgroups.

\subsection{Algorithmic fairness training objectives} \label{sec:training_objectives}
We evaluate training objectives that incorporate algorithmic fairness goals and constraints into their specification.
We do so not to advocate for the use of their use, but rather to develop evidence as to the extent to which theoretical properties and trade-offs manifest empirically.
We focus our efforts on ``in-processing'' approaches \cite{Agarwal2018,pmlr-v54-zafar17a,cotter2019training,cotter2019optimization,Celis2018} rather than on pre- \cite{Zemel2013,Louizos2015,Madras2018,song2019learning,Ilvento2020} or post-processing \cite{Hardt2016,barocas-hardt-narayanan} (\textit{e.g.} threshold-adjustment) approaches because in-processing approaches are well-suited to learning models that achieve the minimum achievable trade-off between measures of model performance and fairness in practical finite-sample settings \cite{Woodworth2017} and further allow for exploration of smooth trade-offs induced by relaxation of the constraint \cite{Agarwal2018,cotter2019optimization}.
We specifically focus on scalable gradient-based learning procedures that use regularized objectives to penalize violation of fairness criteria in a minibatch setting, to enable the use of these procedures for deep neural network models learned with large-scale datasets.
We investigate approaches that, rather than constraining for parity in a metric across subgroups, attempts to improve the worst-case value of the metric over subgroups using distributionally robust optimization (DRO) \cite{Sagawa*2020Distributionally,diana2021minimax,martinez2020minimax,Chen2017_robust,pfohl_dro}.

Following \citet{Pfohl2021a}, the regularized training objective is ERM that incorporates a non-negative penalty term $R$ that assesses the extent to which a fairness criterion of interest is violated and a non-negative parameter $\lambda$ that may be tuned to control the extent to which violation of the criteria is penalized:
\begin{equation} \label{eq:regularized_objective}
    \min_{\theta \in \Theta} \sum_{i=1}^N w_i \ell(y_i, f_{\theta}(x_i)) + \lambda R,
\end{equation}
where $w_i$ are sample weights. 
In our experiments, we use this formulation to penalize violation of equalized odds and differences in AUC and log-loss across subgroups.
To penalize violation of equalized odds, we primarily use a term that penalizes the Maximum Mean Discrepancy (MMD) \cite{gretton2012kernel} between the distribution of scores between each patient subgroup and the overall population conditioned on the observed values of the outcome $Y$, as in \citet{Pfohl2021a}:
\begin{equation} \label{eq:reg_objective_eo}
    \min_{\theta \in \Theta} \sum_{i=1}^N w_i \ell(y, f_{\theta}(x)) + \lambda \frac{1}{K} \sum_{Y_j \in \mathcal{Y}} \sum_{A_k \in \mathcal{A}} \hat{D}_{\mathrm{MMD}}(P(f(X) \mid A=A_k, Y=Y_j) \mid \mid P(f(X)\mid Y=Y_j)).
\end{equation}
A full specification of the MMD-based training objective is included in supplementary section \ref{sec:supp:training_objectives}.

We further use a regularized objective defined on the basis of a penalty that assesses violation of metric parity to penalize differences in the AUC or log-loss between each subgroup with the overall population:
\begin{equation} \label{eq:reg_objective_metric_parity}
    \min_{\theta \in \Theta} \sum_{i=1}^N w_i \ell(y, f_{\theta}(x)) + \lambda \sum_{j=1}^{J} \sum_{A_k \in \mathcal{A}} \big(g_j(f_{\theta}, \mathcal{D}_{A_k}) - g_j(f_{\theta}, \mathcal{D}) \big)^2.
\end{equation}
We also evaluate the use of this objective to penalize violation of equalized odds at relevant thresholds by plugging surrogates of the true and false positive rates into equation ($\ref{eq:reg_objective_metric_parity}$).
A full specification of the relevant objectives is provided in supplementary section \ref{sec:supp:training_objectives}.

Beyond regularized objectives for algorithmic fairness, we evaluate distributionally robust optimization \cite{ben2013robust,Hu2018,Sagawa*2020Distributionally} procedures that encode the goal of maximizing worst-case performance over subgroups as one of learning to be robust over marginal shifts in the proportion of data available from each subgroup.
The use of these objectives reflects a shift in perspective from the goal of requiring that some statistic be equal across subgroups towards one of aiming to identify models that perform well for each subgroup \cite{Sagawa*2020Distributionally,Hu2018,diana2021minimax,martinez2020minimax,pfohl_dro}.
In this work, we leverage the \textit{GroupDRO} framework (hereafter referred to as DRO) developed in \citet{Sagawa*2020Distributionally} and extended in \citet{pfohl_dro}.
The algorithm is implemented as the following alternating updates conducted over minibatches:
\begin{equation} \label{eq:DRO_flexible}
    \lambda_k \leftarrow \lambda_k \exp\big(\eta g(f_{\theta}, \mathcal{D}_{A_k})\big) / \sum_{k=1}^K \exp \big(\eta g(f_{\theta}, \mathcal{D}_{A_k})\big)
\end{equation}
and
\begin{equation} \label{eq:DRO_theta_ipcw}
    \min_{\theta \in \Theta} \sum_{k=1}^{K} \lambda_k \sum_{i = 1}^{n_k} w_i \ell(y_i, f_{\theta}(x_i)),
\end{equation}
where $\eta$ is a non-negative scalar hyperparameter, $\{\lambda_k\}_{k=1}^K$ are non-negative scalars that sum to 1, and $g$ is a performance metric where lower values of the metric indicate better performance.
In our experiments, we evaluate the use of the log-loss and $1-\textrm{AUC}$ as the choice of metric $g$, as in \citet{pfohl_dro}.

\section{Case study in atherosclerotic disease risk estimation}
\subsection{Background on ASCVD risk estimation for statin initiation}
Clinical practice guidelines for the primary prevention of cardiovascular disease recommend the use of estimates of ten-year atherosclerotic cardiovascular disease (ASCVD) risk to inform the initiation of cholesterol-lowering statin therapy \cite{goff20142013,Stone2014,Grundy2019,arnett2019,Lloyd-Jones2019}.
These guidelines primarily recommend the use of risk estimates provided by the Pooled Cohort Equations \cite{goff20142013} and its extensions \cite{yadlowsky2018clinical}.
However, these estimates have been reported to systematically over-estimate or under-estimate risk in ways that are consequential for the appropriateness of downstream treatment decisions. 
This misestimation has been reported to occur both overall \cite{DeFilippis2015,Cook2016,pencina2014application,Rana2016-ae} and for subgroups defined on the basis of race/ethnicity \cite{defilippis2017risk,jung2015acc,afkarian2016diabetes}, sex \cite{DeFilippis2015,Cook2016,mora2018evaluation}, socioeconomic status \cite{Lloyd-Jones2019}, or for patients with comorbidities which influence ASCVD risk or the expected benefit and harms of statin therapy, including diabetes \cite{Rana2016-ae,afkarian2016diabetes}, chronic kidney disease (CKD) \cite{afkarian2016diabetes,jacobson2015national,harper2008managing}, and rheumatoid arthritis (RA) \cite{ozen20162013,Oever2013}.
Approaches undertaken to address these issues include the development of new risk estimators from large, diverse observational cohorts using modern machine learning methods \cite{yadlowsky2018clinical,Ward2020,kakadiaris2018machine,Pfohl2019AIES,Zhao2021}, revisions to guidelines to encourage follow-up testing when the benefits of statin therapy are unclear and shared patient-clinician decision-making to incorporate patient preferences and other context \cite{Lloyd-Jones2019}, and the incorporation of fairness constraints into the model development process \cite{Pfohl2019AIES,barda2021addressing,foryciarz2021ascvd}.

\subsection{Extending the approach}
\subsubsection{Supervised learning with censored binary outcomes} \label{sec:censoring}
When the binary outcome $Y = \mathbbm{1}[T \leq \tau_t]$ is defined as the occurrence of the outcome event at a time $T$ at or before a time horizon $\tau_t \in \mathbb{R}^+$, it is important to account for censoring.
The presence of censoring implies that either the outcome event time $T \in \mathbb{R}^+$ or the censoring time $C \in \mathbb{R}^+$ will be observed, but not both.
The outcome data in an observed dataset $\mathcal{D} = \{(x_i, u_i^t, \delta_i^t, a_i)\}_{i = 1}^N$ is represented by an observed follow-up time $U^t = \min(T, C)$ and an indicator $\Delta^t = \mathbbm{1}[T \leq C]$ that reflects whether the observed follow-up time corresponds to an outcome or a censoring event. 
The binary outcome $Y$ is said to be censored if the censoring time $C$ occurs prior to both the observed follow-up time and the time horizon, \textit{i.e.} $C < T$ and $C < \tau_t$.
We define a composite observed follow-up time $U^y = \min(T, C, \tau_t)$ for the binary outcome and an indicator $\Delta^y = 1 - \mathbbm{1}[C < T]*\mathbbm{1}[C < \tau_t]$ that reflects whether a patient's binary outcome is uncensored. 
A visual depiction of the relationship between the outcome and censoring event times and the value and censoring status of the binary outcome is shown in supplementary figure \ref{fig:censoring}.

The use of inverse probability of censoring weighting (IPCW) allows for the derivation and evaluation of predictive models for censored binary outcomes \cite{Robins1992,Molinaro2004,van2003unified,blanche2013review,uno2007evaluating}, analogous to propensity score weighting procedures used for causal effect estimation \cite{imbens2015causal}.
The appropriate weights are those that are proportional to the inverse probability of remaining uncensored at the time of the composite observed follow-up time.
Specifically, for an estimate of the censoring survival function $G(s, x) = P(C > s \mid X=x)$ we define normalized weights 
\begin{equation} \label{eq:default_weights}
    w_i = \frac{\delta_i^y}{G(u_i^y, x_i)} \Big(\sum_{i=1}^N \frac{\delta_i^y}{G(u_i^y, x_i)}\Big)^{-1}    
\end{equation}
that for a patient $i$ reflects the reciprocal of the conditional probability of remaining uncensored at the time $u^y_i$ given features $x_i$.
To enable this approach, we make the following assumptions: (1) \textit{coarsening at random} \cite{Robins1992,Robins2000} where the outcome event time is independent of the censoring time conditioned on the features, \textit{i.e.} $T \perp C \mid X$, (2) $G(U, X) > 0$ for all data with uncensored binary outcomes (for which $\Delta^y = 1$), and (3) that $f_{\theta}$ is a deterministic transformation.

The IPCW weights may be derived with any procedure that allows for learning a conditional model for the censoring survival function.
In our experiments, we use flexible neural network models in discrete time, such as those described in \citet{Kvamme2019}.
Given these weights, the unconstrained model fitting procedure is weighted ERM.
We extend each of the metrics used for evaluation and or as components of the training objectives presented in section \ref{sec:training_objectives} to account for censoring by incorporating IPCW weights. 
A full specification of the relevant metrics and training objectives is provided in supplementary section \ref{sec:supp:training_objectives}.

\subsubsection{Assessing net benefit in terms of risk reduction} \label{sec:net_benefit_risk_reduction}
For the evaluation of models that predict the risk of ASCVD to inform statin initiation, we introduce an alternative formulation of the net benefit that is defined in terms of the population absolute risk reduction after subtracting out harms represented on the same scale.
We use the guideline-concordant thresholds of 7.5\% and 20\%, which correspond to the bounds of the intermediate and high-risk categories, respectively in clinical practice guidelines \cite{goff20142013,arnett2019,Lloyd-Jones2019}. 
We do so to parameterize the net benefit in terms of clinically-plausible benefit-harm trade-offs. 
Here, we summarize the key aspects of the formulation, but include a full derivation in supplementary section \ref{sec:supp:net_benefit_risk_reduction}.

For this case, the relevant utilities are defined by the absence ($u_0^y$) and presence ($u_1^y$) of an ASCVD event within ten years. 
The expected event rates conditioned on the score $s$ are given by $p_y^0(s)$ and $p_y^1(s)$ in the absence and presence of treatment, respectively. 
The conditional absolute risk reduction is given by $\mathrm{ARR}(s) = p_y^0(s) - p_y^1(s)$.
We assume that the expected harm of the intervention can be represented as a constant $k_{\textrm{harm}}$ that is independent of the risk estimate.
With these assumptions, $U_{\mathrm{cond}}(s) = \big(u_0^y - u_1^y\big)\mathrm{ARR}(s) - k_{\mathrm{harm}}$ and the optimal threshold is given by $\tau_y^* = \mathrm{ARR}^{-1}\big(\frac{k_{\mathrm{harm}}}{u_0^y - u_1^y}\big)$.

We further assume that the intervention induces constant \textit{relative} risk reduction, such that $\mathrm{ARR}(s)=r c(s)$ for a constant $r \in (0,1)$ and the conditional expected utility and optimal threshold are simple transformations of the calibration curve, as was the case for the fixed-cost setting. 
In this case, $U_{\mathrm{cond}}(s) = \big(u_0^y - u_1^y\big) rc(s) - k_{\mathrm{harm}}$ and $\tau_y^* = c^{-1}\big(\frac{k_{\mathrm{harm}}}{r(u_0^y-u_1^y)}\big)$. We derive a formulation of the net benefit in this setting as
\begin{equation} \label{eq:net_benefit_risk_reduction}
    \mathrm{NB}(\tau_y; \tau_y^*) = -(1-\mathrm{NPV}(\tau_y))P(S < \tau_y) - 
        P(S \geq \tau_y)\Big((1-r)\mathrm{PPV}(\tau_y) + r \tau_y^*\Big) + P(Y=1),
\end{equation}
where $\mathrm{NPV}(\tau_y)$ and $\mathrm{PPV}(\tau_y)$ are the negative and positive predictive values evaluated at a threshold $\tau_y$.
The calibrated net benefit is defined analogously:
\begin{align} \label{eq:calibrated_net_benefit_risk_reduction}
    \begin{split}
        \mathrm{cNB}(\tau_y; \tau_y^*) & = -(1-\mathrm{NPV}(c^{-1}(\tau_y)))P(S < c^{-1}(\tau_y)) \hspace{1mm} \ldots \\
        & \hspace{10mm} - P(S \geq c^{-1}(\tau_y))\Big((1-r)\mathrm{PPV}(c^{-1}(\tau_y)) + r \tau_y^*\Big) + P(Y=1).    
    \end{split}
\end{align}

To operationalize this notion of net benefit, we use a simple model for the treatment effect of statin initiation presented in \citet{soran2015cholesterol}. 
This model relates the expected reduction in ASCVD risk to the reduction in low-density lipoprotein cholesterol (LDL-C) that results from statin initiation. 
It assumes that each 1 mmol/L reduction in LDL-C results in an expected 22\% proportional reduction in the ten-year risk of ASCVD, based on evidence from a meta-analysis of randomized control trials \cite{trialists2005efficacy}.
This implies that if the absolute reduction in LDL-C in mmol/L is given by $\kappa$, the relative reduction in ten-year ASCVD risk is given by $r=1-(1-0.22)^\kappa$ \cite{soran2015cholesterol}.
Therefore, the task of describing the expected reduction in risk as a function of the risk estimate can be reduced to the task of describing the expected reduction in LDL-C as a function of the risk estimate.

To describe the expected reduction in LDL-C from statin therapy for the cohort, we separately consider the evidence for the extent to which statins reduce LDL-C as a function of LDL-C alongside the relationship between observed LDL-C values and the risk estimates for the cohort.
As in \citet{soran2015cholesterol}, we assume the use of moderate intensity statin therapy that reduces LDL-C by 43\% on average, independent of the pre-treatment level of LDL-C, consistent with the usage of 20 mg of atorvastatin \cite{soran2015cholesterol,national2014cardiovascular,collins2016interpretation}.
We extract the most recent historical LDL-C result, if present, for each patient in the test set whose binary outcome was uncensored, filtering out extreme results of $<10$ or $>500$ mg/dL LDL-C, resulting in 32,366 results.
We note that the risk estimates produced by the selected model learned with ERM appear to be uncorrelated with observed untreated LDL-C levels in the cohort ($R^2=0.004$; Supplementary Figure \ref{fig:supplement/ldl_scatter}), suggesting that both the expected absolute reduction in LDL-C and the relative risk reduction $r$ may be modeled as constants that are independent of the risk estimates.
We extract a risk-score-independent estimate of the mean LDL-C in the cohort as 3.01 mmol/L, using an IPCW-weighted mean over the extracted LDL-C values.
The value for the expected relative risk reduction that follows from statin initiation is given by $r=1-(1-0.22)^{(3.01*0.43)}=0.275=27.5\%$.

\begin{table}[!th]
\centering
\caption{Characteristics of the cohort drawn from the Optum CDM database. Data are grouped based on sex, racial and ethnic categories, and the presence of type 2 and type 1 diabetes, rheumatoid arthritis (RA), and chronic kidney disease (CKD). Shown, for each subgroup, is the number of patients extracted, the rate at which the ten-year ASCVD outcome is censored, and an inverse probability of censoring weighted estimate of the incidence of the ten-year ASCVD outcome.}
\label{tab:cohort_optum}
\begin{tabular}{lrrr}
\toprule
 Group &      Count &  Censoring rate &  Incidence \\
\midrule
Female                  &  3,253,609 &           0.816 &      0.105 \\
Male                    &  2,549,256 &           0.821 &       0.120 \\
\midrule
Asian                   &    165,198 &           0.814 &     0.0829 \\
Black                   &    438,144 &           0.786 &      0.136 \\
Hispanic                &    433,238 &             0.800 &      0.104 \\
Other                   &    880,116 &           0.936 &      0.115 \\
White                   &  3,886,169 &           0.797 &       0.110 \\
\midrule
Asian, female           &     88,100 &           0.806 &     0.0793 \\
Asian, male             &     77,098 &           0.823 &     0.0874 \\
Black, female           &    262,559 &           0.784 &      0.128 \\
Black, male             &    175,585 &           0.788 &       0.150 \\
Hispanic, female        &    235,736 &           0.792 &      0.102 \\
Hispanic, male          &    197,502 &            0.810 &      0.107 \\
Other, female           &    522,369 &           0.938 &      0.108 \\
Other, male             &    357,747 &           0.932 &      0.125 \\
White, female           &  2,144,845 &           0.794 &      0.102 \\
White, male             &  1,741,324 &           0.802 &      0.119 \\
\midrule
Type 2 diabetes absent  &  5,388,193 &           0.817 &      0.104 \\
Type 2 diabetes present &    414,672 &           0.835 &        0.20 \\
\midrule
Type 1 diabetes absent  &  5,741,282 &           0.818 &       0.110 \\
Type 1 diabetes present &     61,583 &           0.825 &       0.240 \\
\midrule
RA absent               &  5,733,505 &           0.819 &       0.110 \\
RA present              &     69,360 &           0.782 &      0.185 \\
\midrule
CKD absent              &  5,758,773 &           0.819 &       0.110 \\
CKD present             &     44,092 &           0.767 &      0.253 \\
\bottomrule
\end{tabular}
\end{table}

\subsection{Cohort definition}
All data are derived from Optum’s de-identifed Clinformatics\textregistered \hspace{0.5mm}Data Mart Database (Optum CDM), a statistically de-identified large commercial and medicare advantage claims database containing records from 2007 to 2019.
We utilize version 8.1 of the database mapped to the Observational Medical Outcomes Partnership Common Data Model (OMOP CDM) version 5.3.1 \cite{Hripcsak2015,Overhage2012,Reps2018}.
Approval for the use of this data for this study was granted by the Stanford Institutional Review Board protocol \#46829.

% Administrative Panel on Human Subjects in Medical Research (IRB 8 - OHRP \#00006208, protocol \#57916), with a waiver of informed consent.

We apply criteria to extract a cohort for learning estimators of ten-year ASCVD risk that mirrors the population eligible for risk-based allocation of statins based on clinical practice guidelines \cite{arnett2019}.
The characteristics of the extracted cohort are provided in Table \ref{tab:cohort_optum}.
We consider as candidate index events all office visits and outpatient encounters for patients between 40 and 75 years of age at the time of the visit for patients without a prior statin prescription or history of cardiovascular disease (Supplementary Table \ref{tab:ch5:cohort_concepts}).
We restrict the set of candidate index events to those recorded as occurring at or before December 31, 2008 for which least one year of historical data is available, and randomly sample one of the resulting candidate index events per patient for inclusion in the final cohort.

The times of ASCVD and censoring events are identified relative to the index event dates.
ASCVD events are defined as the occurrence of a diagnosis code for myocardial infarction, stroke, or fatal coronary heart disease (Supplementary Table \ref{tab:ch5:cohort_concepts}). 
We consider coronary heart disease to be fatal if death occurs within a year of the recording of the diagnosis code.
Censoring events are identified as the earliest date of statin prescription (Supplementary Table \ref{tab:ch5:cohort_concepts}), death, or the end of the latest enrollment period.
From the extracted ASCVD and censoring times, we construct composite binary outcomes and censoring indicators at ten years, following the logic of section \ref{sec:censoring}.

\subsubsection{Subgroup definitions}
We define discrete subgroups on the basis of (1) a combined race and ethnicity variable based on reported racial and ethnic categories, (2) patient sex, (3) intersectional categories describing intersections of racial and ethnic categories with sex,
(4) history of either type 2 diabetes, type 1 diabetes, RA, or CKD at the index date.
To construct the race and ethnicity attribute, we assign ``Hispanic'' if the recorded OMOP CDM concept for ethnicity is recorded as ``Hispanic or Latino'', and the value of the recorded OMOP CDM racial category otherwise. 
This resulted in a final categorization of ``Asian'', ``Black or African American'', ``Hispanic or Latino'', ``Other'', and ``White'', which we shortened to ``Asian'', ``Black'', ``Hispanic'', ``Other'', and ``White'' for succinctness in the presentation of the results.
We identify patients with a history of type 2 diabetes, type 1 diabetes, rheumatoid arthritis, or chronic kidney disease using the presence of a concept identifier indicative of the condition recorded prior to the index date (Supplementary Table \ref{tab:ch5:cohort_concepts}).
The selected concept identifiers used for identifying type 2 and type 1 diabetes are adapted from \citet{Reps2020}; those used to identify chronic kidney disease are adapted from \citet{suchard2019comprehensive}.

\subsection{Feature extraction}
We apply a procedure similar to the one described in \citet{pfohl_dro} to extract a set of clinical features to use as input to fully-connected feedforward neural networks and logistic regression models.
This procedure concatenates features representing unique OMOP CDM concepts recorded prior to each patient's selected index date. 
We use OMOP CDM concepts corresponding to time-agnostic demographic features (race, ethnicity, sex, and age discretized in five-year intervals) as well as longitudinal recorded diagnoses, medication orders, medical device usage, encounter types, laboratory test orders, flags indicating whether the test results were normal or abnormal based on reference ranges, and other coded clinical observations binned in three time intervals corresponding to 29 to 1 days prior to the index date, 365 days to 30 days prior to the index, and any time prior to the index date. 

\subsection{Data partitioning}
\label{sec:partitioning}
We partition the cohort such that 62.5\% is used as a training set, 12.5\% is used as a validation set, and 25\% of the data is used as a test set.
We subsequently partition the training data into five equally-sized partitions.
We train five models for each hyperparameter configuration, holding out one of the partitions of the training set for use as a development set to assess early stopping criteria, and perform model selection based on algorithm-specific model selection criteria defined over the average performance of the five models on the validation set.

\subsection{Derivation of inverse probability of censoring weights}
We consider the estimation of the risk of ASCVD at a fixed time horizon as an example of a supervised learning problem with a censored binary outcome, using the procedures described in section \ref{sec:censoring}.
To derive IPCW weights, we utilize neural networks trained with the discrete-time likelihood \cite{Kvamme2019,Gensheimer2019,tutz2016modeling} to estimate the censoring survival function conditioned on the full set of features used to fit the model for ten-year ASCVD.
For each cohort, we derive five such models using the training set partitioning strategy described in section \ref{sec:partitioning}. 
We use a fixed model architecture with one hidden layer of 128 hidden units that predicts the discrete-time hazard in twenty intervals whose boundaries are determined by the quantiles of the observed censoring times in the union of the four training set partitions that are not held-out.
We train these models in a minibatch setting and perform early stopping if the discrete-time likelihood does not improve for twenty-five epochs of 100 minibatches.
Subsequently, we define IPCW weights for each patient in the training set by taking the inverse of the predicted censoring survival function at the minimum of the time of censoring, the ASCVD outcome event, or ten years, for each patient, using the model trained on the set of training set partitions that exclude the patient.
The weights for patients in the validation and test sets are derived as the reciprocal of the average estimate of the censoring survival function derived from the five models.

\subsection{Experiments}
Here, we outline the structure of the experiments. 
To serve as baseline comparators for all experiments, we train models using unconstrained IPCW-weighted ERM without stratification.
We refer to this setting as \textit{pooled ERM}.
The first experiment aims to evaluate strategies to learn models that predict the outcome well for subgroups defined following stratification by race, ethnicity, and sex, including intersectional categories, and for patients with ASCVD-promoting comorbidities.
The second experiment aims to assess the implications of penalizing violation of the equalized odds criterion across subgroups defined on the basis of race, ethnicity, and sex. 
In each case, we evaluate the net benefit of statin initiation on the basis of the risk estimates under the assumption that the observed relationship with the benefits of using the ASCVD risk estimator to initiate moderate-intensity statin therapy can be modeled as inducing constant relative risk reduction (section \ref{sec:net_benefit_risk_reduction}), the expected harm of treatment is assumed not to vary on the basis of the risk estimate, and that the trade-off between benefits and harms reflects the choice of a decision threshold of either 7.5\% or 20\%.

\subsubsection{Unconstrained empirical risk minimization without stratification}
We evaluate feedforward neural networks and logistic regression models trained with pooled ERM in a minibatch setting using stochastic gradient descent.
We conduct a grid search over model-specific and algorithm-specific hyperparameters.
For feedforward neural networks, we evaluate a grid of hyperparameters that include learning rates of $1 \times 10^{-4}$ and $1 \times 10^{-5}$, one and three hidden layers of size 128 or 256 hidden units, and a dropout probability of 0.25 or 0.75.
For logistic regression models, we use weight decay regularization \cite{loshchilov2018decoupled} drawn from a grid of values containing 0, 0.01, and 0.001.
The training procedure is conducted in a minibatch setting of up to 150 iterations of 100 minibatches of size 512 using the Adam \cite{kingma2014adam} optimizer in the Pytorch framework \cite{pytorch}. 
We use an early-stopping rule that returns the model with the lowest log-loss evaluated on the development set when the log-loss has not improved for twenty-five epochs of 100 minibatches.
The procedure is repeated separately for each of the five training/development set partitions.
Following training, we apply each model derived from the training procedure to the validation set and select hyperparameters on the basis of the best average log-loss evaluated in the validation set across all training partitions.

\subsubsection{Approaches to improve model performance over patient subgroups} \label{sec:experiment_performance}
To compare with pooled ERM, we evaluate models trained with ERM separately on each subgroup (\textit{stratified ERM}), models trained with IPCW-weighted regularized training objectives that penalize differences in the log-loss or AUC between each subgroup and the overall population, and IPCW-weighted DRO objectives that target the worst-case log-loss or AUC across subgroups.
The hyperparameter grid, early stopping, and model selection procedures conducted for the stratified ERM experiments exactly match those used for the pooled ERM experiments. 
For models trained with regularized objectives or DRO, we use a feedforward neural network with hyperparameters fixed to three hidden layers with 256 hidden units, a dropout probability of 0.25, and a learning rate of $1 \times 10^{-4}$.
For the regularized models, we evaluate a grid of five $\lambda$ values distributed log-uniformly from $1 \times 10^{-2}$ to $10$ and conduct early-stopping on the basis of the value of the penalized loss.
For the DRO experiments, we evaluate unmodified and balanced sampling by subgroup, as well as a grid of value of $\eta$ given by 0.01, 0.1, and 1. 

As in the case of the unpenalized ERM experiments, we fix the batch size to be 512 and evaluate early-stopping criteria in intervals of 100 minibatches and terminate when the criterion has not improved for 25 iterations.
For the fairness-regularized models, we perform early stopping on the basis of the penalized loss that incorporates the regularization term.
To conduct early-stopping for DRO experiments, we use worst-case early-stopping criteria \cite{pfohl_dro} that returns the model with the best worst-case subgroup AUC or log-loss observed over the training procedure when the worst-case value has not improved for twenty-five epochs of 100 minibatches.
We use the worst-case subgroup AUC for early-stopping when the AUC-based training objective is used and the worst-case subgroup log-loss when the DRO objective defined over the log-loss is used.

For model selection on the validation set, we use criteria defined in terms of the worst-case performance (either AUC or log-loss) for both regularized and DRO experiments, over the full set of hyperparameter configurations.
We use the worst average performance produced by averaging validation set performance over the training replicates.
As was the case for early-stopping, we use the worst-case AUC for model selection for the regularized and DRO experiments that incorporate the AUC into their objective, and use the worst-case log-loss for model selection for objectives that incorporate the log-loss into their objective.

\subsubsection{Regularized fairness objectives for equalized odds} 
To evaluate the effect of penalizing violation of equalized odds, we use regularized training objectives that incorporate an IPCW-weighted MMD penalty to penalize differences in the outcome-conditioned distribution of the risk score between each subgroup and the marginal population (equations (\ref{eq:reg_objective_eo}) and (\ref{eq:weighted_mmd_kernel})), as well as a penalty that penalizes differences in the true positive and false negative rates between each subgroup and the marginal population at the guideline-relevant thresholds of 7.5\% and 20\% \cite{arnett2019} using an IPCW-weighted objective that uses a softplus relaxation to the indicator function (equation (\ref{eq:reg_objective_metric_parity_approx})).
To simplify the experiment, we conduct this analysis only with the intersectional categories defined by race, ethnicity, and sex, but separately evaluate the models on the intersectional categories and for race/ethnicity and sex separately.
Furthermore, we fix hyperparameters to those used for the regularized and DRO models in section \ref{sec:experiment_performance} and evaluate five values of the regularization penalty $\lambda$ distributed log-uniformly from $1 \times 10^{-2}$ to $10$.
As before, we fix the batch size to be 512 and evaluate early-stopping criteria in intervals of 100 minibatches and terminate when the value of the penalized loss has not improved for 25 iterations.
For these models, we do not conduct explicit model selection over the regularization path on the basis of validation set performance given that it was of interest to evaluate each value of $\lambda$ separately.

\subsubsection{Evaluation of model performance}
To compute 95\% confidence intervals for model performance metrics, we draw 1,000 bootstrap samples from the test set, stratified by levels of the outcome and subgroup attribute relevant to the evaluation, compute the IPCW-weighted performance metrics for the set of five derived models on each bootstrap sample, and take the 2.5\% and 97.5\% empirical quantiles of the resulting distribution that results from pooling over both the models and bootstrap replicates.
We construct analogous confidence intervals for the difference in the model performance relative to pooled ERM by computing the difference in the performance computed on the same bootstrap sample and taking the 2.5\% and 97.5\% empirical quantiles of the distribution of the differences. 
To construct confidence intervals for the worst-case performance over subgroups, we extract the worst-case performance for each bootstrap sample.

We assess model performance in the test set in terms of IPCW-weighted variants of the AUC, the average log-loss, the absolute calibration error (ACE) \cite{Pfohl2021a,Austin2019,Yadlowsky2019}, true positive rate, false positive rate, calibration curve, and the net benefit.
Estimates of the calibration curve used to compute the ACE and calibrated net benefit rely on an estimate of the calibration curve learned via a logistic regression estimator trained on the test data to predict the outcome from a logit-transformed outputs of the predictive model as inputs. 
The inverse of the calibration curve used to compute the calibrated net benefit is derived analytically based on the coefficients of the learned logistic regression model.
To compute the ACE, we use an IPCW-weighted average of the absolute value of the differences between the model output and the calibration curve.

\section{Results}
\begin{figure*}[!t]
	\centering
	\includegraphics[width=0.95\linewidth]{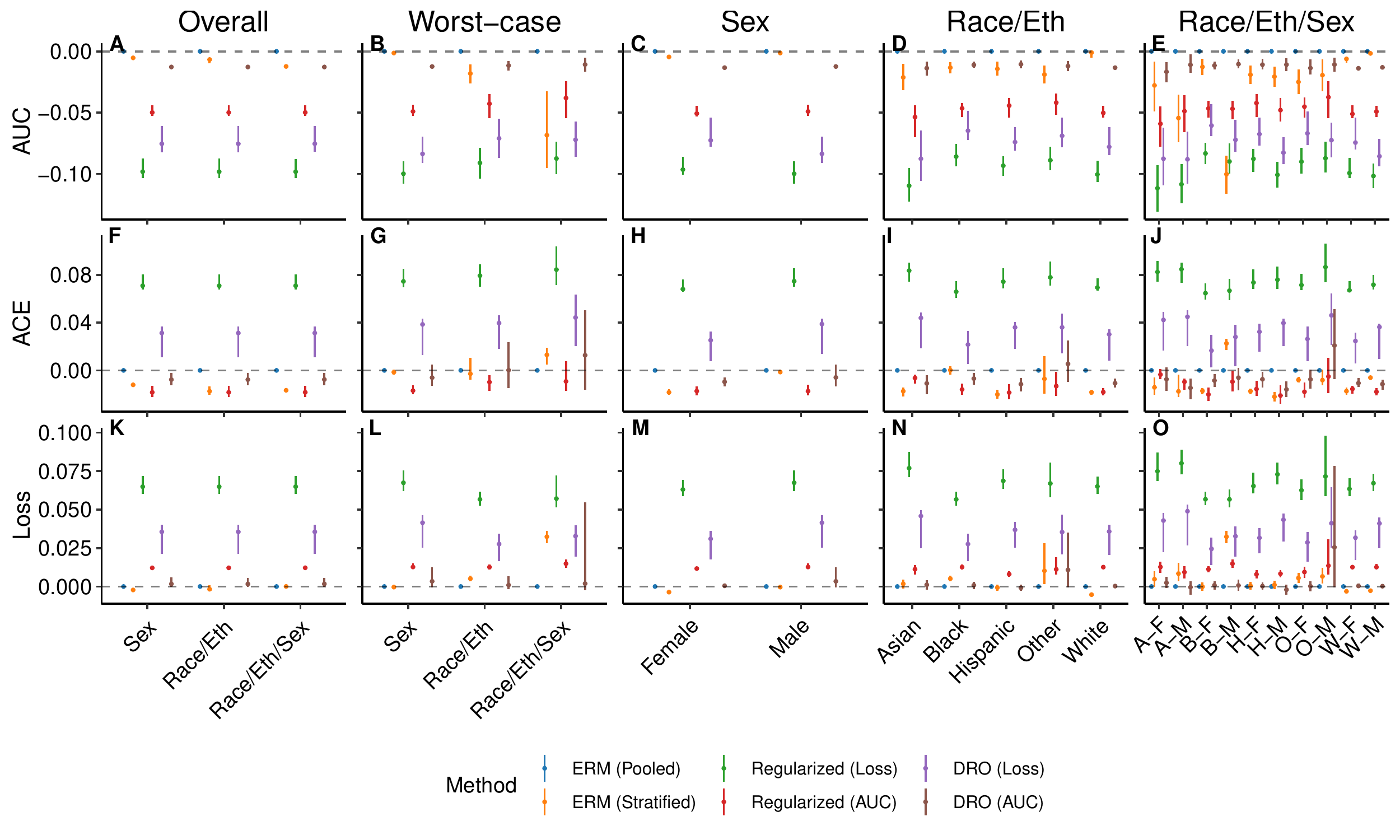}
	\caption{
    	The performance of models that estimate ten-year ASCVD risk, stratified by race, ethnicity, and sex, relative to the results attained by the application of unpenalized ERM to the overall population.
    	Results shown are the relative AUC, absolute calibration error (ACE), and log-loss assessed in the overall population, on each subgroup, and in the worst-case over subgroups following the application of unconstrained pooled or stratified ERM, regularized objectives that penalize differences in the log-loss or AUC across subgroups, or DRO objectives that optimize for the worst-case log-loss or AUC across subgroups.
        Error bars indicate 95\% confidence intervals derived with the percentile bootstrap with 1,000 iterations.
	}
	\label{fig:main_text/performance/race_eth_sex_relative}
\end{figure*}

\begin{figure*}[!t]
	\centering
	\includegraphics[width=0.95\linewidth]{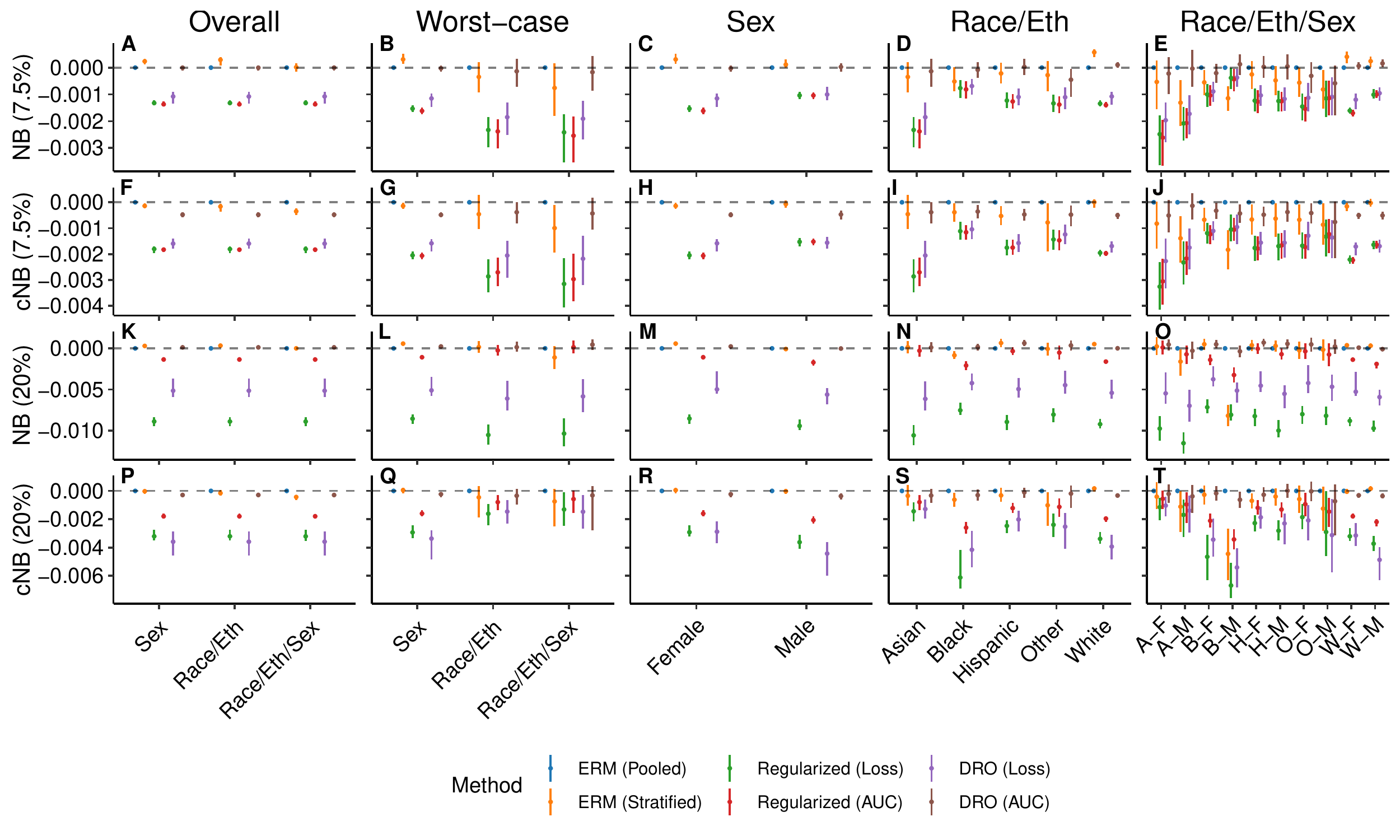}
	\caption{
    	The net benefit of models that estimate ten-year ASCVD risk, stratified by race, ethnicity, and sex, relative to the results attained by the application of unpenalized ERM to the overall population.
    	Results shown are the net benefit (NB) and calibrated net benefit (cNB), parameterized by the choice of a decision threshold of 7.5\% or 20\%, assessed in the overall population, on each subgroup, and in the worst-case over subgroups following the application of unconstrained pooled or stratified ERM, regularized objectives that penalize differences in the log-loss or AUC across subgroups, or DRO objectives that optimize for the worst-case log-loss or AUC across subgroups.
        Error bars indicate 95\% confidence intervals derived with the percentile bootstrap with 1,000 iterations.
	}
	\label{fig:main_text/net_benefit/race_eth_sex_relative}
\end{figure*}

%% Equalized odds experiment
% Performance grid
\begin{figure*}[!t]
	\centering
	\includegraphics[width=0.96\linewidth]{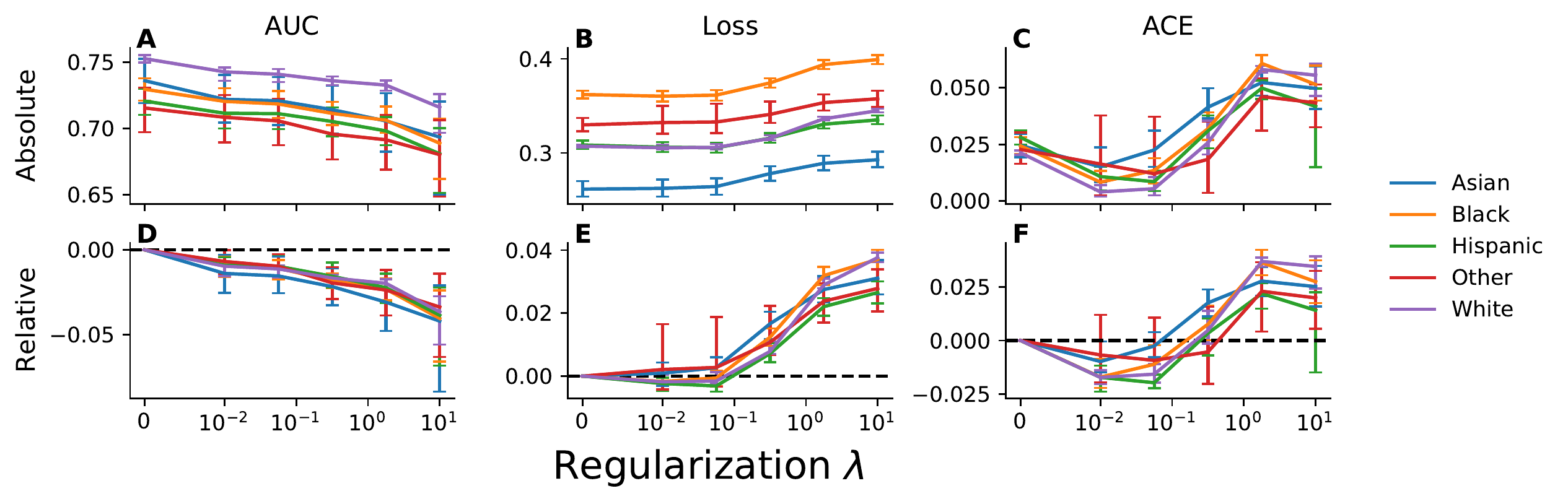}
	\caption{
	    Model performance evaluated across racial and ethnic subgroups for models trained with an objective that penalizes violation of equalized odds across intersectional subgroups defined on the basis of race, ethnicity, and sex using a MMD-based penalty.
	    Plotted, for each subgroup and value of the regularization parameter $\lambda$, is the area under the receiver operating characteristic curve (AUC), log-loss, and absolute calibration error (ACE).
	    Relative results are reported relative to those attained for unconstrained empirical risk minimization. 
	    Error bars indicate 95\% confidence intervals derived with the percentile bootstrap with 1,000 iterations.
	}
	\label{fig:main_text/eo_rr/race_eth/mmd/eo_performance_lambda}
\end{figure*}

% Performance-EO grid
\begin{figure*}[!t]
	\centering
	\includegraphics[width=0.95\linewidth]{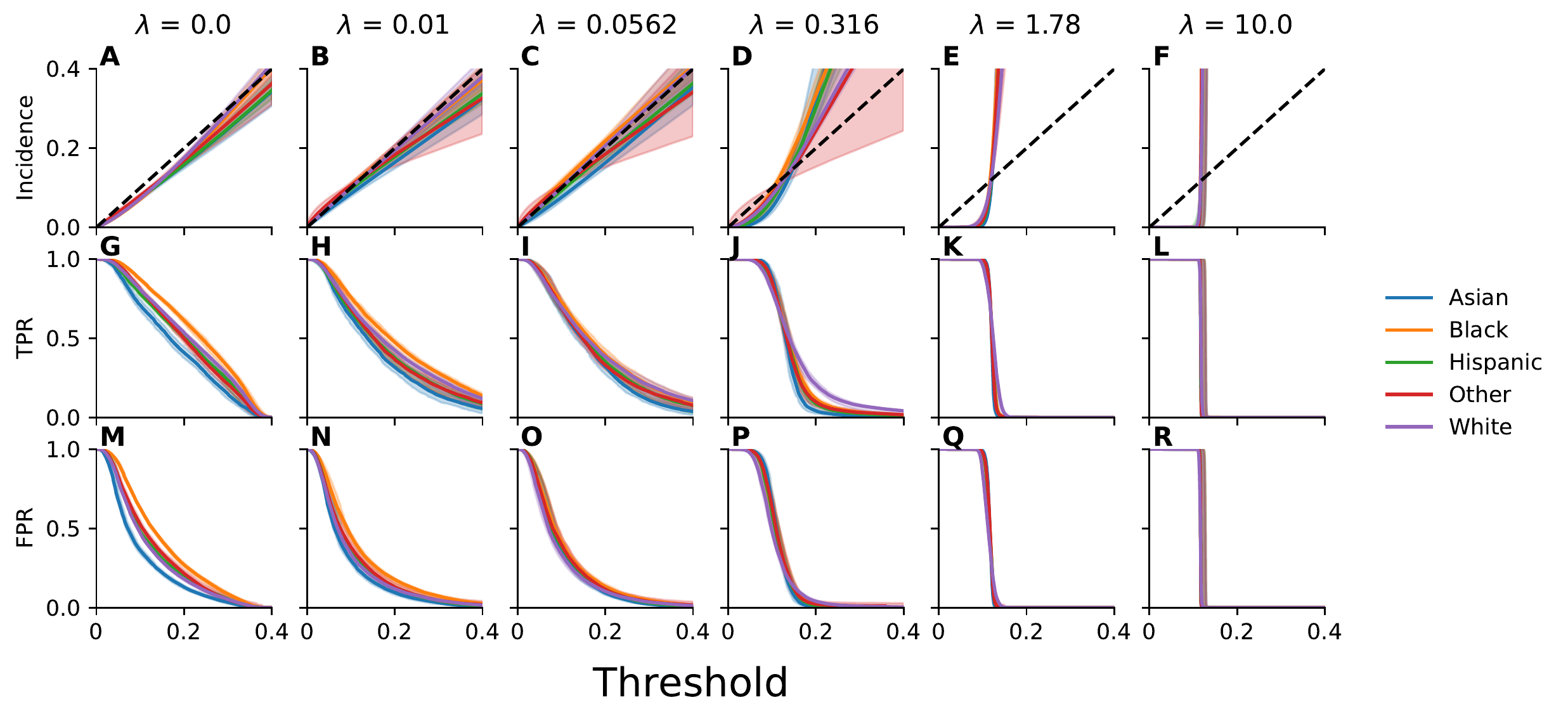}
	\caption{
	    Calibration curves, true positive rates, and false positive rates evaluated for a range of thresholds across racial and ethnic subgroups for models trained with an objective that penalizes violation of equalized odds across intersectional subgroups defined on the basis of race, ethnicity, and sex using a MMD-based penalty.
	    Plotted, for each subgroup and value of the regularization parameter $\lambda$, are the calibration curve (incidence), true positive rate (TPR), and false positive rate (FPR) as a function of the decision threshold.
	    Error bands indicate 95\% confidence intervals derived with the percentile bootstrap with 1,000 iterations.
	}
	\label{fig:main_text/eo_rr/race_eth/mmd/calibration_tpr_fpr}
\end{figure*}

% Net benefit metrics as a function of lambda
\begin{figure*}[!t]
	\centering
	\includegraphics[width=0.95\linewidth]{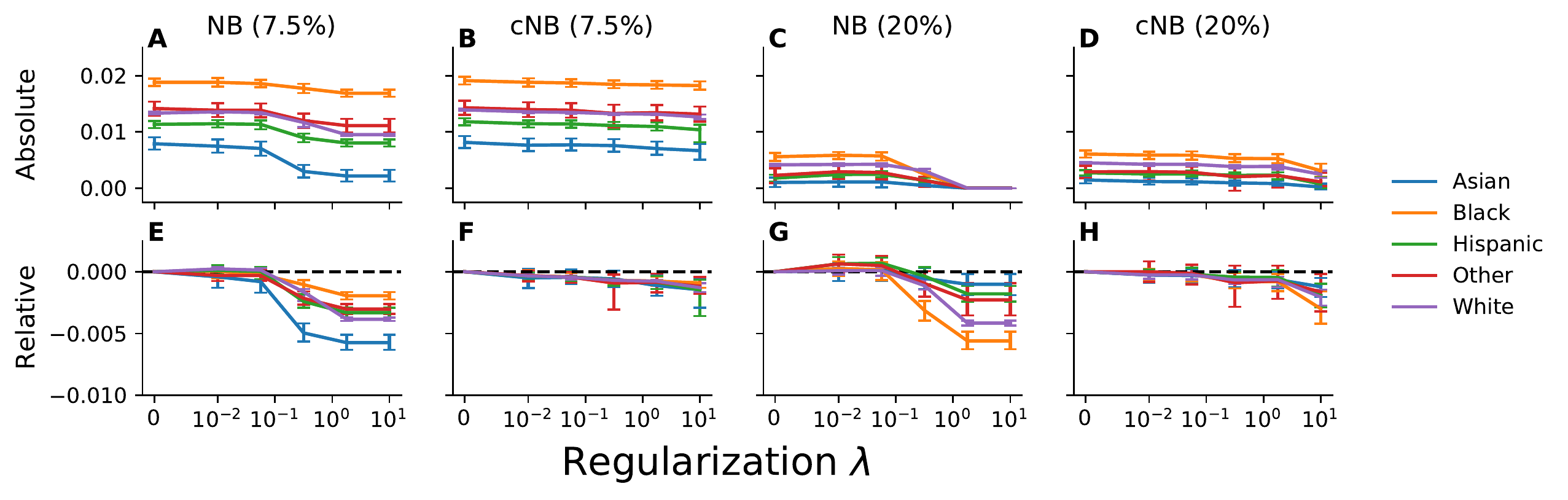}
	\caption{
	    The net benefit evaluated across racial and ethnic subgroups, parameterized by the choice of a decision threshold of 7.5\% or 20\%, for models trained with an objective that penalizes violation of equalized odds across intersectional subgroups defined on the basis of race, ethnicity, and sex using a MMD-based penalty.
	    Plotted, for each subgroup, is the net benefit (NB) and calibrated net benefit (cNB) as a function of the value of the regularization parameter $\lambda$.
	    Relative results are reported relative to those attained for unconstrained empirical risk minimization.
	    Error bars indicate 95\% confidence intervals derived with the percentile bootstrap with 1,000 iterations.
	}
	\label{fig:main_text/eo_rr/race_eth/mmd/eo_net_benefit_lambda}
\end{figure*}

% Equalized odds satisfaction
\begin{figure*}[!t]
	\centering
	\includegraphics[width=0.95\linewidth]{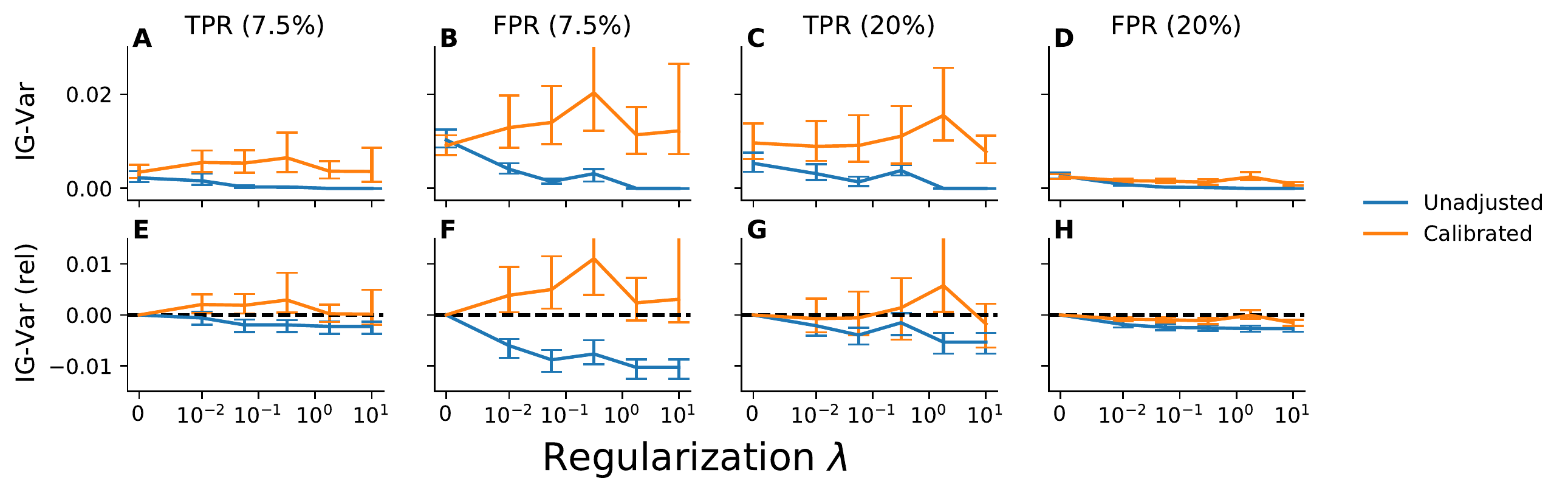}
	\caption{
        Satisfaction of equalized odds evaluated across racial and ethnic subgroups for models trained with an objective that penalizes violation of equalized odds across intersectional subgroups defined on the basis of race, ethnicity, and sex using a MMD-based penalty.
        Plotted is the intergroup variance (IG-Var) in the true positive and false positive rates at decision thresholds of 7.5\% and 20\%.
        Recalibrated results correspond to those attained for models for which the threshold has been adjusted to account for the observed miscalibration.
        Relative results are reported relative to those attained for unconstrained empirical risk minimization.
	    Error bars indicate 95\% confidence intervals derived with the percentile bootstrap with 1,000 iterations.
	}
	\label{fig:main_text/eo_rr/race_eth/mmd/eo_tpr_fpr_var_lambda}
\end{figure*}

% Decision curves at t=0.075
\begin{figure*}[!ht]
	\centering
	\includegraphics[width=0.95\linewidth]{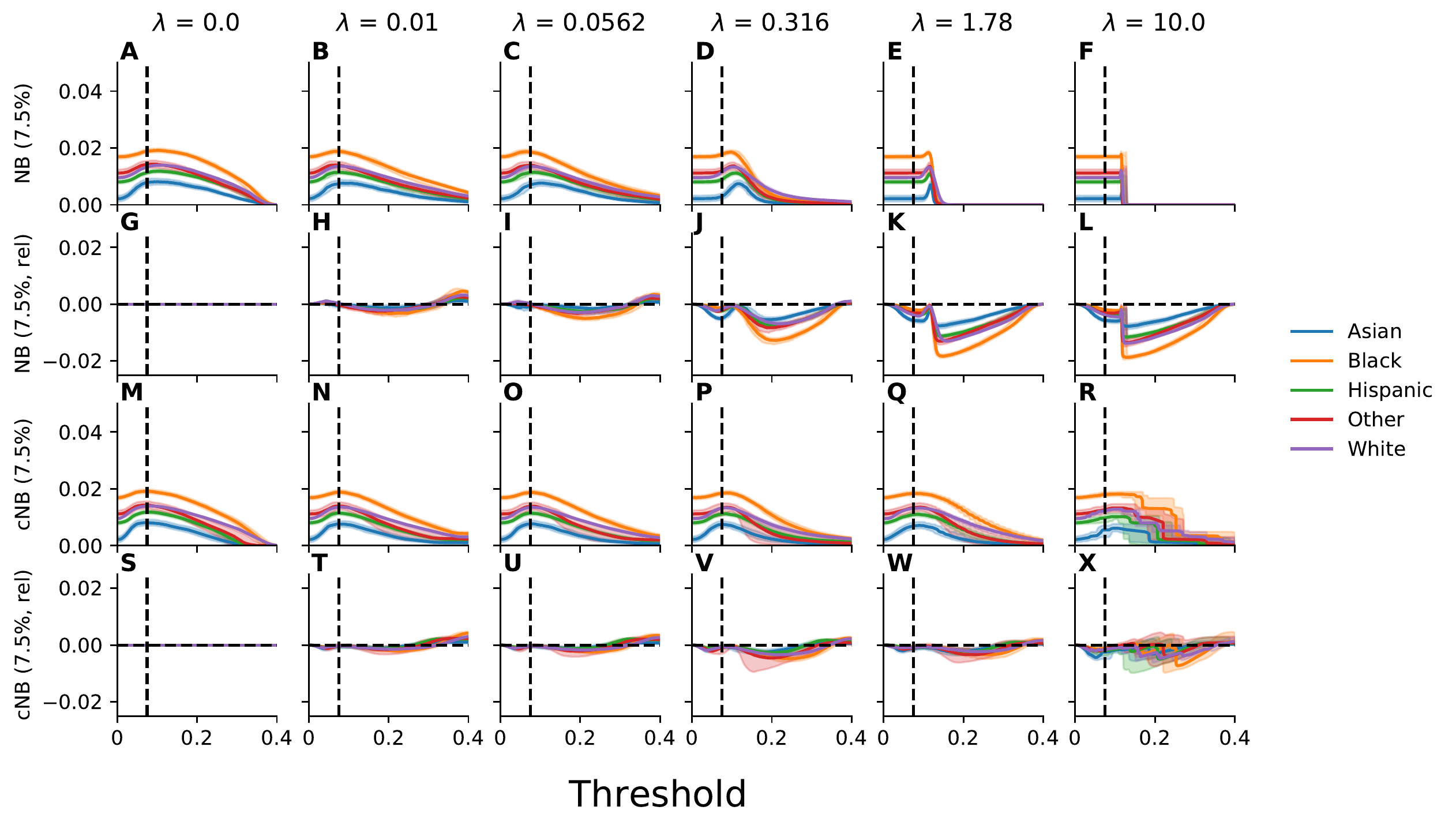}
	\caption{
	    The net benefit evaluated for a range of thresholds across racial and ethnic subgroups,  parameterized by the choice of a decision threshold of 7.5\%, for models trained with an objective that penalizes violation of equalized odds across intersectional subgroups defined on the basis of race, ethnicity, and sex using a MMD-based penalty.
	    Plotted, for each subgroup and value of the regularization parameter $\lambda$, is the net benefit (NB) and calibrated net benefit (cNB) as a function of the decision threshold.
	    Results reported relative to the results for unconstrained empirical risk minimization are indicated by ``rel''.
	    Error bars indicate 95\% confidence intervals derived with the percentile bootstrap with 1,000 iterations.
	}
	\label{fig:main_text/eo_rr/race_eth/mmd/decision_curves_075}
\end{figure*}

\subsection{Approaches to improve model performance over subgroups}
We conducted an experiment to assess whether approaches that penalize differences in AUC or log-loss across subgroups or optimize for the worst-case value of these metrics improve upon empirical risk minimization approaches in terms of the model performance and net benefit measures.
In the main text, we report the results assessed relative to those derived from unpenalized ERM applied to the entire population for subgroups defined in terms of race, ethnicity, and sex (Figure \ref{fig:main_text/performance/race_eth_sex_relative}), as well as for subgroups with ASCVD-promoting comorbidities (Supplementary Figure \ref{fig:supplement/performance/comorbidities_relative}).
Absolute performance estimates are reported in the supplementary material (Supplementary Figure \ref{fig:supplement/performance/race_eth_sex_absolute} and Supplementary Figure \ref{fig:supplement/performance/comorbidities_absolute}).

We find that the use of unconstrained empirical risk minimization using data from the entire population typically results in models with the greatest AUC for each subgroup, but stratified ERM procedures that train a separate model for each subgroup achieve an AUC that does not differ substantially in some cases, particularly for majority subgroups (Figure \ref{fig:main_text/performance/race_eth_sex_relative}D,E and Supplementary Figure \ref{fig:supplement/performance/comorbidities_relative}C,D,E,F).
The models trained with regularized fairness objectives or DRO and selected on the basis of the worst-case AUC or log-loss do not improve on the AUC assessed for each subgroup, and typically perform substantially worse, with the least extreme degradation observed for those models trained with the AUC-based DRO training objective (Figure \ref{fig:main_text/performance/race_eth_sex_relative}C,D,E and Supplementary Figure \ref{fig:supplement/performance/comorbidities_relative}C,D,E,F).
Despite the lack of improvement in AUC, we observe that subgroup-specific ERM and both regularized and DRO-based objectives that incorporate the AUC into their training objective often result in improved model calibration for some subgroups (\ref{fig:main_text/performance/race_eth_sex_relative}F,G,H,I,J and Supplementary Figure \ref{fig:supplement/performance/comorbidities_relative}G,I,J,K,L).
Similarly, subgroup-specific training does result in minor improvements in the log-loss for some subgroups relative to ERM applied to the entire population, but these results are typically observed only for larger subgroups when they are present (Figure \ref{fig:main_text/performance/race_eth_sex_relative}M,D,O and Supplementary Figure \ref{fig:supplement/performance/comorbidities_relative}O,R).

The implication of these effects can be understood holistically through an assessment of the net benefit of statin therapy initiated on the basis of the risk estimates.
Overall, no approach consistently confers more net benefit than unpenalized ERM applied to the entire population for each subgroup, when the net benefit is assessed for the benefit-harm tradeoffs corresponding to either of the thresholds of 7.5\% or 20\%, but subgroup-specific training and AUC-based DRO approaches do lead to minor improvements in some cases (Figure \ref{fig:main_text/net_benefit/race_eth_sex_relative}C,D,E,M,N,O and Supplementary Figure \ref{fig:supplement/net_benefit/comorbidities_relative}C,F,O,R).
However, we note that, for each subgroup, no approach improves on the calibrated net benefit, \textit{i.e.} the net benefit achieved following adjustment of the decision threshold to account for the observed miscalibration, relative to unpenalized ERM applied to the entire population (Figure \ref{fig:main_text/net_benefit/race_eth_sex_relative}H,I,J,R,S,T and Supplementary Figure \ref{fig:supplement/net_benefit/comorbidities_relative}I,J,K,L,U,V,W,X).
This indicates that for those cases where an alternative strategy results in an increase in the net benefit conferred relative to that which is achieved for the pooled ERM strategy, it is a consequence of the improvement in calibration at the threshold of interest.

\subsection{Regularized fairness objectives for equalized odds}
We further conducted an experiment to assess the implications of the use of a training objective that penalizes violation of equalized odds across intersectional subgroups defined by race, ethnicity, and sex. 
In the main text, we present the results corresponding to an MMD-based penalty evaluated over subgroups defined by race and ethnicity, but include in the supplementary material analogous results corresponding to evaluation over intersectional categories and for sex (Supplementary \Crefrange*{fig:supplement/eo_rr/race_eth_sex/mmd/eo_performance_lambda}{fig:supplement/eo_rr/sex/mmd/decision_curves}).
Furthermore, the supplementary material includes analogous results for experiments that penalize equalized odds at both of the thresholds of 7.5\% and 20\% using softplus relaxations of the true positive and false positive rates (Supplementary \Crefrange*{fig:supplement/eo_rr/race_eth/threshold_rate/eo_performance_lambda}{fig:supplement/eo_rr/sex/threshold_rate/decision_curves}).

We observe that as the strength of the penalty $\lambda$ increases, the AUC assessed for each subgroup monotonically decreases (Figure \ref{fig:main_text/eo_rr/race_eth/mmd/eo_performance_lambda}A,D).
With a minor degree of equalized-odds promoting regularization (\textit{i.e.} $\lambda=0.01, 0.0562$), calibration actually improves relative to the result for unpenalized ERM (Figure \ref{fig:main_text/eo_rr/race_eth/mmd/eo_performance_lambda}C,F) and there is little to no change in the log-loss for each subgroup despite the reduction in AUC (Figure \ref{fig:main_text/eo_rr/race_eth/mmd/eo_performance_lambda}B,E).
This is reflected in the calibration curves presented in Figure \ref{fig:main_text/eo_rr/race_eth/mmd/calibration_tpr_fpr}, where we observe modest miscalibration consistent with overestimation of risk for each subgroup for the unconstrained model (Figure \ref{fig:main_text/eo_rr/race_eth/mmd/calibration_tpr_fpr}A) with improvements in the calibration of the model with a minor degree of regularization (Figure \ref{fig:main_text/eo_rr/race_eth/mmd/calibration_tpr_fpr}B,C).
However, for large degrees of regularization (\textit{i.e.} $\lambda=1.78$ and $\lambda=10$), both the calibration and log-loss assessed for each subgroup deteriorates, although the reduction in AUC remains modest (Figure \ref{fig:main_text/eo_rr/race_eth/mmd/eo_performance_lambda}).
In this case, the variability in the risk estimates sharply decreases to concentrate around the incidence of the outcome for larger degrees of regularization, which is reflected in the shape of the calibration curve and error rates as a function of the threshold (Figure \ref{fig:main_text/eo_rr/race_eth/mmd/calibration_tpr_fpr}F,L,R), consistent with overestimation for patients with risk lower than the incidence and underestimation for patients with risk greater than the incidence.

For the unconstrained model, the true positive rates and false positive rates at each threshold are ranked across subgroups in accordance with the observed incidence for each subgroup, such that the Black population has the largest true positive rate and false positive rate while the Asian population has the lowest true positive rate and false positive rate (Figure \ref{fig:main_text/eo_rr/race_eth/mmd/calibration_tpr_fpr}G,H).
The regularized training objective is successful at enforcing the equalized odds constraint, in that the variability in false positive and true positives rates trends towards zero as the strength of the penalty increases (Figures \ref{fig:main_text/eo_rr/race_eth/mmd/calibration_tpr_fpr} and Figure
\ref{fig:main_text/eo_rr/race_eth/mmd/eo_tpr_fpr_var_lambda}).
 
For the benefit-harm tradeoff implied by the use of either a threshold of 7.5\% or 20\%, we observe clear reductions in net benefit for each subgroup for large values of $\lambda$ (Figure \ref{fig:main_text/eo_rr/race_eth/mmd/eo_net_benefit_lambda}A,C,E,G).
With minor amounts of regularization, we observe little to no reduction in net benefit parameterized by either a threshold of 7.5\% or 20\%, and the point estimates for 20\% even suggest a relative increase in net benefit compared to unpenalized ERM
(Figure \ref{fig:main_text/eo_rr/race_eth/mmd/eo_net_benefit_lambda}E,G).
However, for large degrees of regularization, we observe large reductions in net benefit relative to that which is attained from unpenalized ERM, but the magnitude of these differences are attenuated when the thresholds applied for each subgroup are adjusted to account for miscalibration (Figure \ref{fig:main_text/eo_rr/race_eth/mmd/eo_net_benefit_lambda}B,D,F,H).
We further observe that the calibrated net benefit for equalized odds penalized models does not improve on unpenalized ERM at any value of $\lambda$ (Figure \ref{fig:main_text/eo_rr/race_eth/mmd/eo_net_benefit_lambda}C,F,D,H).
Overall, the reduction in net benefit observed directly due to operating at a suboptimal decision threshold, as a result of miscalibration, is generally larger than the reduction in net benefit that results due to the reduction in the AUC of the model at larger values of $\lambda$.
Furthermore, we note that threshold adjustment to recover net benefit lost due to the miscalibration resulting from the use of the training objective that penalizes equalized odds violation does not preserve the satisfaction of the equalized odds fairness constraint, as the variability in error rates at the adjusted thresholds is observed to be similar to or more variable than that which results from unpenalized ERM (Figure \ref{fig:main_text/eo_rr/race_eth/mmd/eo_tpr_fpr_var_lambda}).

To gain further insight into these phenomena, we plot the net benefit for a range of decision thresholds, assuming that the benefit-harm tradeoff is fixed to one implied by the use of a threshold of 7.5\% (Figure \ref{fig:main_text/eo_rr/race_eth/mmd/decision_curves_075}).
In the supplementary material, we include analogous results for the threshold of 20\% (Supplementary Figure \ref{fig:supplement/eo_rr/race_eth/mmd/decision_curves_20})), as well as standard decisions curves defined such that the net benefit plotted for each point on the curve corresponds to the benefit-harm tradeoff implied by the corresponding threshold on the x-axis (Supplementary Figure \ref{fig:supplement/eo_rr/race_eth/mmd/decision_curves})).
As expected for the analysis corresponding to a threshold of 7.5\%, the calibrated net benefit is maximized for each subgroup at a threshold on the risk estimates corresponding to the point where the observed incidence of the outcome conditioned on the risk estimate is 7.5\% (Figure \ref{fig:main_text/eo_rr/race_eth/mmd/decision_curves_075}M,N,O,P,Q,R).
Furthermore, when the model overestimates risk at a threshold of 7.5\% due to miscalibration, such as was the case for the unpenalized ERM model and for the models trained with a large penalty on equalized odds violation, the threshold that maximizes the net benefit is one greater than 7.5\% (Figure \ref{fig:main_text/eo_rr/race_eth/mmd/decision_curves_075}A,D,E,F).
In these cases, adjusting the threshold on the penalized models to compensate for miscalibration recovers the majority of difference in net benefit relative to the model derived with unpenalized ERM.

\section{Discussion}
The results suggest that in settings where the observed model miscalibration may be adjusted for with subgroup-specific recalibration or via threshold-adjustment, no approach to learning an ASCVD risk estimator confers more net benefit for each subgroup than unpenalized ERM applied to the entire population.
This claim follows from the observation that no alternative approach resulted in greater \textit{calibrated} net benefit for any subgroup.
We find that the net benefit for each population is maximized for each subgroup at a threshold on the risk score that is consistent with the analysis presented in section \ref{sec:utility}.

In cases where we observe improvements in the unadjusted net benefit over ERM, or little to no change despite a reduction in AUC, the differences directly follow from improvements in the calibration of the model derived from the alternative approach.
We observe such effects for models trained with objectives that penalize equalized odds to a minor degree, those trained with stratified ERM procedures that train a separate model for each subgroup, as well as for regularized fairness objectives and DRO procedures that operate over the AUC assessed for each subgroup.
Taken together, these results indicate that models derived from unpenalized ERM should not necessarily be assumed to be well-calibrated in practice, further highlighting the importance of model development, selection, and post-processing strategies that aims to identify the best-fitting, well-calibrated model for each subgroup.

Algorithmic fairness assessments in healthcare based on the equalized odds criterion are likely to be misleading.
If the model is calibrated and fits well for each subgroup, differences in those error rates are expected when the observed outcome incidence differs \cite{Liu2019}.
Similarly, effort undertaken to minimize equalized odds violation is likely to introduce harm when it results in unrecognized miscalibration or reduction of model fit.
If the differences in outcome incidence reflect measurement error that differs systematically across subgroups \cite{Mitchell2019,obermeyer2019dissecting}, then violation of equalized odds may be present.
However, for such a case, we argue for conducting a calibration-based fairness assessment with respect to a proxy of the targeted unobserved outcome that is not subject to differential measurement error across subgroups, as in \citet{obermeyer2019dissecting}.
When those differences in outcome incidence, which may or may not be mismeasured or observable, are a result of population-level differences in disease burden across patient subgroups as a result of structural disparities \cite{Bailey2017,Bailey2020}, we argue that the appropriate response is to endeavor to understand both the cause of those disparities and the impact of potential interventions on the structural factors that perpetuate health disparities \cite{world2010conceptual,Kalluri2020,Goodman2018}.

While this work motivates the use of approaches that reason about algorithmic fairness in terms of calibration characteristics \cite{barda2021addressing,pmlr-v80-hebert-johnson18a}, such assessments are not comprehensive.
For instance, calibration-based assessments do not account for differences in benefit that arise due to differences in the discrimination performance nor in differences in unmodeled heterogeneity in the outcome across subgroups  \cite{Corbett-Davies:2017:ADM:3097983.3098095,corbett2018measure}.
The presence of measurement error can also mask consequential violation of sufficiency with respect to the targeted unobserved outcome that is not subject to measurement error \cite{Mitchell2019,obermeyer2019dissecting}.
Furthermore, when a predictive model is used for referral to a clinical service that cannot process more than a fixed number of cases due to resource constraints, \textit{e.g.} as in \citet{jung2021framework}, then it may not be practical to operate at the utility-maximizing threshold.
In that case, differences in the magnitude of the unrealized utility across subgroups are likely if the distribution of risk differs across subgroups, even if sufficiency holds and a consistent global threshold is applied across subgroups. 
Such a capacity constrained situation poses a set of ethical conflicts and trade-offs that should be navigated with participatory processes incorporating the preferences and attitudes of a diverse set of stakeholders \cite{Cagliero2021-hc}.

\section{Acknowledgements}
We thank the Stanford Center for Population Health Sciences Data Core and the Stanford Research Computing Center for supporting the data and computing infrastructure used in this work.
This work is supported by the National Heart, Lung, and Blood Institute R01 HL144555 and the Stanford Medicine Program for AI in Healthcare.
Any opinions, findings, and conclusions or recommendations expressed in this material are those of the authors and do not necessarily reflect the views of the funding bodies.

\section{Data availability}
Individuals wishing to access the data used in this work may sign a data use agreement with Stanford and Optum to access the data for replication or confirmatory studies on the Stanford Secure Data Ecosystem.

\section{Code availability}
We make all code available at \url{https://github.com/som-shahlab/net_benefit_ascvd}.

\section{Competing interests statement}
The authors declare no competing interests.

\bibliographystyle{unsrtnat}
\bibliography{references}

\clearpage
\appendix
\counterwithin{figure}{section}
\counterwithin{table}{section}
\renewcommand*\thetable{\Alph{section}\arabic{table}}
\renewcommand*\thefigure{\Alph{section}\arabic{figure}}
\renewcommand{\figurename}{Supplementary Figure}

\section*{Supplementary material}

\section{Supplementary methods}
\subsection{Simulation study} \label{sec:simulation}
\begin{figure}[!thbp]
    \centering
    \includegraphics[width=0.95\linewidth]{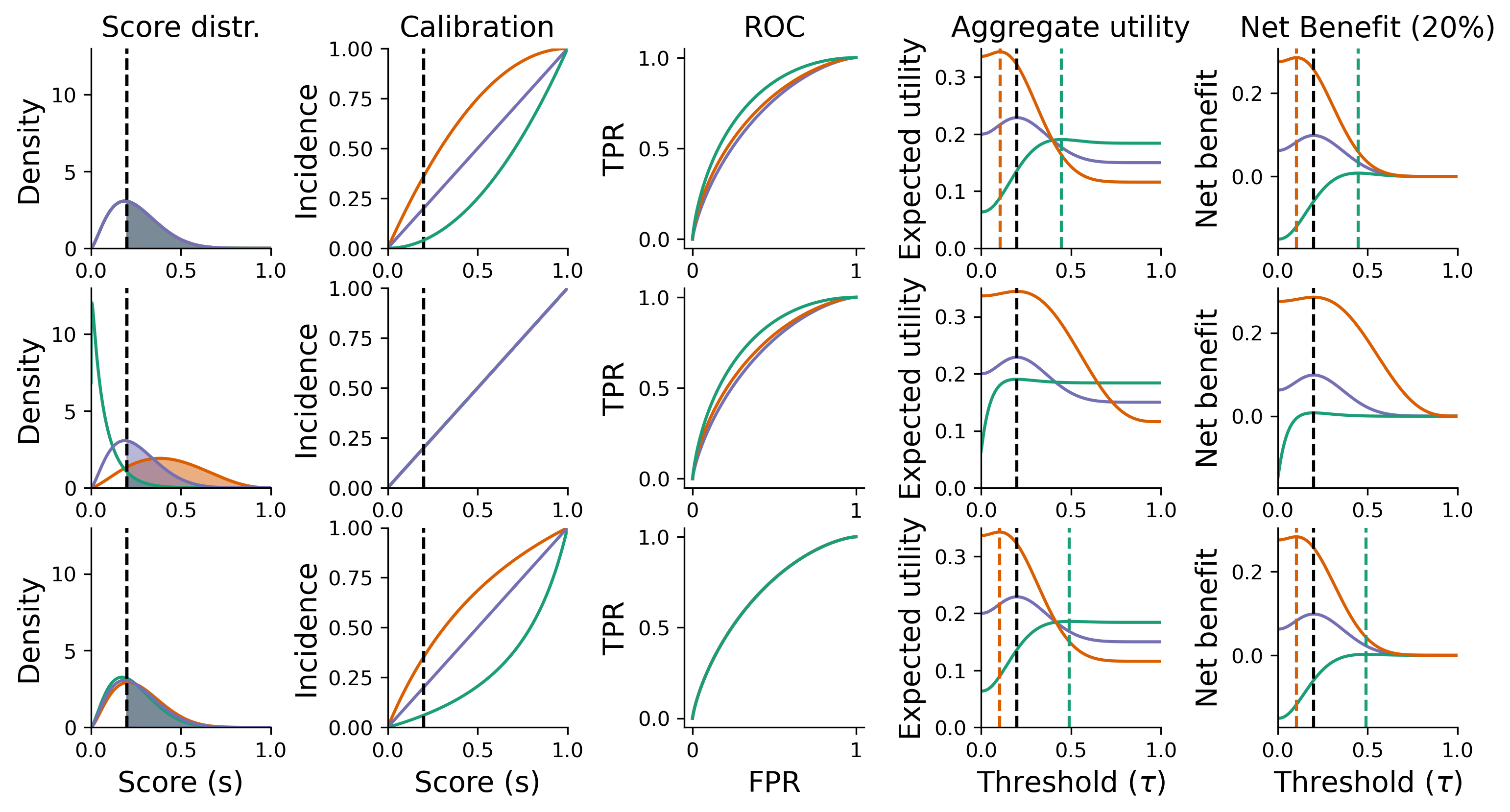}
    \caption{
        The relationship between score distributions, calibration, receiver operating characteristic curves, aggregate expected utility, and net benefit in simulation when demographic parity (first row), group calibration (second row), and equalized odds (third row) are satisfied. Dashed lines indicate optimal decision thresholds.
    }
    \label{fig:simulation}
\end{figure}

We conduct a simulation study in order to help clarify the relationship between the key concepts considered in this work. 
We consider evaluation of a predictive model of a binary outcome without censoring using the notion of utility and net benefit discussed in section \ref{sec:utility}.
For evaluation in all settings, we evaluate aggregate expected utility using the fixed-cost utility function with $u_{\textrm{TP}}=0.8$, $u_{\textrm{TN}}=0.2$, and $u_{\textrm{FP}}=u_{\textrm{FN}}=0$, corresponding to an optimal threshold of 0.2, as well as the net benefit parameterized by $\tau^*=0.2$.

We evaluate over three subgroups in three different settings (Supplementary Figure \ref{fig:simulation}).
In the first setting, we assume the model satisfies demographic parity and the score distribution is given by $S \sim \textrm{Beta}(2.5, 7.5)$, corresponding to an outcome incidence of 25\%.
We assume that the model is perfectly calibrated for one subgroup (purple) and that risk is systematically underestimated (orange) or overestimated (green) for the other two subgroups.
We encode the miscalibration by setting $c(s)=-(s-1)^2+1$ for underestimation and $c(s)=2s+(s-1)^2-1$ for overestimation.
For the second setting, we consider a hypothetical recalibration procedure for each subgroup that constructs a new set of scores by setting the value of the score to be that of the subgroup calibration curve, propagating the changes to the score distributions using change of variables for probability density functions.
For the third context, we consider the effect of enforcing an equalized odds constraint by setting the conditional distributions $P(S\mid Y)$ for each subgroup to be equal to that of the subgroup for which the model was perfectly calibrated in the first setting, where demographic parity was satisfied, as the ROC curve for that subgroup defines the convex hull of the ROC curves across all subgroups.
In this case, the resulting changes to the score distributions and calibration curves are computed using simple conditional probability rules.

The results of the simulation study are consistent with the presentation of section \ref{sec:implications}.
In particular, we note that when the model is calibrated for each subgroup, the utility and net benefit maximizing decision threshold is the same for each subgroup.
However, when demographic parity or equalized odds are satisfied and sufficiency is violated, the optimal threshold differs from 0.2 for the subgroups for which the model is miscalibrated.
We note that satisfying equalized odds is similar to that of demographic parity in terms of the effect on the score distributions, aggregate utility, and net benefit.

\subsection{Assessing net benefit in terms of risk reduction} \label{sec:supp:net_benefit_risk_reduction}
This section expands upon the presentation in section \ref{sec:net_benefit_risk_reduction} and includes full derivations for equations (\ref{eq:net_benefit_risk_reduction}) and (\ref{eq:calibrated_net_benefit_risk_reduction}).
We present a conceptual model that reasons about the utility of an intervention allocated on the basis of a decision rule applied to a predictive model in terms of a downstream reduction in the risk of the predicted outcome as a result of the intervention.
We show that the use of this conceptual model results in similar conclusions as the fixed expected cost/utility setting.

We define $u_1^y$ be the utility associated with the presence of the outcome $Y$ and $u_0^y$ be the utility associated with its absence. 
The probability of the outcome in the absence of the intervention is given by $p_y^0(s)=c(s)$ where $c(s)$ is the calibration curve. 
The probability of the outcome in the presence of intervention is given by $p_y^1(s)$, where the precise form of $p_y^1(s)$ is governed by the effectiveness of the intervention.
We further assume that there is some harm $Z$, representing all costs, harms, or side effects, that occurs with probability $p_z^1(s)$ and utility $u_1^z$ following the intervention and with $p_z^0(s)$ and utility $u_0^z$ in the absence of the intervention.

This formulation implies the conditional utilities
\begin{equation} \label{eq:conditional_utility_neg_rr_expanded}
    U_{\mathrm{cond}}^0(s) = p_y^0(s) \big(u_1^y - u_0^y\big) + u_0^y + p_z^0(s) \big(u_1^z - u_0^z\big) + u_0^z,
\end{equation}
\begin{equation} \label{eq:conditional_utility_pos_rr_expanded}
    U_{\mathrm{cond}}^1(s) = p_y^1(s) \big(u_1^y - u_0^y\big) + u_0^y + p_z^1(s) \big(u_1^z - u_0^z\big) + u_0^z,
\end{equation}
and
\begin{equation} \label{eq:conditional_utility_rr_expanded}
    U_{\mathrm{cond}}(s) = \big(u_0^y - u_1^y\big)\big(p_y^0(s) - p_y^1(s)\big) + \big(u_0^z - u_1^z\big)\big(p_z^0(s) - p_z^1(s)\big).
\end{equation}

Setting equation (\ref{eq:conditional_utility_rr_expanded}) to zero shows that the value of the optimal threshold $\tau_y^*$ is governed by the following relationship
\begin{equation} \label{eq:optimal_threshold_rr_expanded}
    \big(u_0^y - u_1^y\big)\big(p_y^0(\tau_y^*) - p_y^1(\tau_y^*)\big) =  \big(u_0^z - u_1^z\big)\big(p_z^1(\tau_y^*) - p_z^0(\tau_y^*)\big) \rightarrow \frac{p_y^0(\tau_y^*) - p_y^1(\tau_y^*)}{p_z^1(\tau_y^*) - p_z^0(\tau_y^*)} = \frac{u_0^z - u_1^z}{u_0^y - u_1^y},
\end{equation}
when $U_{\mathrm{cond}}(s)$ is monotonically increasing in $s$.
To interpret this expression, consider that $\mathrm{ARR}(s) = p_y^0(s) - p_y^1(s)$ is the absolute reduction in risk as a result of the intervention, $p_z^0(s) - p_z^1(s)$ indicates a corresponding increase in the risk of harm, and $u_0^y - u_1^y$ and $u_0^z - u_1^z$ indicate the utilities associated with avoiding $Y$ and $Z$, respectively.
It follows that the optimal threshold is the one where the benefits of the intervention are balanced against its harms.

To simplify the model, we now assume that $p_z^1(s) - p_z^0(s)$ do not depend on the risk score, indicating that the expected harm $k_{\mathrm{harm}}=\big(u_0^z - u_1^z\big)\big(p_z^1 - p_z^0\big)$ is a constant.

With this assumption, the conditional utility may be represented as
\begin{equation} \label{eq:conditional_utility_rr}
    U_{\mathrm{cond}}(s) = \big(u_0^y - u_1^y\big)\big(p_y^0(s) - p_y^1(s)\big) - k_{\mathrm{harm}}.
\end{equation}

Setting equation (\ref{eq:conditional_utility_rr}) to zero shows that the value of the optimal threshold $\tau_y^*$ is governed by the following relationship, consistent with \citet{vickers2007method}:
\begin{equation} \label{eq:optimal_threshold_rr}
    p_y^0(\tau_y^*) - p_y^1(\tau_y^*) = \frac{k_{\mathrm{harm}}}{u_0^y - u_1^y}.
\end{equation}
This expression relates the absolute risk reduction $\mathrm{ARR}(s) = p_y^0(s) - p_y^1(s)$ evaluated at the optimal threshold $\tau_y$ to both the expected harm of intervention $k_{\mathrm{harm}}$ and the utility of avoiding the outcome $u_0^y - u_1^y$.
It follows that the optimal threshold $\tau_y^*$ is given by 
\begin{equation}
    \tau_y^* = \mathrm{ARR}^{-1}\Big(\frac{k_{\mathrm{harm}}}{u_0^y - u_1^y}\Big).
\end{equation}
Furthermore, the aggregate utility over the population when treating at a threshold $\tau_y$ can be derived as 
\begin{align} \label{eq:aggregate_utility_rr}
    \begin{split}
        & U_{\mathrm{agg}}(\tau_y) = \Big(u_1^y-u_0^y\Big)\Big(\int_0^{\tau_y} p_y^0(s) P(s) ds + \int_{\tau_y}^1 p_y^1(s) P(s) ds\Big) - k_{\mathrm{harm}} + p_z^0(u_1^z-u_0^z) + u_0^y + u_0^z \\
        & = \Big(u_1^y-u_0^y\Big)\Big(\E[p_y^0(s) \mid S < \tau_y]P(S < \tau_y) + \E[p_y^1(s) \mid S \geq \tau_y]P(S \geq \tau_y)\Big) \ldots \\ & \hspace{20mm} -k_{\mathrm{harm}} + p_z^0(u_1^z-u_0^z) + u_0^y + u_0^z.
    \end{split}
\end{align}

To construct a net benefit measure that represents the aggregate utility given that $\tau_y$ is the optimal threshold, we divide equation (\ref{eq:aggregate_utility_rr}) by $u_0^y-u_1^y$, perform a substitution following equation (\ref{eq:optimal_threshold_rr}), and define a constant $k$ such that the net benefit of the treat-none strategy is zero:
\begin{equation}
        \mathrm{NB}(\tau_y; \tau_y^*) = -\E[p_y^0(s) \mid S < \tau_y]P(S < \tau_y) -\E[p_y^1(s) \mid S \geq \tau_y]P(S \geq \tau_y) - \mathrm{ARR}(\tau_y^*) P(S \geq \tau_y) + k.
\end{equation}
From this expression, it follows that the appropriate value of $k$ is given by $P(Y=1) = \E[p_y^0(s) \mid S < 1]P(S < 1)$, giving the following expression for the net benefit:
\begin{align} \label{eq:net_benefit_rr}
    \begin{split}
        \mathrm{NB}(\tau_y; \tau_y^*) & = -\E[p_y^0(s) \mid S < \tau_y]P(S < \tau_y) -\E[p_y^1(s) \mid S \geq \tau_y]P(S \geq \tau_y) \ldots \\
        & \hspace{15mm} - \mathrm{ARR}(\tau_y^*) P(S \geq \tau_y) + P(Y=1).
    \end{split}
\end{align}
The expression for the calibrated variant of the net benefit is given by:
\begin{align} \label{eq:net_benefit_rr_recalib}
    \begin{split}
        \mathrm{cNB}(\tau_y; \tau_y^*) & = -\E[p_y^0(s) \mid S < c^{-1}(\tau_y)]P(S < c^{-1}(\tau_y)) -\E[p_y^1(s) \mid S \geq c^{-1}(\tau_y)]P(S \geq c^{-1}(\tau_y)) \ldots \\
        & \hspace{15mm} - \mathrm{ARR}(\tau_y^*) P(S \geq c^{-1}(\tau_y)) + P(Y=1).
    \end{split}
\end{align}
This formulation differs from that of \citet{vickers2007method} in that that work defines the treat-all strategy as having a net benefit of zero whereas we do so for the treat-none strategy in order to maintain consistency with the net benefit defined for the fixed-cost utility function.

\subsubsection{Constant relative risk reduction}
Given only observational data that corresponds to an untreated population, it is necessary to provide assumptions on the form of $p_y^1(s)$ in order to assess the net benefit of a model using a utility function defined in terms of the risk reduction induced by the intervention.
A simple choice for the relationship between $p_y^0(s)$ and $p_y^1(s)$ is one where the intervention reduces the risk of the outcome by a constant multiplicative factor $r \in (0, 1)$, such that $p_y^1(s)=(1-r)p_y^0(s)$.
Given this assumption, $p_y^0(s) = c(s)$, $p_y^1(s) = (1-r)c(s)$, and $\mathrm{ARR}(s) = r c(s)$.

With these assumptions, it follows that $U_{\mathrm{cond}}(s)$ is a linear transformation of the calibration curve, just as was the case for the fixed-cost utility function:
\begin{equation} \label{eq:conditional_utility_rr_const}
    U_{\mathrm{cond}}(s) = \big(u_0^y - u_1^y\big) rc(s) - k_{\mathrm{harm}}.
\end{equation}
Furthermore, the optimal threshold is given by
\begin{equation}
    \tau_y^* = c^{-1}\Big(\frac{k_{\mathrm{harm}}}{r(u_0^y-u_1^y)}\Big),
\end{equation}
which can be simplified to 
\begin{equation}
    \tau_y^* = \frac{k_{\mathrm{harm}}}{r(u_0^y-u_1^y)}
\end{equation}
when the model is assumed to be calibrated.

Making the appropriate substitutions into equation (\ref{eq:net_benefit_rr}), and noting that $\E[c(s) \mid S < \tau_y] = \E[Y \mid S < \tau_y]$ and $\E[c(s) \mid S \geq \tau_y] = \E[Y \mid S \geq \tau_y]$, gives an expression for the net benefit:
\begin{align}
    \begin{split}
        \mathrm{NB}(\tau_y; \tau_y^*) & = -\E[Y \mid S < \tau_y]P(S < \tau_y) - 
        P(S \geq \tau_y)\Big((1-r)\E[Y \mid S \geq \tau_y] + \mathrm{ARR}(\tau_y^*)\Big) + P(Y=1) \\
        & = -(1-\mathrm{NPV}(\tau_y))P(S < \tau_y) - 
        P(S \geq \tau_y)\Big((1-r)\mathrm{PPV}(\tau_y) + r \tau_y^* \Big) + P(Y=1) 
    \end{split}
\end{align}
where $\mathrm{NPV}(\tau_y)$ and $\mathrm{PPV}(\tau_y)$ designate the negative and positive predictive values that result from operating at the decision threshold $\tau_y$. Note that this matches equation (\ref{eq:net_benefit_risk_reduction}) and that the calibrated variant in equation (\ref{eq:calibrated_net_benefit_risk_reduction}) is analogous.

\subsection{Regularized training objectives for fairness} \label{sec:supp:training_objectives}
Here, we present the regularized training objective that allows for the flexible specification of penalties on the violation of fairness criteria.
To begin, we consider the MMD-based penalty for equalized odds presented in equation (\ref{eq:reg_objective_eo}). 
The MMD uses the distance between the mean embedding of samples from two distributions in a kernel space to define a statistic that takes a value of zero in a population setting if and only if two distributions are the same \cite{gretton2012kernel}.
To construct a regularizer, we use an empirical estimate of the squared population MMD \cite{gretton2012kernel}
\begin{align} \label{eq:mmd_kernel}
    \begin{split}
    \hat{D}_{\mathrm{MMD}}(\mathcal{D}_0 \mid \mid \mathcal{D}_1) & = \E_{(z, z')\sim \mathcal{D}_0,\mathcal{D}_0} [k\big(z, z'\big)] - \\
        & 2 \E_{(z, z') \sim \mathcal{D}_0, \mathcal{D}_1}[k\big(z, z'\big)] + \\
        & \E_{(z, z')\sim \mathcal{D}_1, \mathcal{D}_1} [k\big(z, z'\big)],
    \end{split}
\end{align}
where $k(z, z')= \exp(-\gamma\lVert z - z'\rVert)$ is the Gaussian Radial Basis Function kernel defined for a positive scalar hyperparameter $\gamma$, and $\E_{(z, z')\sim \mathcal{D}_0, \mathcal{D}_1}$ indicates an empirical mean following sampling a pair of data ($z, z'$) from $\mathcal{D}_0$ and $\mathcal{D}_1$, respectively. In our experiments, we set $\gamma=1$.
As noted in \citet{gretton2012kernel}, this statistic is non-negative, but has a small upward bias. 

To account for censoring, we use a weighted extension of the maximum mean discrepancy, as a modification to each of the expectations over the pairwise evaluation of the kernel function (equation \ref{eq:mmd_kernel}).
As an example, the term $\E_{(z, z') \sim \mathcal{D}_0, \mathcal{D}_1}[k\big(z, z'\big)]$ can be replaced with $\sum_{z_i, z_j \in \{\mathcal{D}_0, \mathcal{D}_1\}} w_{ij} k(z_i, z_j)$ for weights defined as 
\begin{equation} \label{eq:weighted_mmd_kernel}
    w_{ij} = 
    \frac{\delta_i^y}{G(u_i^y, x_i)} 
    \frac{\delta_j^y}{G(u_j^y, x_j)} 
    \Big(
        \sum_{z_i \in \mathcal{D}_0}
        \sum_{z_j \in \mathcal{D}_1}
            \frac{\delta_i^y}{G(u_i^y, x_i)} 
            \frac{\delta_j^y}{G(u_j^y, x_j)}
    \Big)^{-1}.
\end{equation}
To define the full MMD, each of the three expectations in equation (\ref{eq:mmd_kernel}) are replaced with weighted variants analogous to equation (\ref{eq:weighted_mmd_kernel}).

Now, we consider the operationalization of equation (\ref{eq:reg_objective_metric_parity}) to penalize differences in model performance metrics. 
Recall that equation (\ref{eq:reg_objective_metric_parity}) is well-suited to penalties that rely on smooth and differentiable $g_j$.
Unfortunately, naively plugging-in threshold-based metrics, including the classification rate, true positive rate, false positive rate, PPV, as well as those defined as ranking performance, including the AUC, does not produce a practical regularized objective due to the presence of the indicator function embedded in the definition of each of those metrics.
As an example, consider that the classification rate $\E[\mathbbm{1}[f_{\theta}(X) > \tau_y]]$ can be represented as $\E[\mathbbm{1}[f_{\theta}(X) - \tau_y > 0]]$ or $\E[h_{\mathrm{step}}\big(f_{\theta}(X) - \tau_y\big)]$ when $h_{\mathrm{step}}$ is the step-indicator function $h_{\mathrm{step}}(z)=\mathbbm{1}[z>0]$.
The shape of this function is such that it does not provide a useful signal for stochastic gradient descent, given that its derivative is zero everywhere that its derivative is defined.

\begin{figure}[!t]
    \centering
    \includegraphics[width=0.5\linewidth]{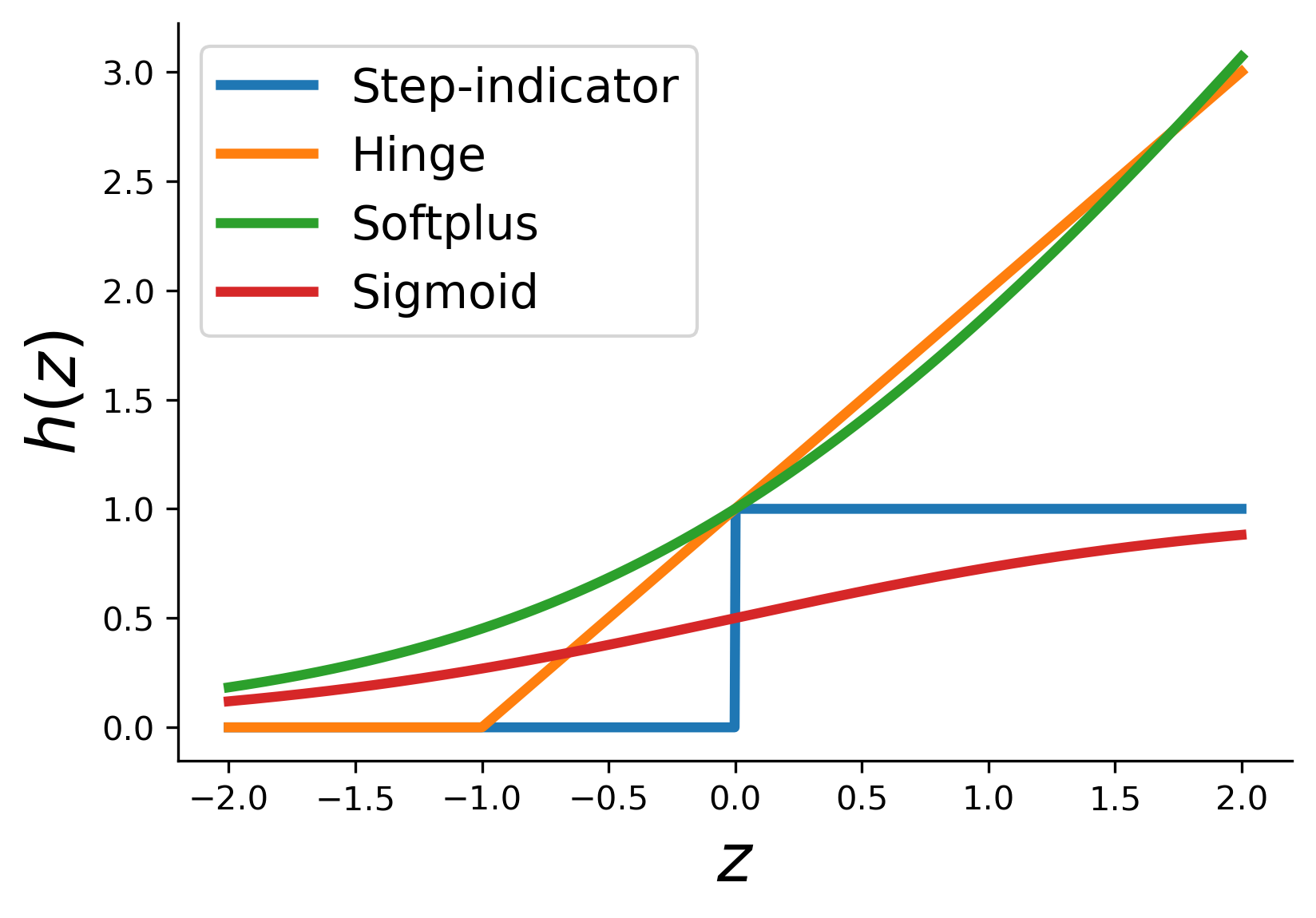}
    \caption{
        Surrogates to the indicator function.
    }
    \label{fig:surrogates}
\end{figure}

One approach to addressing this issue is to use a smooth and differentiable surrogate \cite{cotter2019optimization,eban2017scalable} to the step-indicator that either upper bounds or approximates it. 
A visual depiction of several options is provided in Figure \ref{fig:surrogates}.
The use of either the hinge, $h_{\mathrm{hinge}}(z) = \max(0, 1 + z)$), or the scaled softplus, $h_{\mathrm{softplus}}(z) = \log(1 + \exp(z))/ \log(2)$, provides a smooth and differentiable upper bound to the indicator. 
Because $h_{\mathrm{step}}(z) \leq h_{\mathrm{hinge}}(z)$ and $h_{\mathrm{step}}(z) \leq h_{\mathrm{softplus}}(z)$, a metric $g$ defined as a sum over evaluations of $h_{\mathrm{step}}(z)$ can be upper bounded by a metric $\hat{g}$ defined as a sum over evaluations of $h_{\mathrm{hinge}}(z)$ or $h_{\mathrm{softplus}}(z)$.
Furthermore, the use of the sigmoid function, $h_{\mathrm{sigmoid}}(z) = \frac{1}{(1 + \exp(-z)}$ does not directly bound the indicator function, but rather provides a smooth approximation to it (Figure \ref{fig:surrogates}), that can be similarly incorporated into a relaxed, approximate metric $\hat{g}$. 

Given a relaxed metric $\hat{g}$, the corresponding training objective is given by
\begin{equation} \label{eq:reg_objective_metric_parity_approx}
    \min_{\theta \in \Theta} \sum_{i=1}^N w_i \ell(y, f_{\theta}(x)) + \lambda \sum_{j=1}^{J} \sum_{A_k \in \mathcal{A}} \Big( \hat{g}_j(f_{\theta}, \mathcal{D}_{A_k}) - \hat{g}_j(f_{\theta}, \mathcal{D}) \Big)^2.
\end{equation}
With this relaxed objective, it is straightforward to penalize differences in threshold-based performance metrics, such as the true positive rate and false positive rates, or to penalize differences in AUC measures.
The true positive and false positive rates can each be written as $\sum_{i=1}^N w_i h(f_{\theta}(x_i) - \tau_y)$ for the following weights \cite{uno2007evaluating}:
\begin{equation} \label{eq:tpr_relaxed}
    w_i = \frac{\mathbbm{1}[y_i=1]\delta_i^y}{G(u_i^y, x_i)} \Big(\sum_{i=1}^N\frac{\mathbbm{1}[y_i=1]\delta_i^y}{G(u_i^y, x_i)}\Big)^{-1}
\end{equation}
for the true positive rate, and 
\begin{equation} \label{eq:fpr_relaxed}
    w_i = \frac{\mathbbm{1}[y_i=0]\delta_i^y}{G(u_i^y, x_i)} \Big(\sum_{i=1}^N\frac{\mathbbm{1}[y_i=0]\delta_i^y}{G(u_i^y, x_i)}\Big)^{-1}
\end{equation}
for the false positive rate.
The corresponding censoring-adjusted definition of the AUC \cite{blanche2013review} that incorporates IPCW is given by $\sum_{i=1}^{N} \sum_{j=1}^{N} w_{ij} h(f_{\theta}(x_i) - f_{\theta}(x_j))$ for
\begin{equation} \label{eq:auc_ipcw}
    w_{ij} = 
    \frac{\delta_i^y \mathbbm{1}[y_i=1]}{G(u_i^y, x_i)} 
    \frac{\delta_j^y \mathbbm{1}[y_j=0]}{G(u_j^y, x_j)} 
    \Big(
        \sum_{i=1}^{N}
        \sum_{j=1}^{N}
            \frac{\delta_i^y \mathbbm{1}[y_i=1]}{G(u_i^y, x_i)} 
            \frac{\delta_j^y \mathbbm{1}[y_j=0]}{G(u_j^y, x_j)}
    \Big)^{-1}.
\end{equation}

\clearpage
\section{Supplementary tables}

\begin{table}[!th]
\centering
\caption{
Code and concept identifiers used to construct the cohort. 
Parentheses indicate the source vocabulary for the listed identifiers. We use the International Classification of Diseases version 9 (ICD-9), the Anatomical Therapeutic Chemical Classification System (ATC), Logical Observation Identifiers Names and Codes (LOINC), and OMOP CDM concept identifiers.
Asterisks indicate the union of all possible suffixes and brackets indicate a range of included suffixes.
For identifiers that are not OMOP CDM concept identifiers, we map the listed identifiers to standard OMOP CDM concepts using the mappings provided by the OMOP CDM vocabulary.
Each set of OMOP CDM concepts used in the cohort definition is defined by the union of the mapped standard OMOP CDM concepts and their descendants in the OMOP CDM vocabulary followed by the exclusion of any excluded concepts and their descendants from the set.
}
\label{tab:ch5:cohort_concepts}
\begin{tabular}{ll}
\toprule
Concept & Code or concept identifiers \\
\midrule
Stroke (ICD-9) & \makecell[lt]{ 430*, 431*, 432*, 433* (except 433.*0), \\ 434* (except 434.*0), 436*} \\
Myocardial Infarction (ICD-9) & 410* \\
Coronary Heart Disease (ICD-9) & 411*, 413*, 414* \\
Cardiovascular Disease (ICD-9) & \makecell[lt]{410*, 411*, 413*, 414*, 430*, \\ 431*, 432*, 433*, 434*, 436*, 427.31, 428*} \\
Statin (ATC) & C10AA0[1-8] \\
Type 1 diabetes (OMOP) & 201254, 40484648, 201254, 435216 \\
Gestational diabetes (OMOP) & 4058243 \\
Type 2 diabetes (OMOP) & \makecell[lt]{443238, 201820, 442793 \\ (exclude all type 1 and gestational concepts)} \\
Chronic kidney disease (OMOP) & \makecell[lt]{192279, 192359, 193253, 194385, 195314, \\ 201313, 261071, 4103224, 4263367, 46271022 \\ (exclude 195014, 195289, 195737, 197320, \\ 197930, 4066005, 37116834, 43530912, 45769152)} \\
Rheumatoid Arthritis (OMOP) & 80809 \\
Low-density lipoprotein cholesterol (LOINC) & 18262-6, 13457-7, 2089-1 \\
\bottomrule
\end{tabular}
\end{table}

\clearpage
\section{Supplementary figures}

\begin{figure*}[!th]
    \centering
    \includegraphics[width=0.75\linewidth]{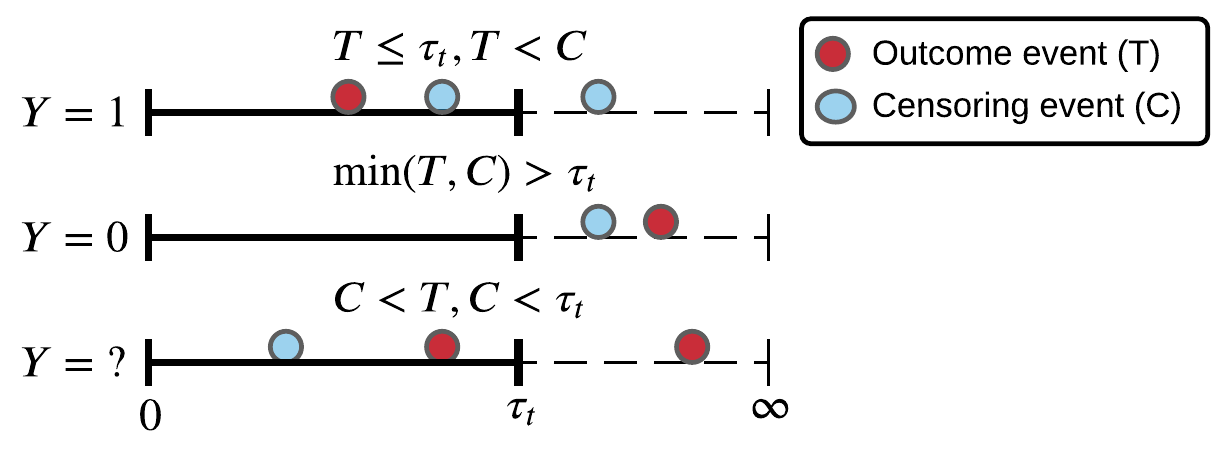}
    \caption{
        The effect of censoring on the observation of binary outcomes
    }
    \label{fig:censoring}
\end{figure*}

\begin{figure*}[!ht]
	\centering
	\includegraphics[width=0.5\linewidth]{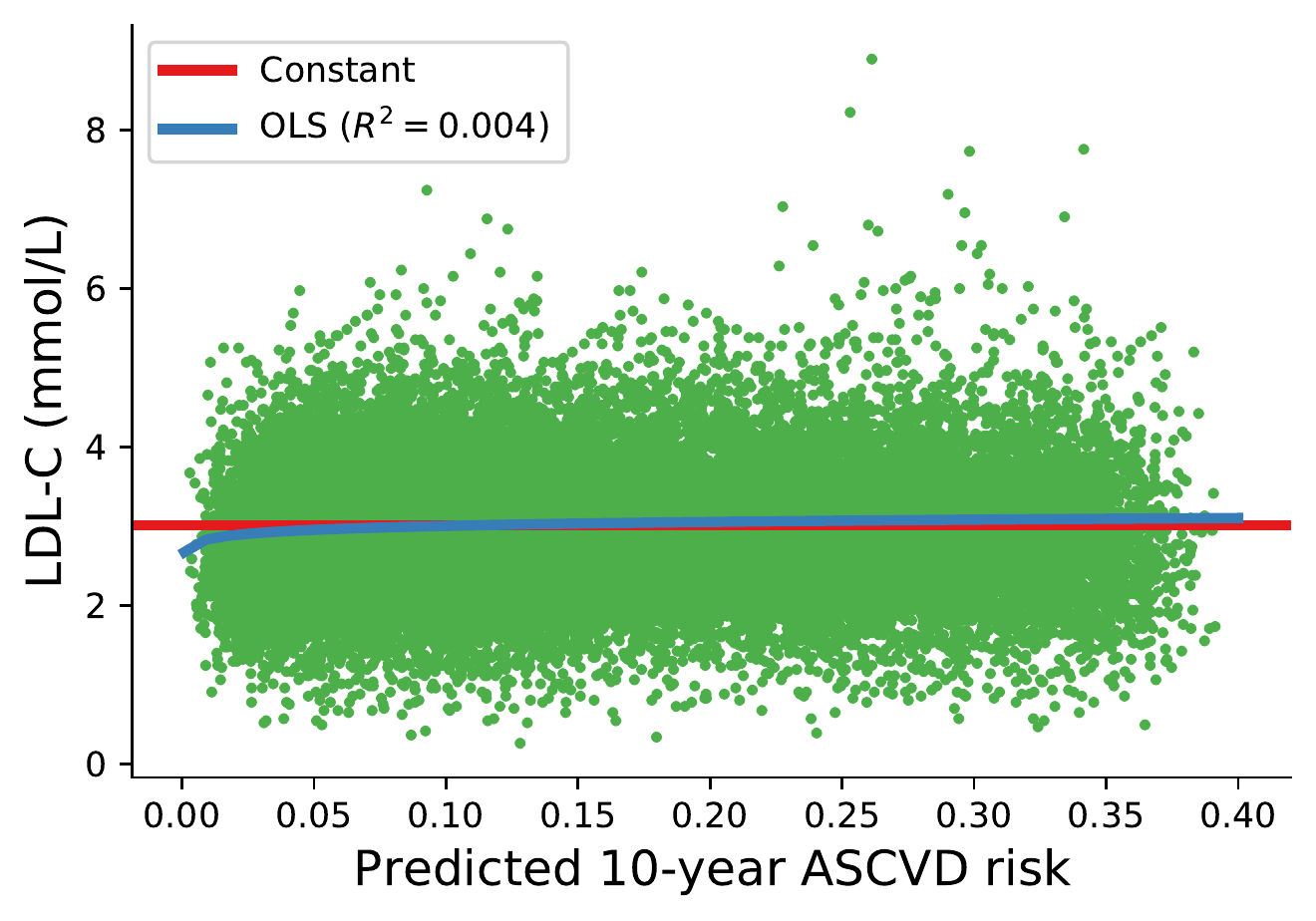}
	\caption{
	    The result of the most recent low density lipoprotein cholesterol (LDL-C) measurement versus the estimated risk of ASCVD within ten years.
	}
	\label{fig:supplement/ldl_scatter}
\end{figure*}

% Comorbidities - performance
\begin{figure*}[!t]
	\centering
	\includegraphics[width=0.95\linewidth]{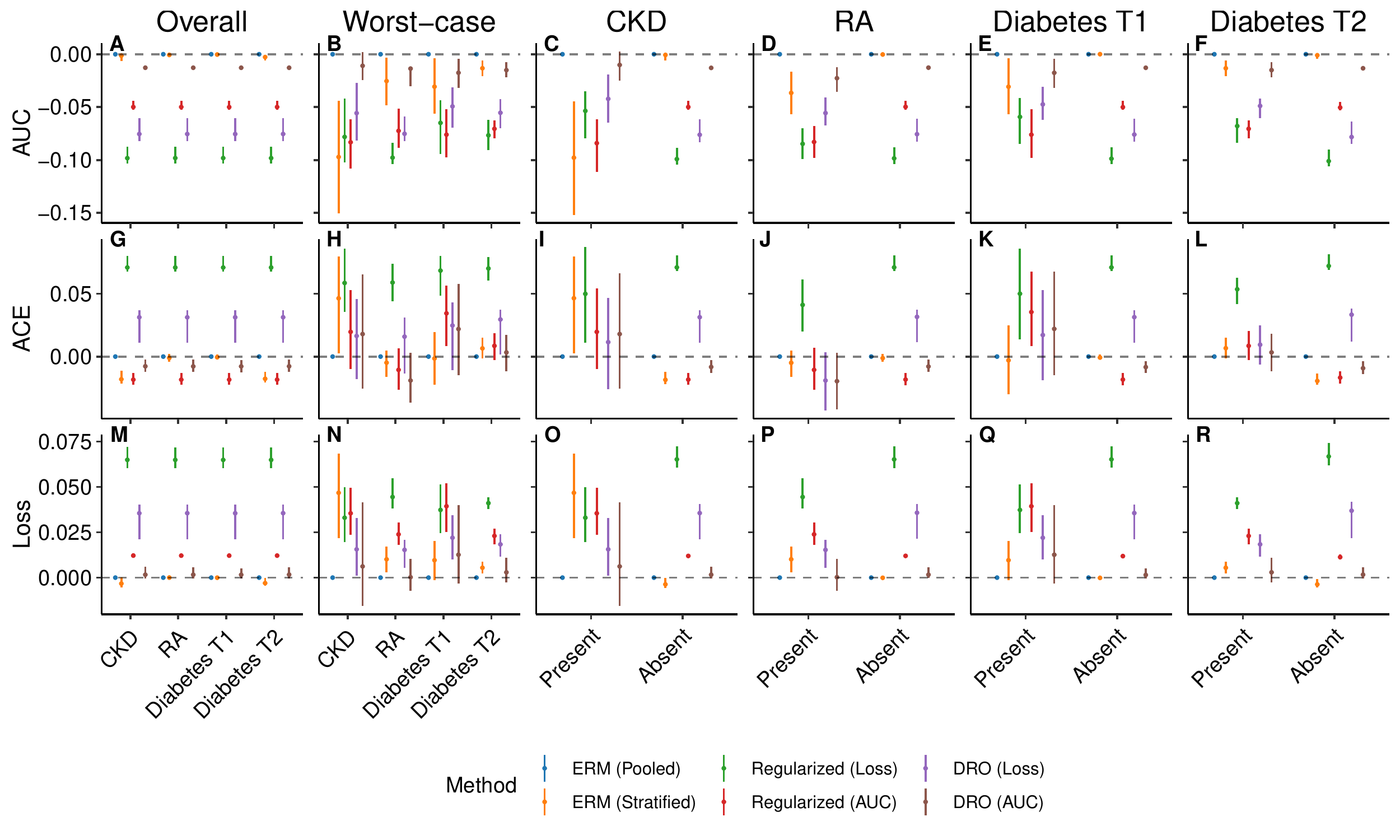}
	\caption{
    	The performance of models that estimate ten-year ASCVD risk, for subgroups defined by the presence or absence of chronic kidney disease (CKD), rheumatoid arthritis (RA), or type 1 (T1) or type 2 (T2) diabetes, relative to the results attained by the application of unpenalized ERM to the overall population.
    	Results shown are the relative AUC, absolute calibration error (ACE), and log-loss assessed in the overall population, on each subgroup, and in the worst-case over subgroups following the application of unpenalized ERM, regularized objectives that penalize differences in the log-loss or AUC across subgroups, or DRO objectives that optimize for the worst-case log-loss or AUC across subgroups.
        Error bars indicate 95\% confidence intervals derived with the percentile bootstrap with 1,000 iterations.
	}
	\label{fig:supplement/performance/comorbidities_relative}
\end{figure*}

% race/eth/sex - performance absolute
\begin{figure*}[!t]
	\centering
	\includegraphics[width=0.95\linewidth]{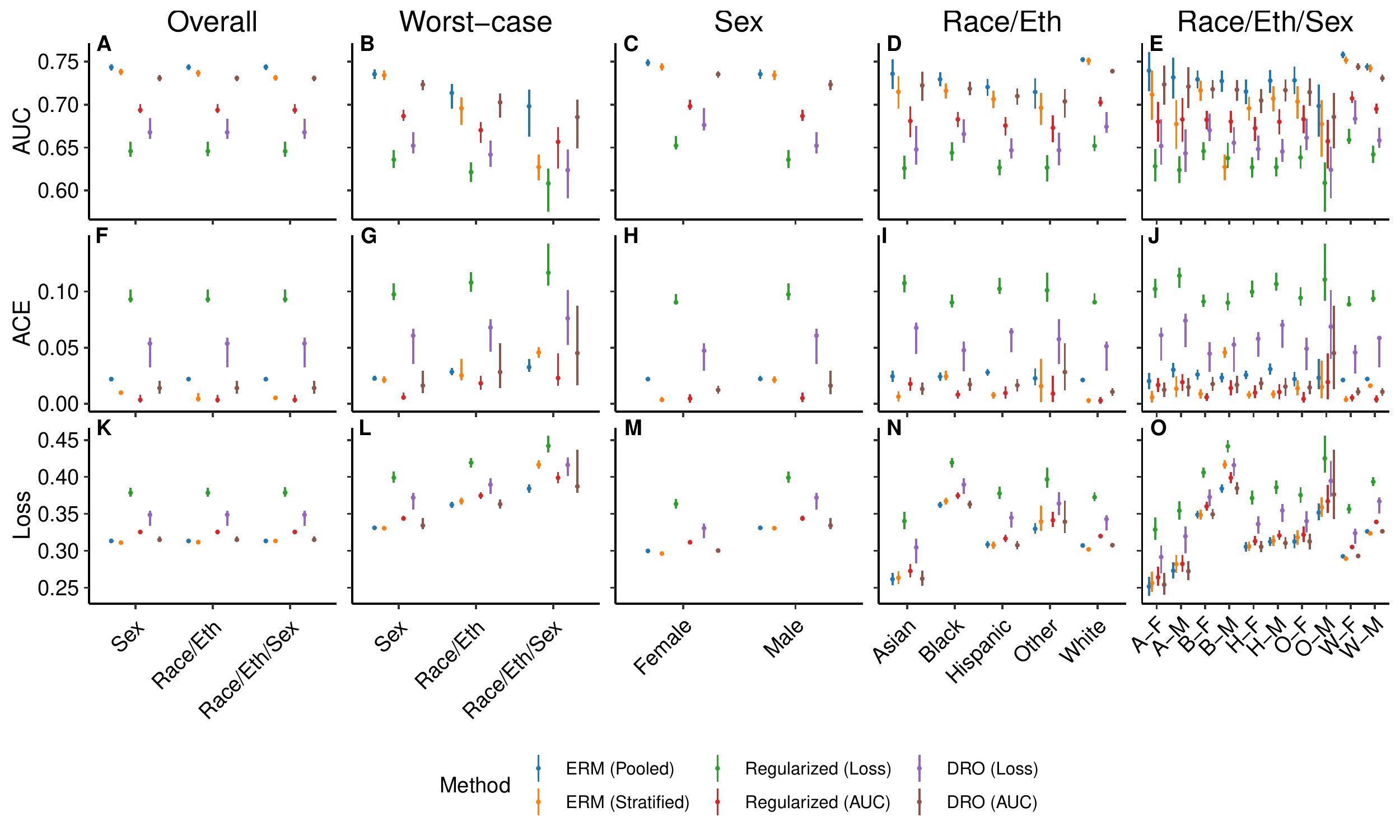}
	\caption{
    	The performance of models that estimate ten-year ASCVD risk, stratified by race, ethnicity, and sex.
    	Results shown are the AUC, absolute calibration error (ACE), and log-loss assessed in the overall population, on each subgroup, and in the worst-case over subgroups following the application of unconstrained pooled or stratified ERM, regularized objectives that penalize differences in the log-loss of AUC across subgroups, or DRO objectives that optimize for the worst-case log-loss of AUC across subgroups.
        Error bars indicate 95\% confidence intervals derived with the percentile bootstrap with 1,000 iterations.
	}
	\label{fig:supplement/performance/race_eth_sex_absolute}
\end{figure*}

% comorbidities - performance absolute
\begin{figure*}[!t]
	\centering
	\includegraphics[width=0.95\linewidth]{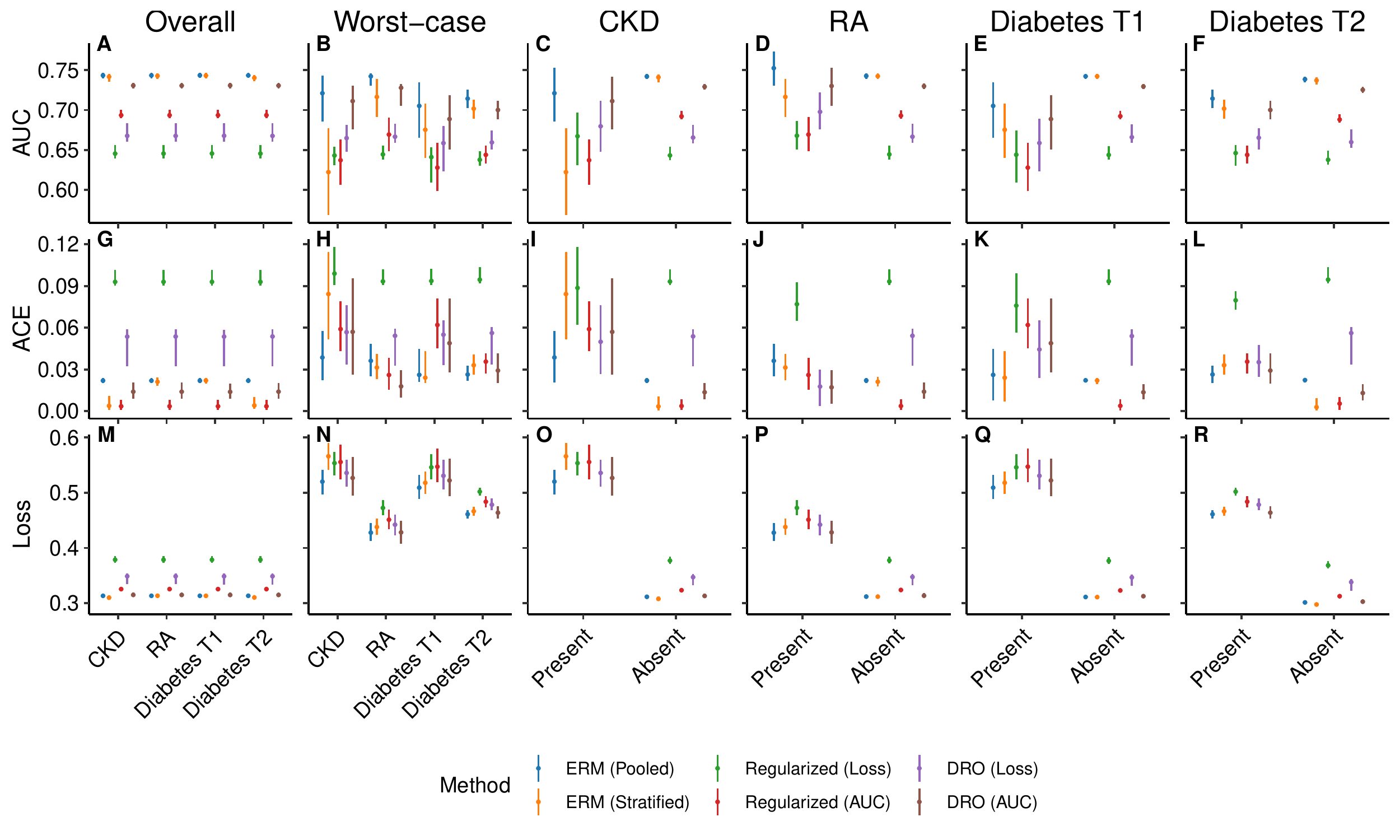}
	\caption{
    	The performance of models that estimate ten-year ASCVD risk for subgroups defined by the presence or absence of chronic kidney disease (CKD), rheumatoid arthritis (RA), or type 1 (T1) or type 2 (T2) diabetes.
    	Results shown are the AUC, absolute calibration error (ACE), and log-loss assessed in the overall population, on each subgroup, and in the worst-case over subgroups following the application of unpenalized ERM, regularized objectives that penalize differences in the log-loss of AUC across subgroups, or DRO objectives that optimize for the worst-case log-loss of AUC across subgroups.
        Error bars indicate 95\% confidence intervals derived with the percentile bootstrap with 1,000 iterations.
	}
	\label{fig:supplement/performance/comorbidities_absolute}
\end{figure*}

% Net benefit 
\begin{figure*}[!t]
	\centering
	\includegraphics[width=0.95\linewidth]{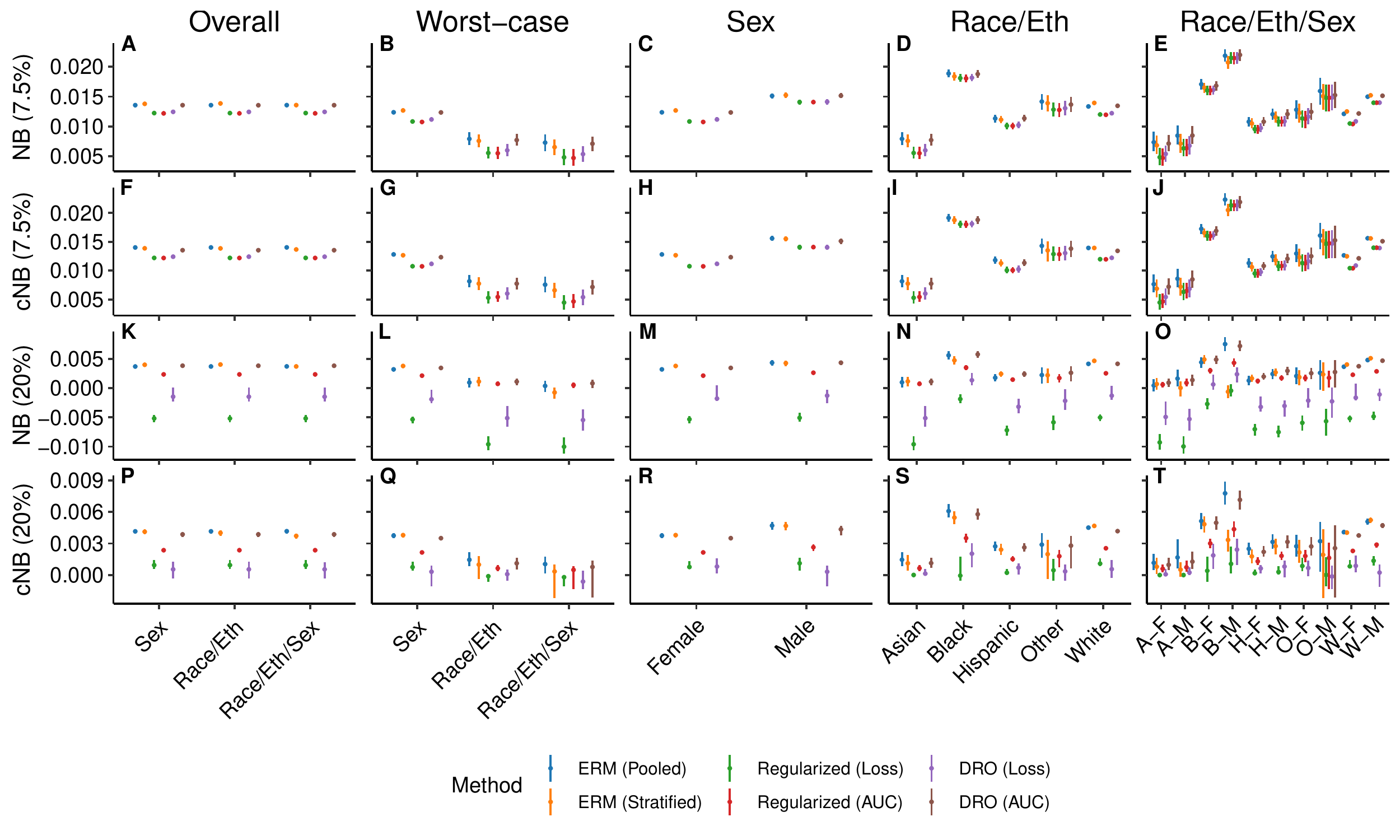}
	\caption{
    	The net benefit of models that estimate ten-year ASCVD risk, stratified by race, ethnicity, and sex.
    	Results shown are the net benefit (NB) and calibrated net benefit (cNB), parameterized by the choice of a decision threshold of 7.5\% or 20\%, assessed in the overall population, on each subgroup, and in the worst-case over subgroups following the application of unconstrained pooled or stratified ERM, regularized objectives that penalize differences in the log-loss of AUC across subgroups, or DRO objectives that optimize for the worst-case log-loss of AUC across subgroups.
        Error bars indicate 95\% confidence intervals derived with the percentile bootstrap with 1,000 iterations.
	}
	\label{fig:supplement/net_benefit/race_eth_sex_absolute}
\end{figure*}

\begin{figure*}[!t]
	\centering
	\includegraphics[width=0.95\linewidth]{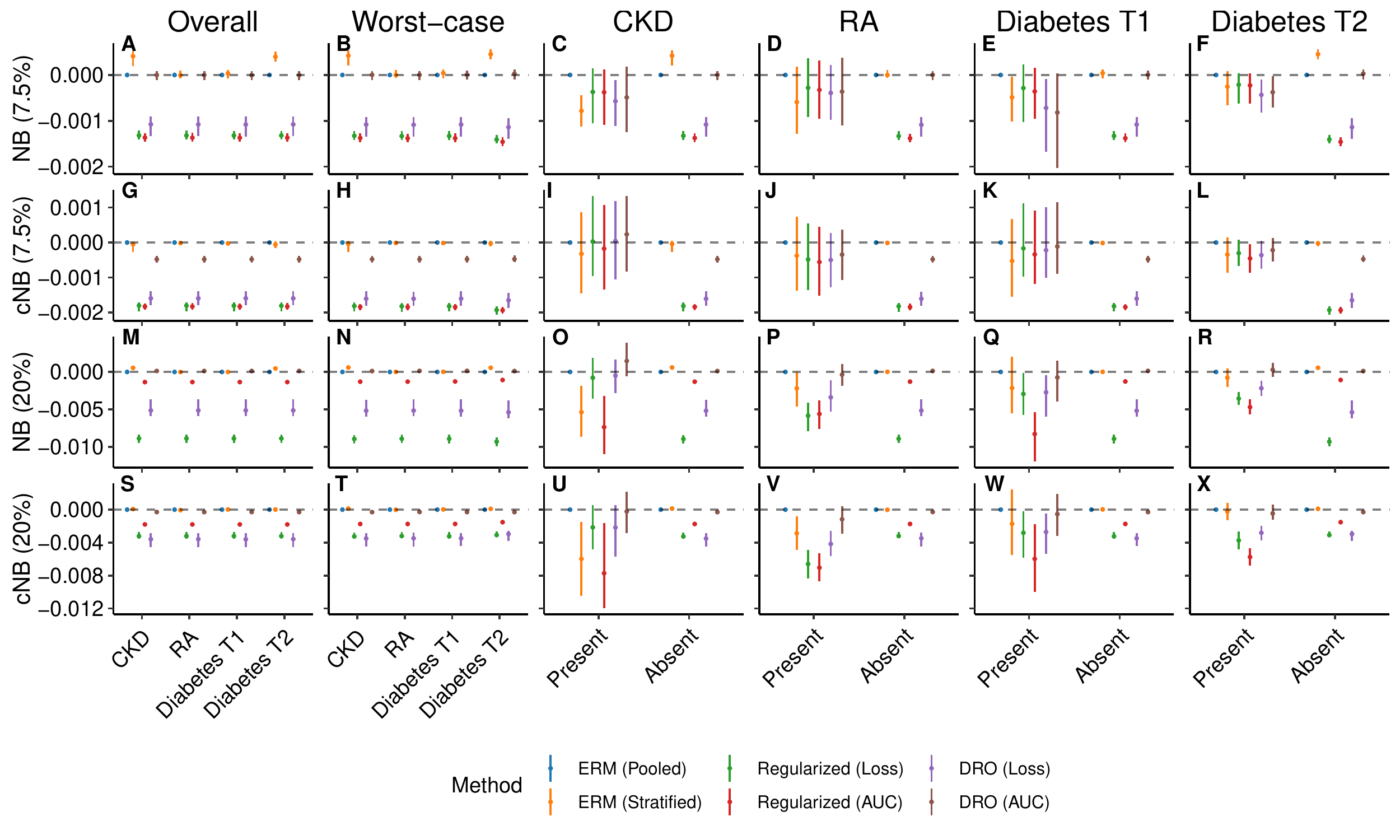}
	\caption{
    	The net benefit of models that estimate ten-year ASCVD risk, for subgroups defined by the presence or absence of chronic kidney disease (CKD), rheumatoid arthritis (RA), or type 1 (T1) or type 2 (T2) diabetes, relative to the results attained by the application of unpenalized ERM to the overall population.
    	Results shown are the net benefit (NB) and calibrated net benefit (cNB), parameterized by the choice of a decision threshold of 7.5\% or 20\%, assessed in the overall population, on each subgroup, and in the worst-case over subgroups following the application of unconstrained pooled or stratified ERM, regularized objectives that penalize differences in the log-loss or AUC across subgroups, or DRO objectives that optimize for the worst-case log-loss or AUC across subgroups.
        Error bars indicate 95\% confidence intervals derived with the percentile bootstrap with 1,000 iterations.
	}
	\label{fig:supplement/net_benefit/comorbidities_relative}
\end{figure*}

\begin{figure*}[!t]
	\centering
	\includegraphics[width=0.95\linewidth]{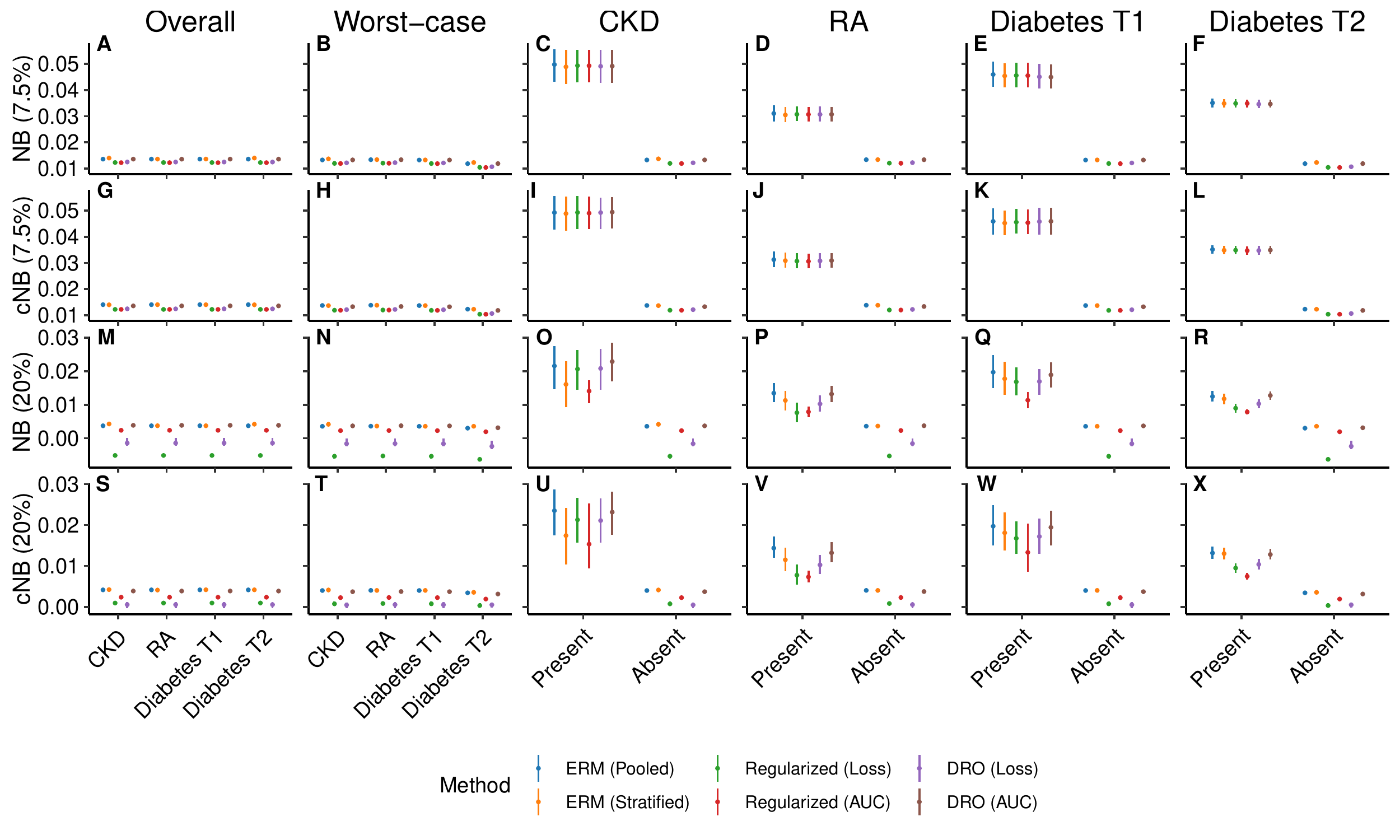}
	\caption{
    	The net benefit of models that estimate ten-year ASCVD risk for subgroups defined by the presence or absence of chronic kidney disease (CKD), rheumatoid arthritis (RA), or type 1 (T1) or type 2 (T2) diabetes.
    	Results shown are the net benefit (NB) and calibrated net benefit (cNB), parameterized by the choice of a decision threshold of 7.5\% or 20\%, assessed in the overall population, on each subgroup, and in the worst-case over subgroups following the application of unconstrained pooled or stratified ERM, regularized objectives that penalize differences in the log-loss of AUC across subgroups, or DRO objectives that optimize for the worst-case log-loss of AUC across subgroups.
        Error bars indicate 95\% confidence intervals derived with the percentile bootstrap with 1,000 iterations.
	}
	\label{fig:supplement/net_benefit/comorbidities_absolute}
\end{figure*}

%% Equalized odds experiments %%
%% Race/eth MMD
\begin{figure*}[!ht]
    
	\centering
	\includegraphics[width=0.95\linewidth]{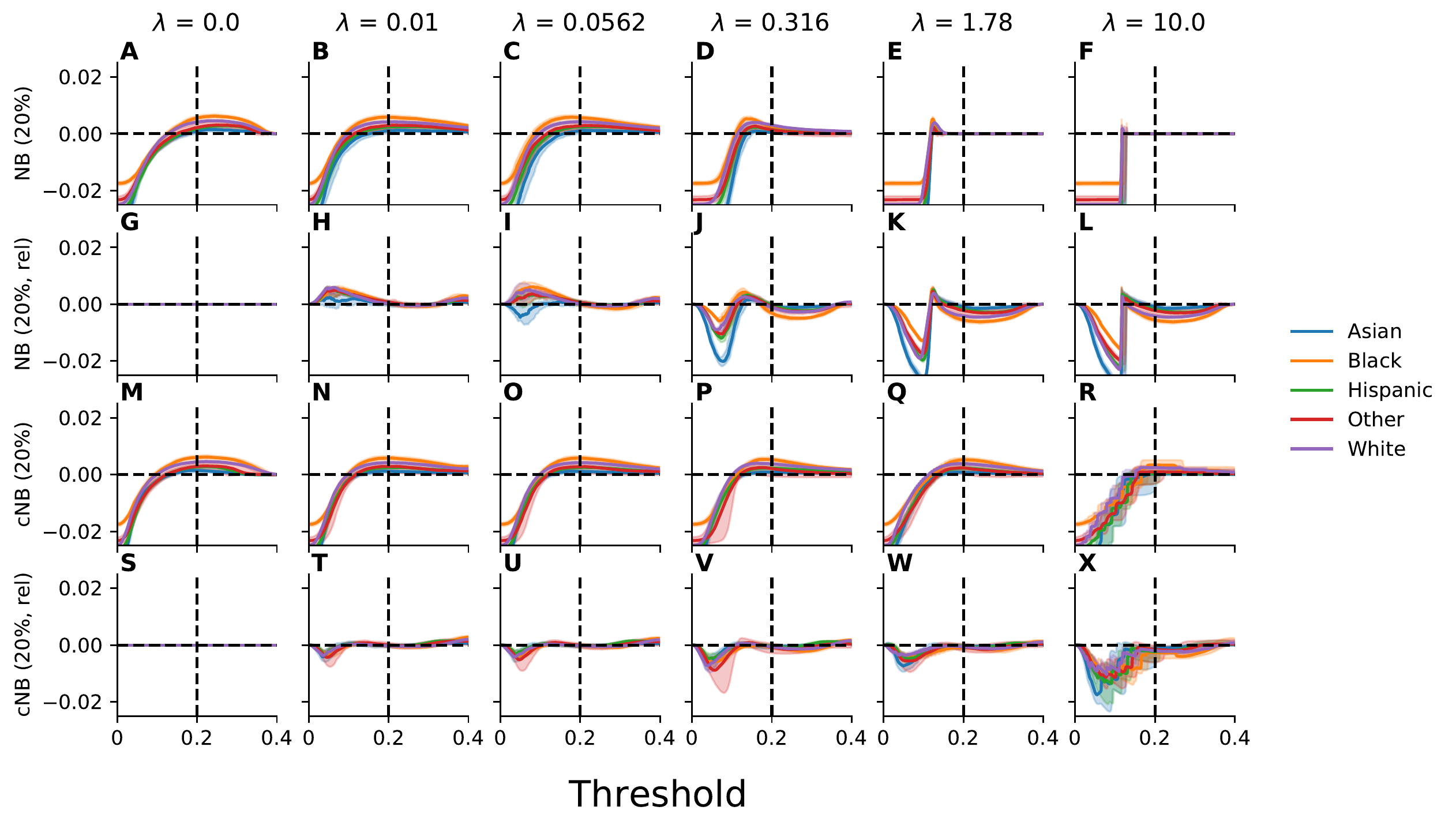}
	\caption{
	    The net benefit evaluated for a range of thresholds across racial and ethnic subgroups, parameterized by the choice of a decision threshold of 20\%, for models trained with an objective that penalizes violation of equalized odds across intersectional subgroups defined on the basis of race, ethnicity, and sex using a MMD-based penalty.
	    Plotted, for each subgroup and value of the regularization parameter $\lambda$, is the net benefit (NB) and calibrated net benefit (cNB) as a function of the decision threshold.
	    Results reported relative to the results for unconstrained empirical risk minimization are indicated by ``rel''.
	    Error bars indicate 95\% confidence intervals derived with the percentile bootstrap with 1,000 iterations.
	}
	\label{fig:supplement/eo_rr/race_eth/mmd/decision_curves_20}
\end{figure*}

\begin{figure*}[!ht]
    
	\centering
	\includegraphics[width=0.95\linewidth]{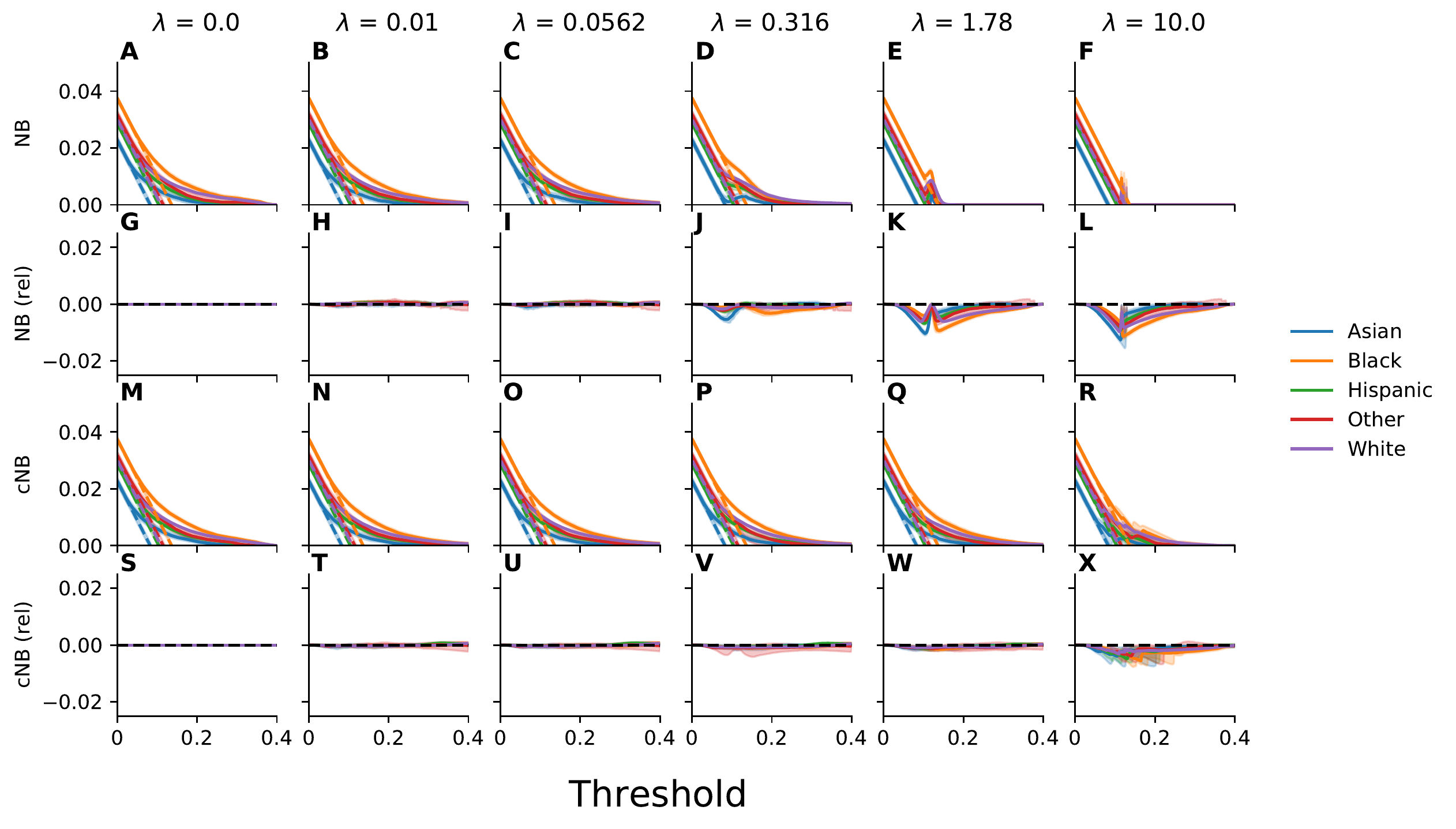}
	\caption{
	    Decision curve analysis to assess net benefit of models across racial and ethnic subgroups for models trained with an objective that penalizes violation of equalized odds across intersectional subgroups defined on the basis of race, ethnicity, and sex using a MMD-based penalty.
	    Plotted, for each subgroup and value of the regularization parameter $\lambda$, is the net benefit (NB) and calibrated net benefit (cNB) as a function of the decision threshold.
	    The net benefit of treating all patients is designated by dashed lines.
	    Results reported relative to the results for unconstrained empirical risk minimization are indicated by ``rel''.
	    Error bars indicate 95\% confidence intervals derived with the percentile bootstrap with 1,000 iterations.
	}
	\label{fig:supplement/eo_rr/race_eth/mmd/decision_curves}
\end{figure*}

%% Intersectional MMD
\begin{figure*}[!ht]
	\centering
	\includegraphics[width=0.95\linewidth]{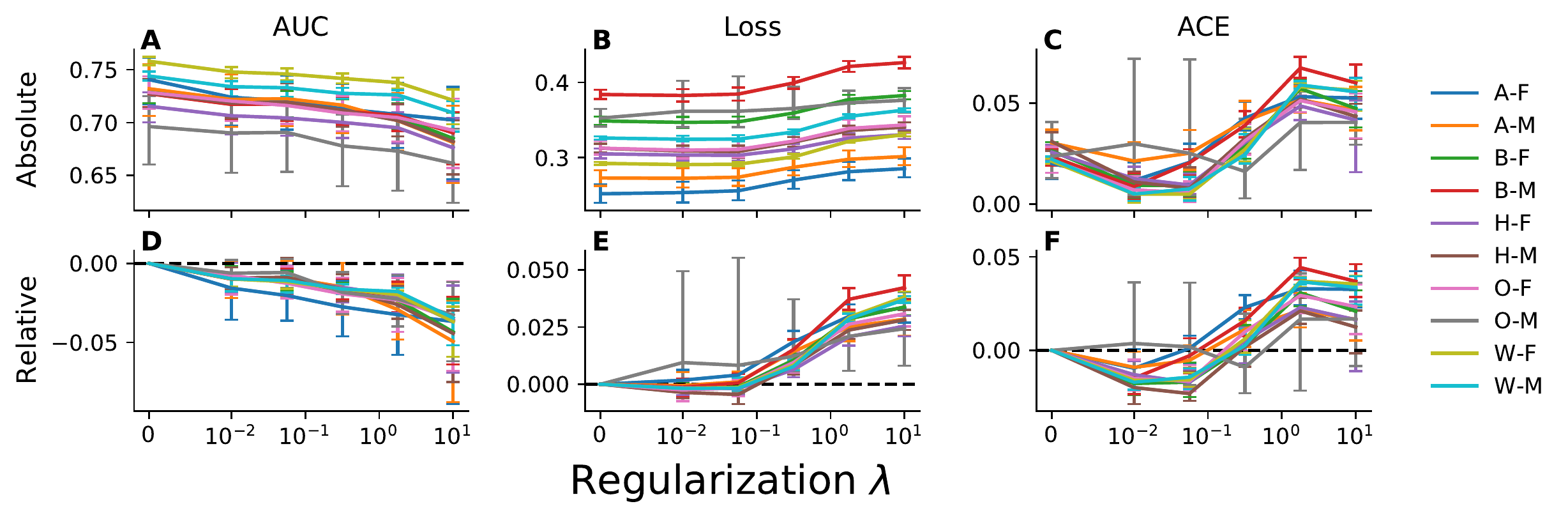}
	\caption{
	    The performance of models trained with an objective that penalizes violation of equalized odds across intersectional subgroups defined on the basis of race, ethnicity, and sex using a MMD-based penalty.
	    Plotted, for each subgroup and value of the regularization parameter $\lambda$, is the area under the receiver operating characteristic curve (AUC), log-loss, and absolute calibration error (ACE).
	    Relative results are reported relative to those attained for unconstrained empirical risk minimization. 
	    Labels correspond to Asian (A), Black (B), Hispanic (H), Other (O), White (W), Male (M), and Female (F) patients.
	    Error bars indicate 95\% confidence intervals derived with the percentile bootstrap with 1,000 iterations.
	}
	\label{fig:supplement/eo_rr/race_eth_sex/mmd/eo_performance_lambda}
\end{figure*}

\begin{figure*}[!t]
    \centering
    \includegraphics[width=0.95\linewidth]{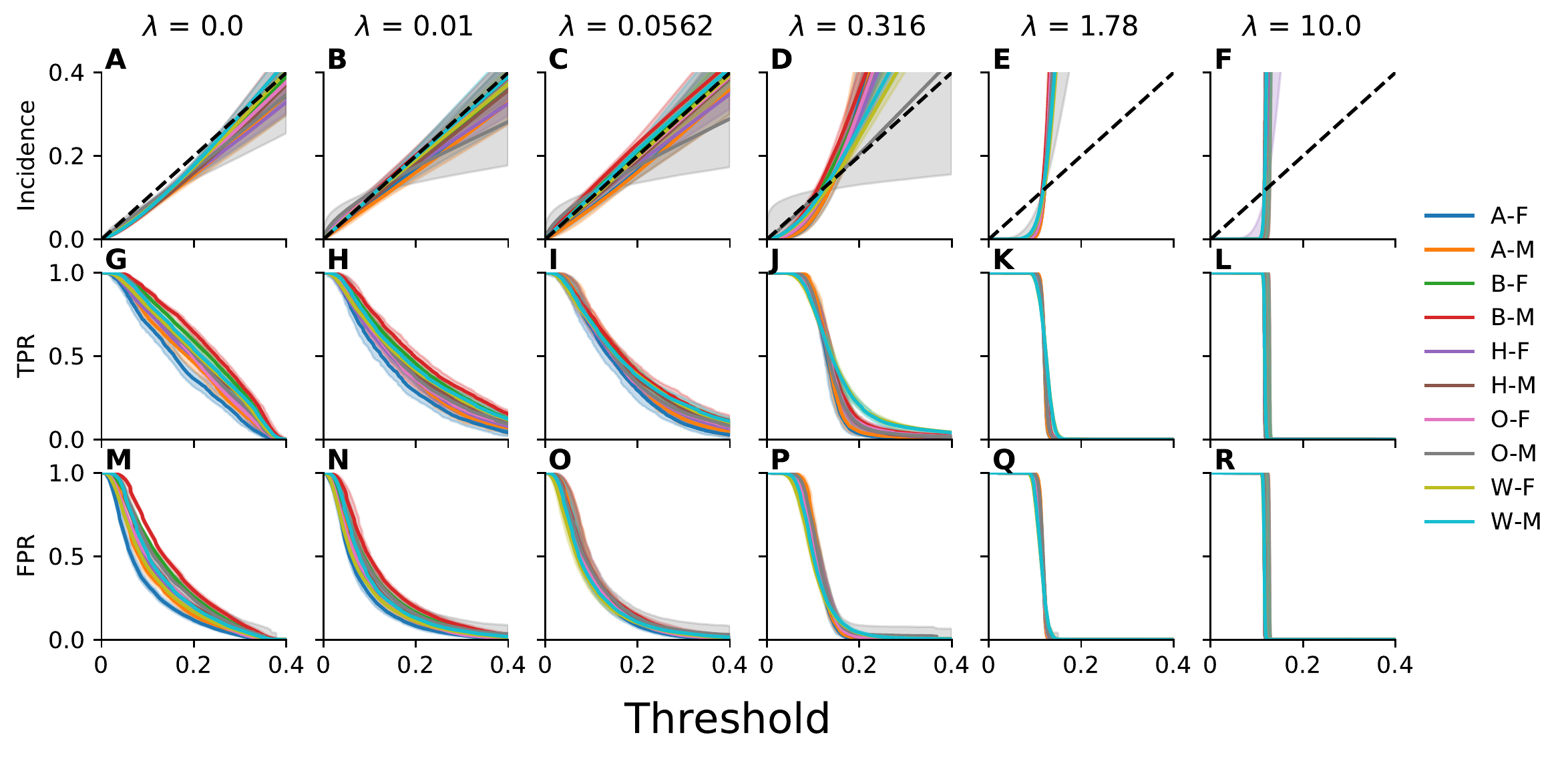}
    \caption{
        Calibration curves, true positive rates, and false positive rates evaluated for a range of thresholds for models trained with an objective that penalizes violation of equalized odds across intersectional subgroups defined on the basis of race, ethnicity, and sex using a MMD-based penalty.
        Plotted, for each subgroup and value of the regularization parameter $\lambda$, are the calibration curve (incidence), true positive rate (TPR), and false positive rate (FPR) as a function of the decision threshold.
        Error bars indicate 95\% confidence intervals derived with the percentile bootstrap with 1,000 iterations.
    }
    \label{fig:supplement/eo_rr/race_eth_sex/mmd/calibration_tpr_fpr}
\end{figure*}

\begin{figure*}[!ht]
	\centering
	\includegraphics[width=0.95\linewidth]{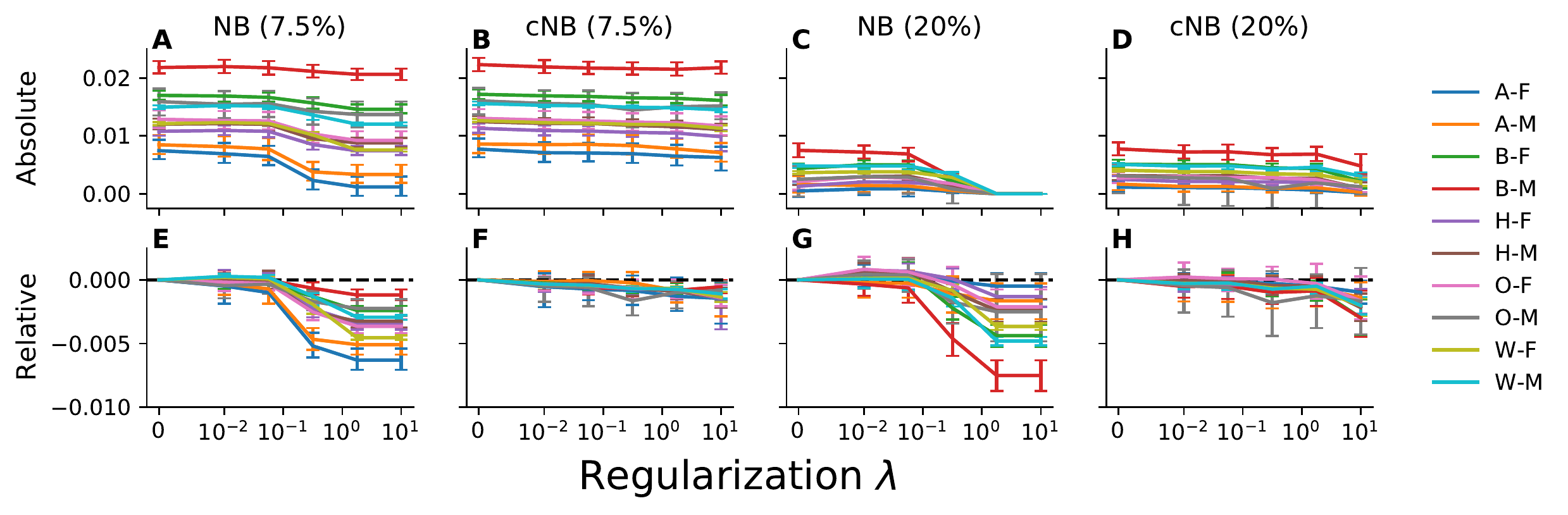}
	\caption{
	    The net benefit of models trained with an objective that penalizes violation of equalized odds across intersectional subgroups defined on the basis of race, ethnicity, and sex using a MMD-based penalty, parameterized by the choice of a decision threshold of 7.5\% or 20\%.
	    Plotted, for each subgroup is the net benefit (NB) and calibrated net benefit (rNB) as a function of the value of the regularization parameter $\lambda$, .
	    Relative results are reported relative to those attained for unconstrained empirical risk minimization.
	    Labels correspond to Asian (A), Black (B), Hispanic (H), Other (O), White (W), Male (M), and Female (F) patients.
	    Error bars indicate 95\% confidence intervals derived with the percentile bootstrap with 1,000 iterations.
	}
	\label{fig:supplement/eo_rr/race_eth_sex/mmd/eo_net_benefit_lambda}
\end{figure*}

\begin{figure*}[!ht]
	\centering
	\includegraphics[width=0.95\linewidth]{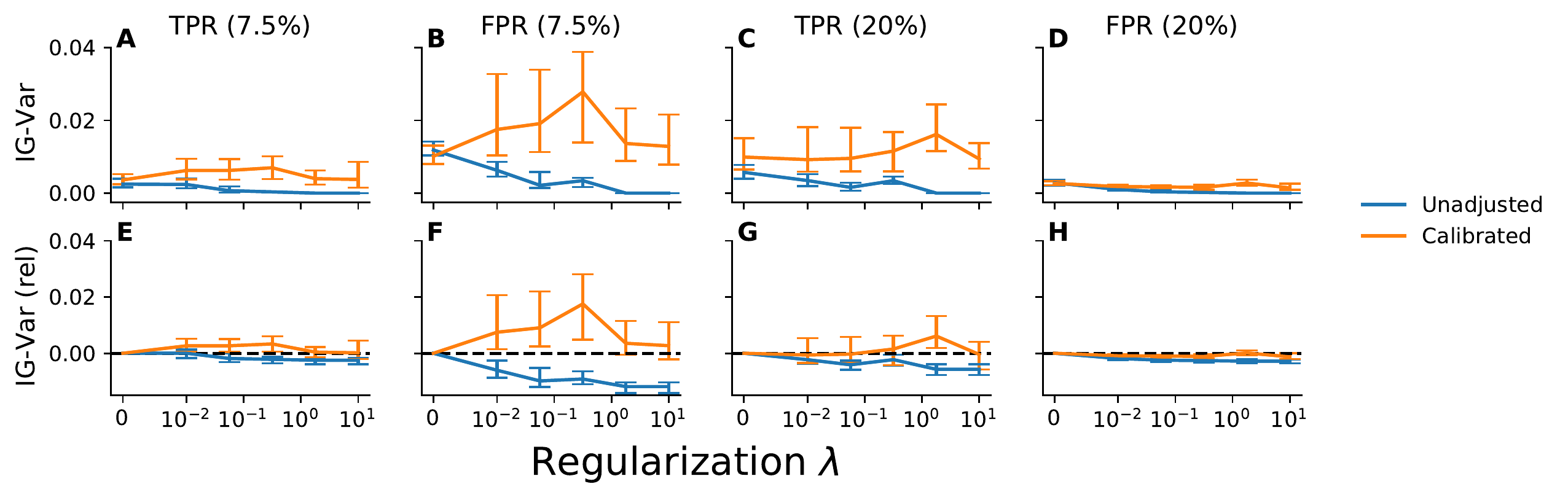}
	\caption{
        Satisfaction of equalized odds for models trained with an objective that penalizes violation of equalized odds across intersectional subgroups defined on the basis of race, ethnicity, and sex using a MMD-based penalty.
        Plotted is the intergroup variance (IG-Var) in the true positive and false positive rates at decision thresholds of 7.5\% and 20\%.
        Recalibrated results correspond to those attained for models for which the threshold has been adjusted to account for the observed miscalibration.
        Relative results are reported relative to those attained for unconstrained empirical risk minimization.
        Labels correspond to Asian (A), Black (B), Hispanic (H), Other (O), White (W), Male (M), and Female (F) patients.
	    Error bars indicate 95\% confidence intervals derived with the percentile bootstrap with 1,000 iterations.
	}
	\label{fig:supplement/eo_rr/race_eth_sex/mmd/eo_tpr_fpr_var_lambda}
\end{figure*}

\begin{figure*}[!ht]
	\centering
	\includegraphics[width=0.95\linewidth]{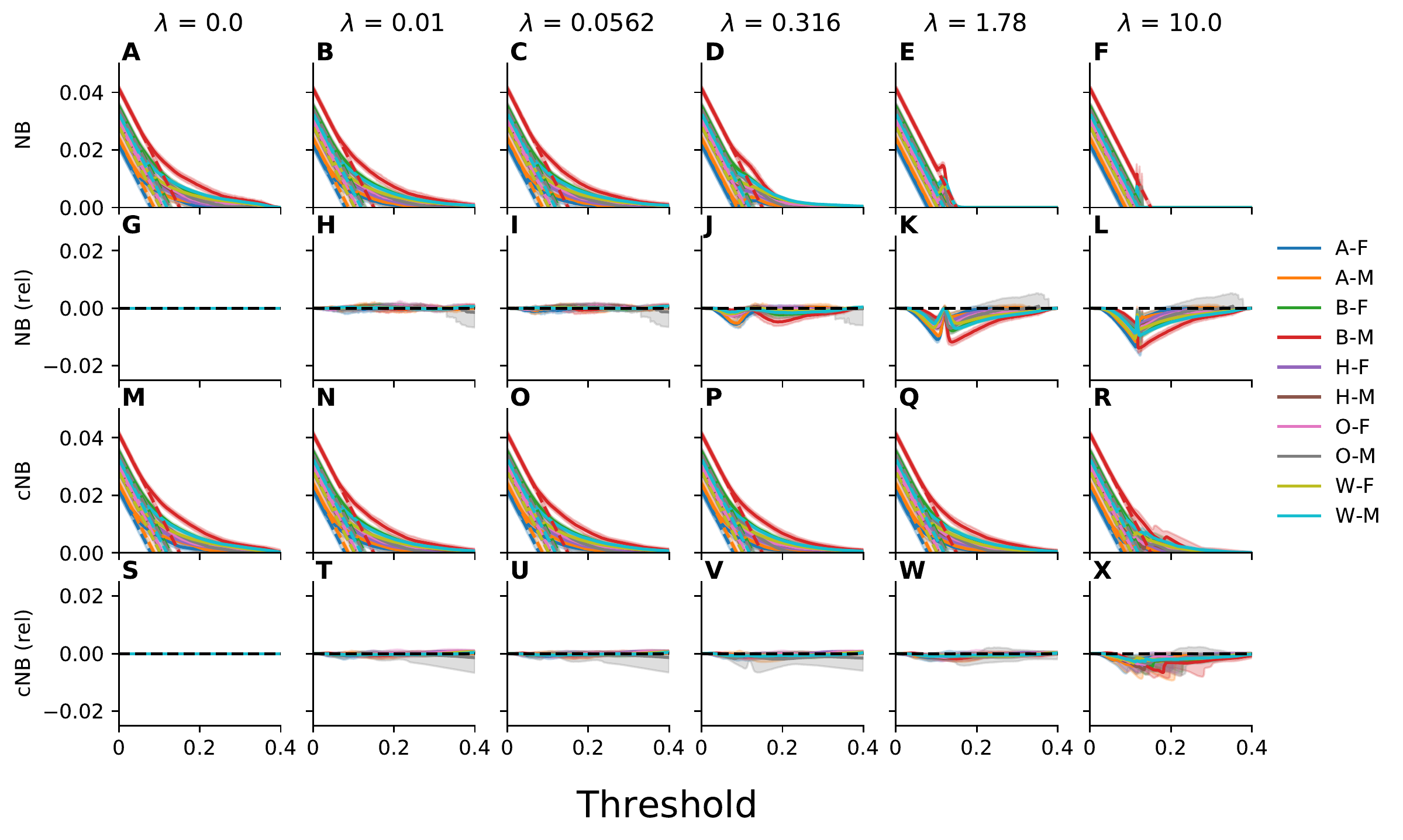}
	\caption{
	    Decision curve analysis to assess net benefit of models trained with an objective that penalizes violation of equalized odds across intersectional subgroups defined on the basis of race, ethnicity, and sex using a MMD-based penalty.
	    Plotted, for each subgroup and value of the regularization parameter $\lambda$, is the net benefit (NB) and calibrated net benefit (rNB) as a function of the decision threshold.
	    The net benefit of treating all patients is designated by dashed lines.
	    Results reported relative to the results for unconstrained empirical risk minimization are indicated by ``rel''.
	    Labels correspond to Asian (A), Black (B), Hispanic (H), Other (O), White (W), Male (M), and Female (F) patients.
	    Error bars indicate 95\% confidence intervals derived with the percentile bootstrap with 1,000 iterations.
	}
	\label{fig:supplement/eo_rr/race_eth_sex/mmd/decision_curves}
\end{figure*}

\begin{figure*}[!ht]
	\centering
	\includegraphics[width=0.95\linewidth]{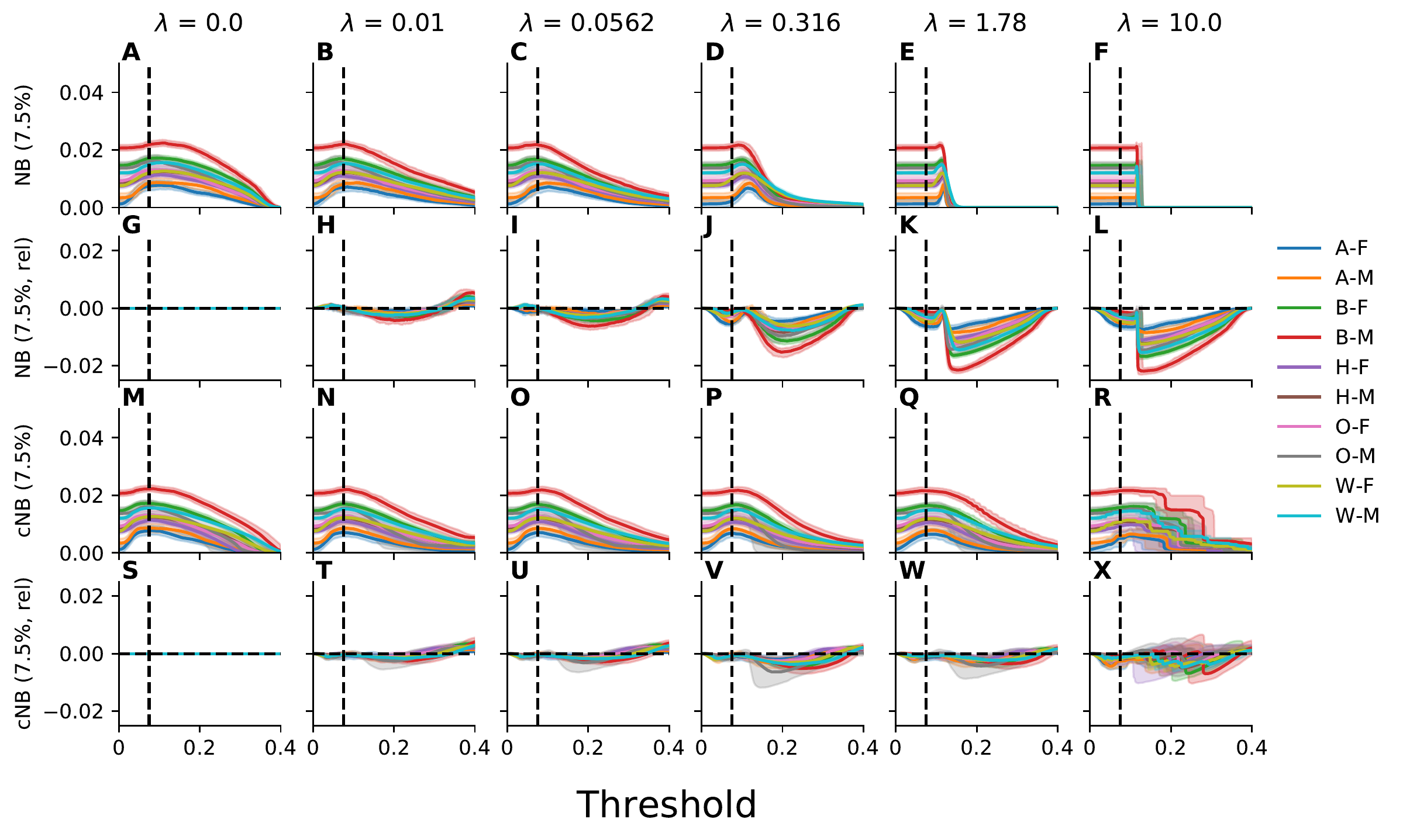}
	\caption{
	    The net benefit of models trained with an objective that penalizes violation of equalized odds across intersectional subgroups defined on the basis of race, ethnicity, and sex using a MMD-based penalty, for the net benefit parameterized by the choice of a decision threshold of 7.5\%.
	    Plotted, for each subgroup and value of the regularization parameter $\lambda$, is the net benefit (NB) and calibrated net benefit (rNB) as a function of the decision threshold.
	    The net benefit of treating all patients is designated by dashed lines.
	    Results reported relative to the results for unconstrained empirical risk minimization are indicated by ``rel''.
	    Labels correspond to Asian (A), Black (B), Hispanic (H), Other (O), White (W), Male (M), and Female (F) patients.
	    Error bars indicate 95\% confidence intervals derived with the percentile bootstrap with 1,000 iterations.
	}
	\label{fig:supplement/eo_rr/race_eth_sex/mmd/decision_curves_075}
\end{figure*}

\begin{figure*}[!ht]
	\centering
	\includegraphics[width=0.95\linewidth]{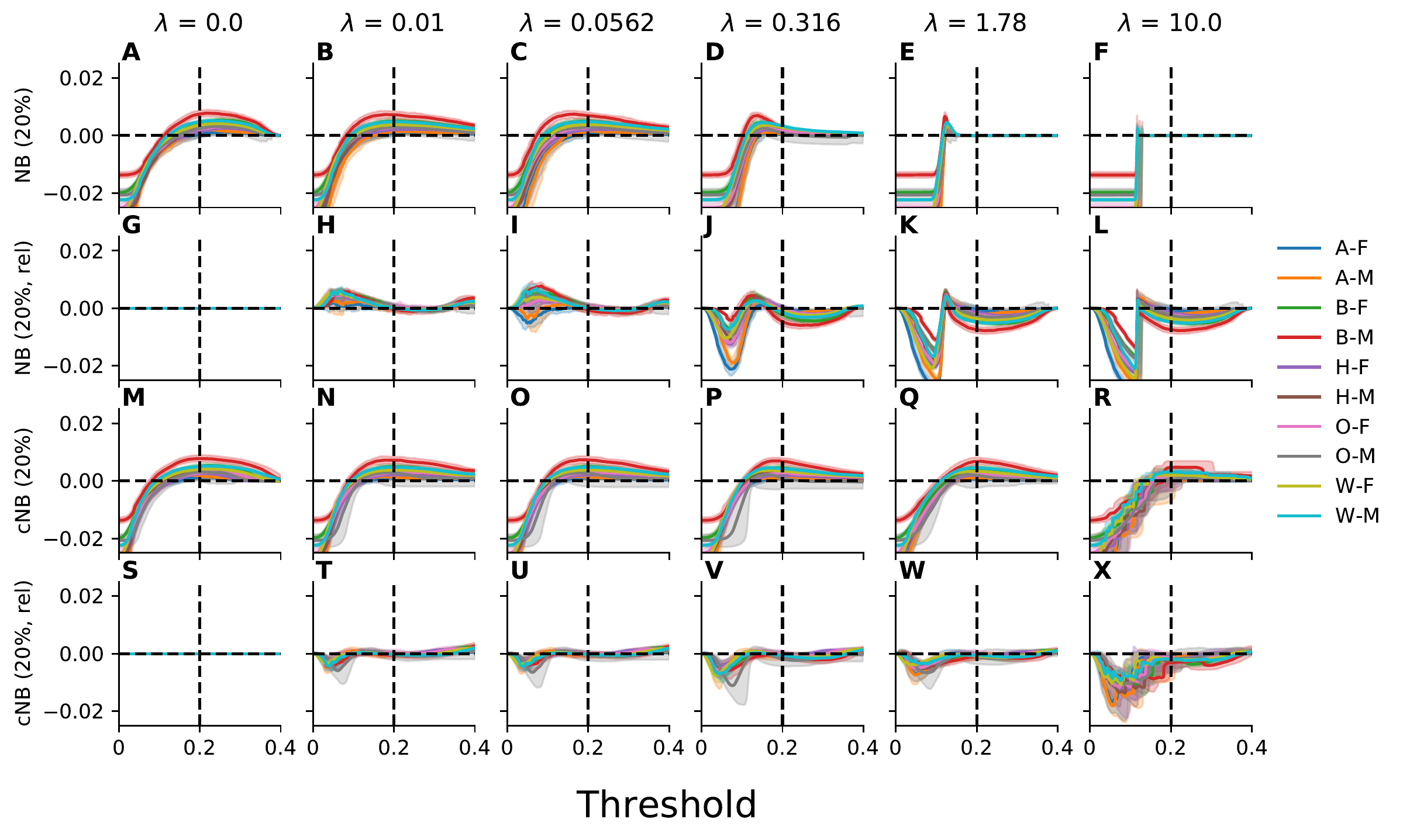}
	\caption{
	    The net benefit of models trained with an objective that penalizes violation of equalized odds across intersectional subgroups defined on the basis of race, ethnicity, and sex using a MMD-based penalty, for the net benefit parameterized by the choice of a decision threshold of 20\%.
	    Plotted, for each subgroup and value of the regularization parameter $\lambda$, is the net benefit (NB) and calibrated net benefit (rNB) as a function of the decision threshold.
	    The net benefit of treating all patients is designated by dashed lines.
	    Results reported relative to the results for unconstrained empirical risk minimization are indicated by ``rel''.
	    Labels correspond to Asian (A), Black (B), Hispanic (H), Other (O), White (W), Male (M), and Female (F) patients.
	    Error bars indicate 95\% confidence intervals derived with the percentile bootstrap with 1,000 iterations.
	}
	\label{fig:supplement/eo_rr/race_eth_sex/mmd/decision_curves_20}
\end{figure*}

% Sex MMD
\begin{figure*}[!t]
    \centering
    \includegraphics[width=0.95\linewidth]{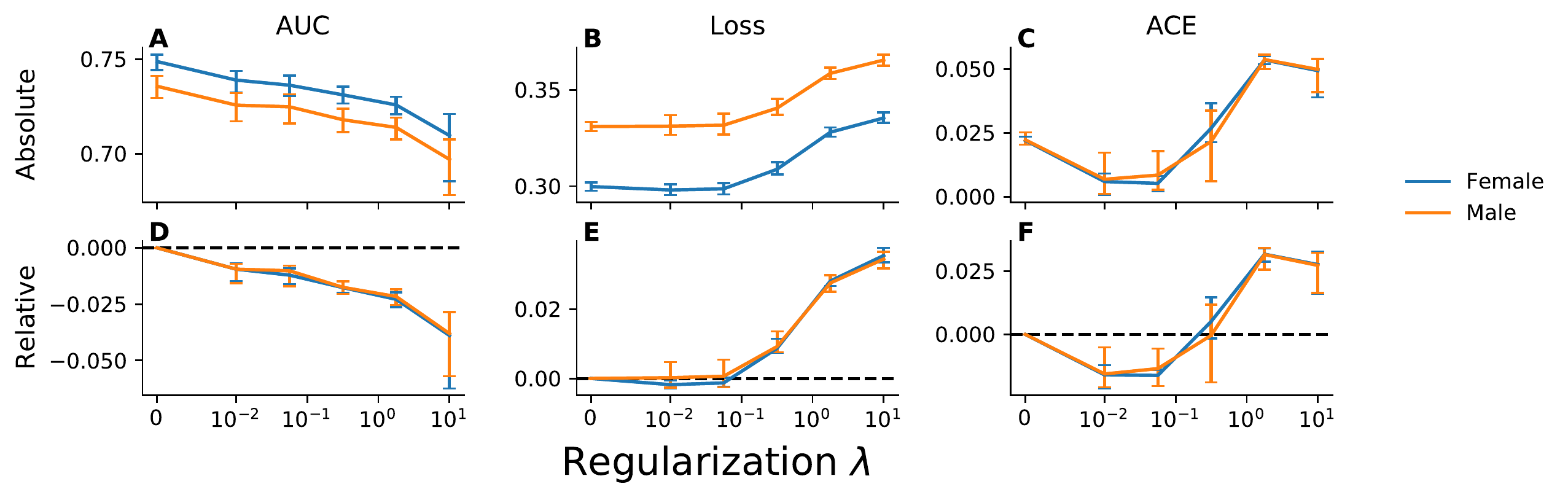}
    \caption{
        Model performance evaluated across subgroups defined by sex for models trained with an objective that penalizes violation of equalized odds across intersectional subgroups defined on the basis of race, ethnicity, and sex using a MMD-based penalty.
        Plotted, for each subgroup and value of the regularization parameter $\lambda$, is the area under the receiver operating characteristic curve (AUC), log-loss, and absolute calibration error (ACE).
        Relative results are reported relative to those attained for unconstrained empirical risk minimization. 
        Error bars indicate 95\% confidence intervals derived with the percentile bootstrap with 1,000 iterations.
    }
    \label{fig:supplement/eo_rr/sex/mmd/eo_performance_lambda}
\end{figure*}

% Performance-EO grid
\begin{figure*}[!t]
    \centering
    \includegraphics[width=0.95\linewidth]{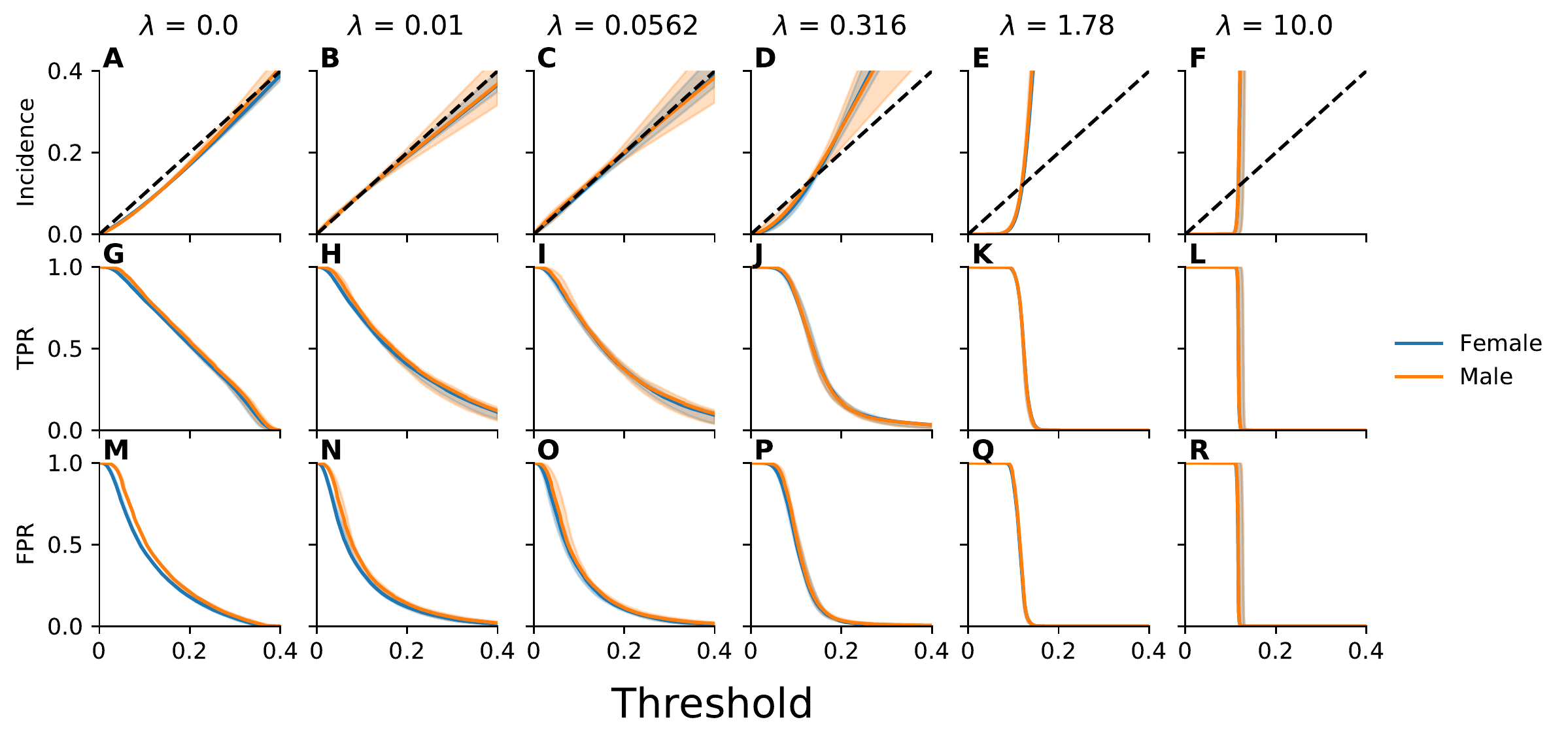}
    \caption{
        Calibration curves, true positive rates, and false positive rates evaluated for a range of thresholds across subgroups defined by sex for models trained with an objective that penalizes violation of equalized odds across intersectional subgroups defined on the basis of race, ethnicity, and sex using a MMD-based penalty.
        Plotted, for each subgroup and value of the regularization parameter $\lambda$, are the calibration curve (incidence), true positive rate (TPR), and false positive rate (FPR) as a function of the decision threshold.
        Error bars indicate 95\% confidence intervals derived with the percentile bootstrap with 1,000 iterations.
    }
    \label{fig:supplement/eo_rr/sex/mmd/calibration_tpr_fpr}
\end{figure*}

% Net benefit metrics as a function of lambda
\begin{figure*}[!t]
    \centering
    \includegraphics[width=0.95\linewidth]{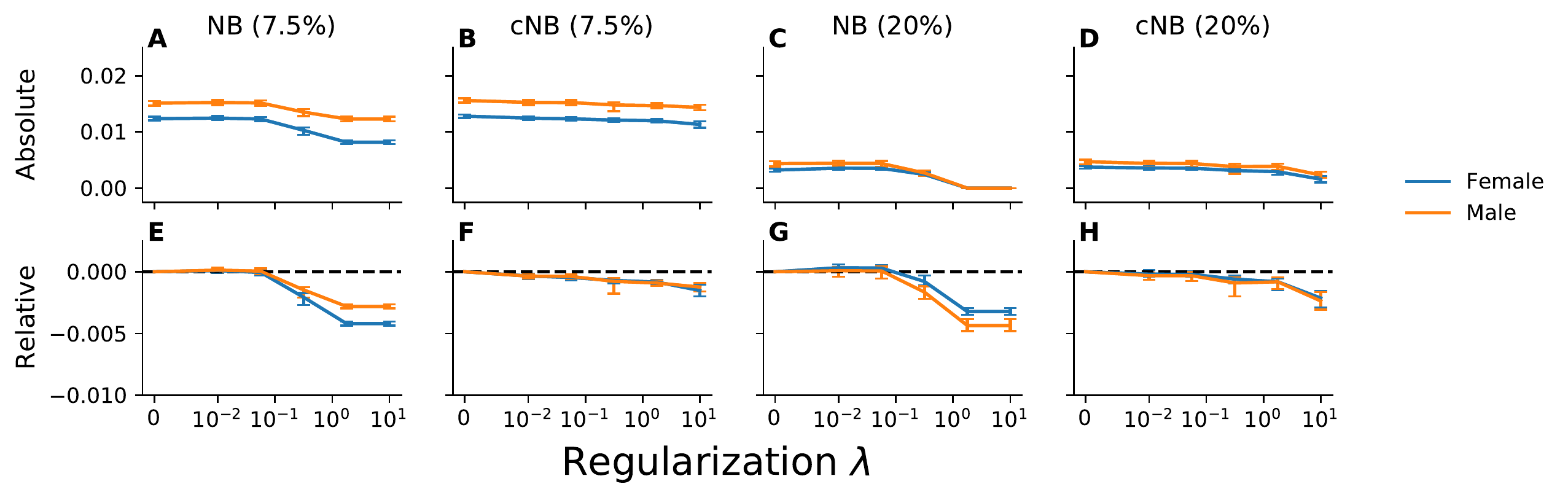}
    \caption{
        The net benefit evaluated across subgroups defined by sex, parameterized by the choice of a decision threshold of 7.5\% or 20\%, for models trained with an objective that penalizes violation of equalized odds across intersectional subgroups defined on the basis of race, ethnicity, and sex using a MMD-based penalty.
        Plotted, for each subgroup is the net benefit (NB) and calibrated net benefit (cNB) as a function of the value of the regularization parameter $\lambda$.
        Relative results are reported relative to those attained for unconstrained empirical risk minimization.
        Error bars indicate 95\% confidence intervals derived with the percentile bootstrap with 1,000 iterations.
    }
    \label{fig:supplement/eo_rr/sex/mmd/eo_net_benefit_lambda}
\end{figure*}

% Equalized odds satisfaction
\begin{figure*}[!t]
    \centering
    \includegraphics[width=0.95\linewidth]{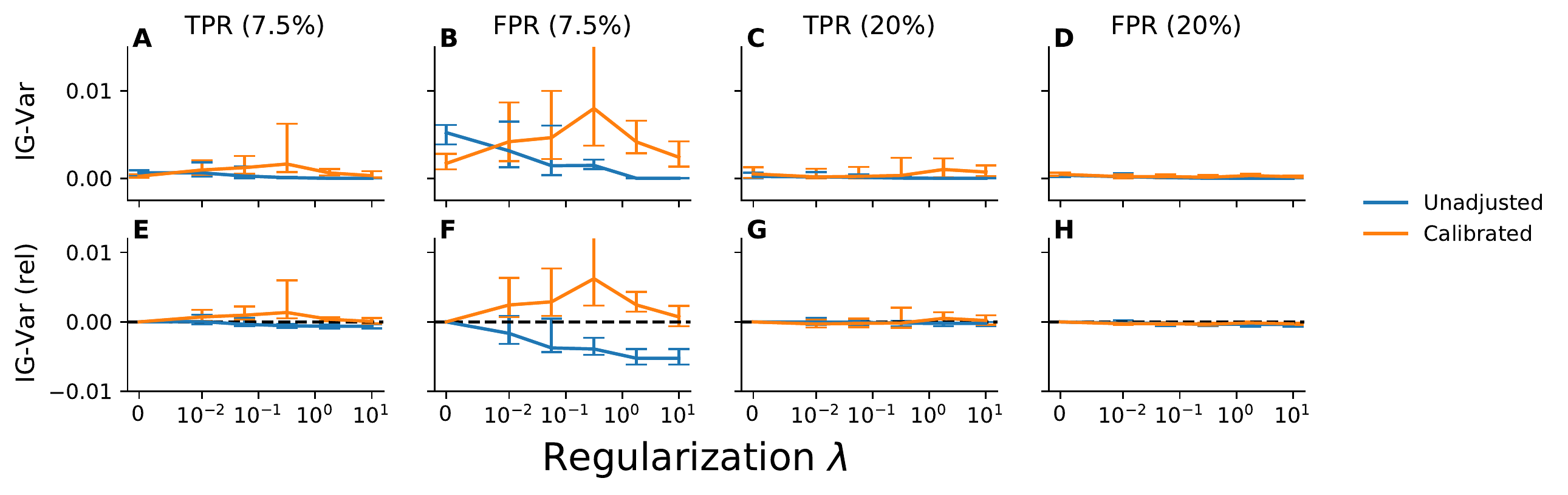}
    \caption{
        Satisfaction of equalized odds evaluated across subgroups defined by sex for models trained with an objective that penalizes violation of equalized odds across intersectional subgroups defined on the basis of race, ethnicity, and sex using a MMD-based penalty.
        Plotted is the intergroup variance (IG-Var) in the true positive and false positive rates at decision thresholds of 7.5\% and 20\%.
        Recalibrated results correspond to those attained for models for which the threshold has been adjusted to account for the observed miscalibration.
        Relative results are reported relative to those attained for unconstrained empirical risk minimization.
        Error bars indicate 95\% confidence intervals derived with the percentile bootstrap with 1,000 iterations.
    }
    \label{fig:supplement/eo_rr/sex/mmd/eo_tpr_fpr_var_lambda}
\end{figure*}

% Decision curves at t=0.075
\begin{figure*}[!ht]
    \centering
    \includegraphics[width=0.95\linewidth]{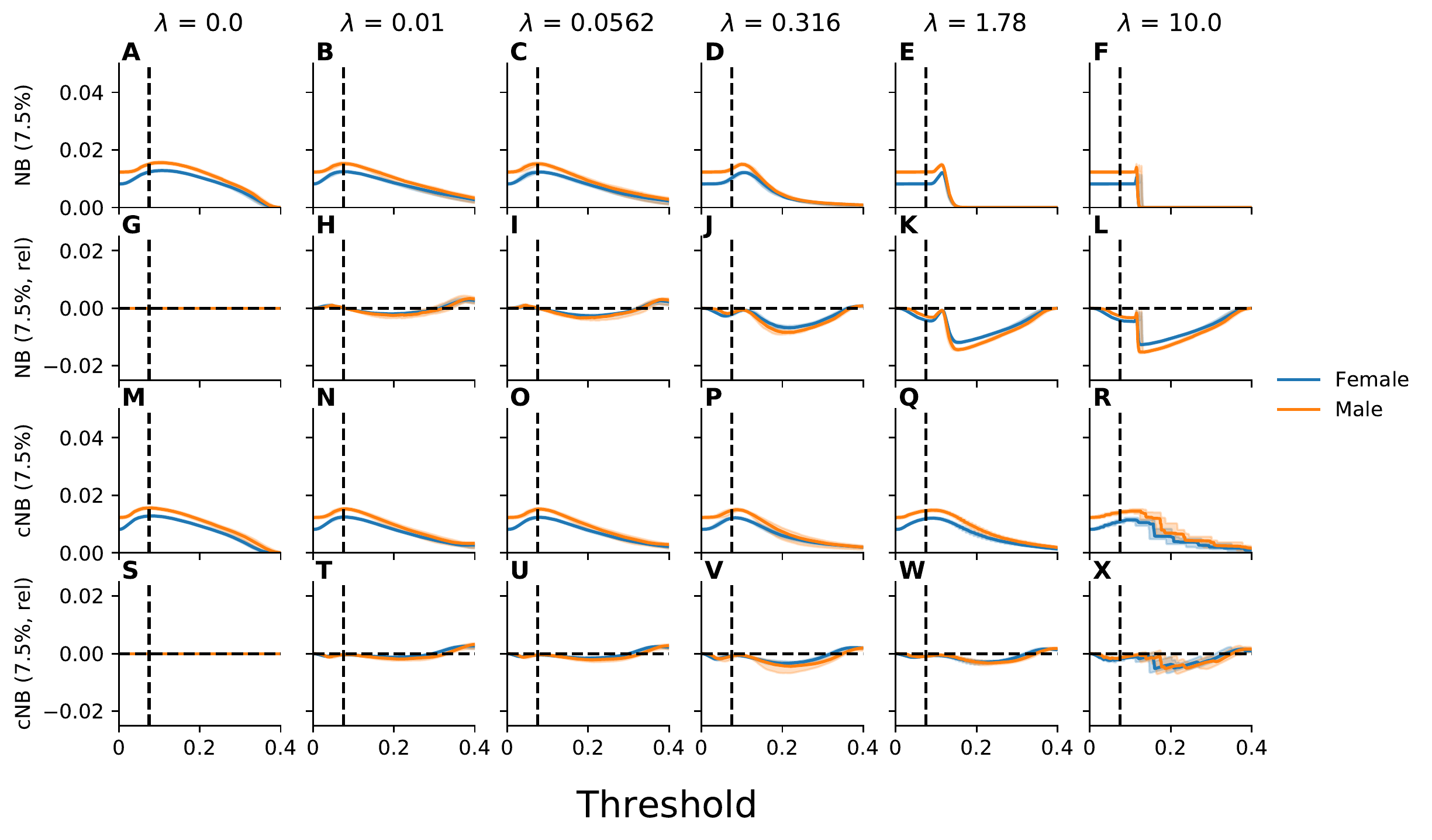}
    \caption{
        The net benefit evaluated for a range of thresholds across subgroups defined by sex, parameterized by the choice of a decision threshold of 7.5\%, for models trained with an objective that penalizes violation of equalized odds across intersectional subgroups defined on the basis of race, ethnicity, and sex using a MMD-based penalty.
        Plotted, for each subgroup and value of the regularization parameter $\lambda$, is the net benefit (NB) and calibrated net benefit (cNB) as a function of the decision threshold.
        Results reported relative to the results for unconstrained empirical risk minimization are indicated by ``rel''.
        Error bars indicate 95\% confidence intervals derived with the percentile bootstrap with 1,000 iterations.
    }
    \label{fig:supplement/eo_rr/sex/mmd/decision_curves_075}
\end{figure*}

\begin{figure*}[!ht]
    
	\centering
	\includegraphics[width=0.95\linewidth]{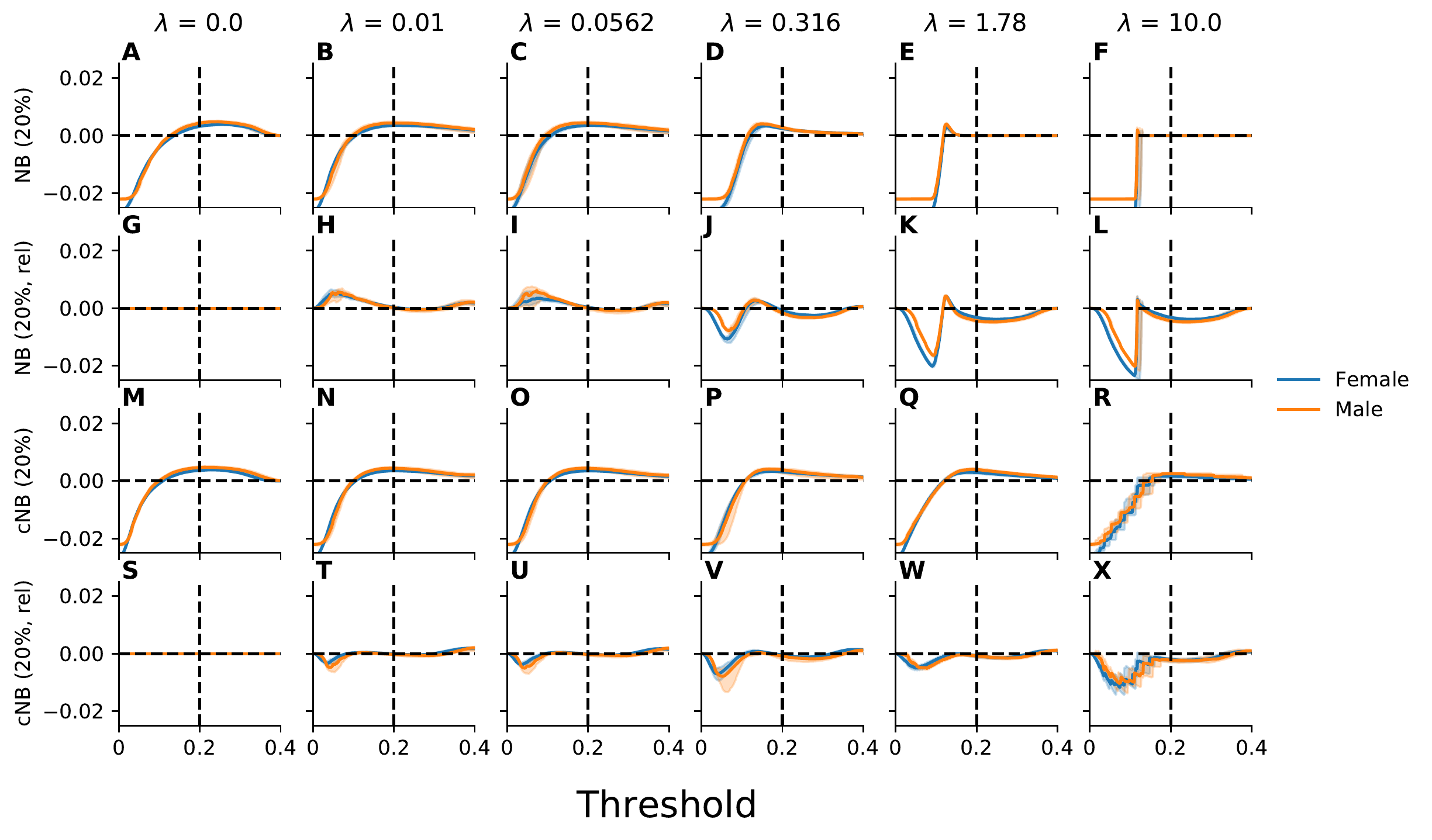}
	\caption{
	    The net benefit evaluated for a range of thresholds across subgroups defined by sex, parameterized by the choice of a decision threshold of 20\%, for models trained with an objective that penalizes violation of equalized odds across intersectional subgroups defined on the basis of race, ethnicity, and sex using a MMD-based penalty.
	    Plotted, for each subgroup and value of the regularization parameter $\lambda$, is the net benefit (NB) and calibrated net benefit (cNB) as a function of the decision threshold.
	    Results reported relative to the results for unconstrained empirical risk minimization are indicated by ``rel''.
	    Error bars indicate 95\% confidence intervals derived with the percentile bootstrap with 1,000 iterations.
	}
	\label{fig:supplement/eo_rr/sex/mmd/decision_curves_20}
\end{figure*}

\begin{figure*}[!ht]
    
	\centering
	\includegraphics[width=0.95\linewidth]{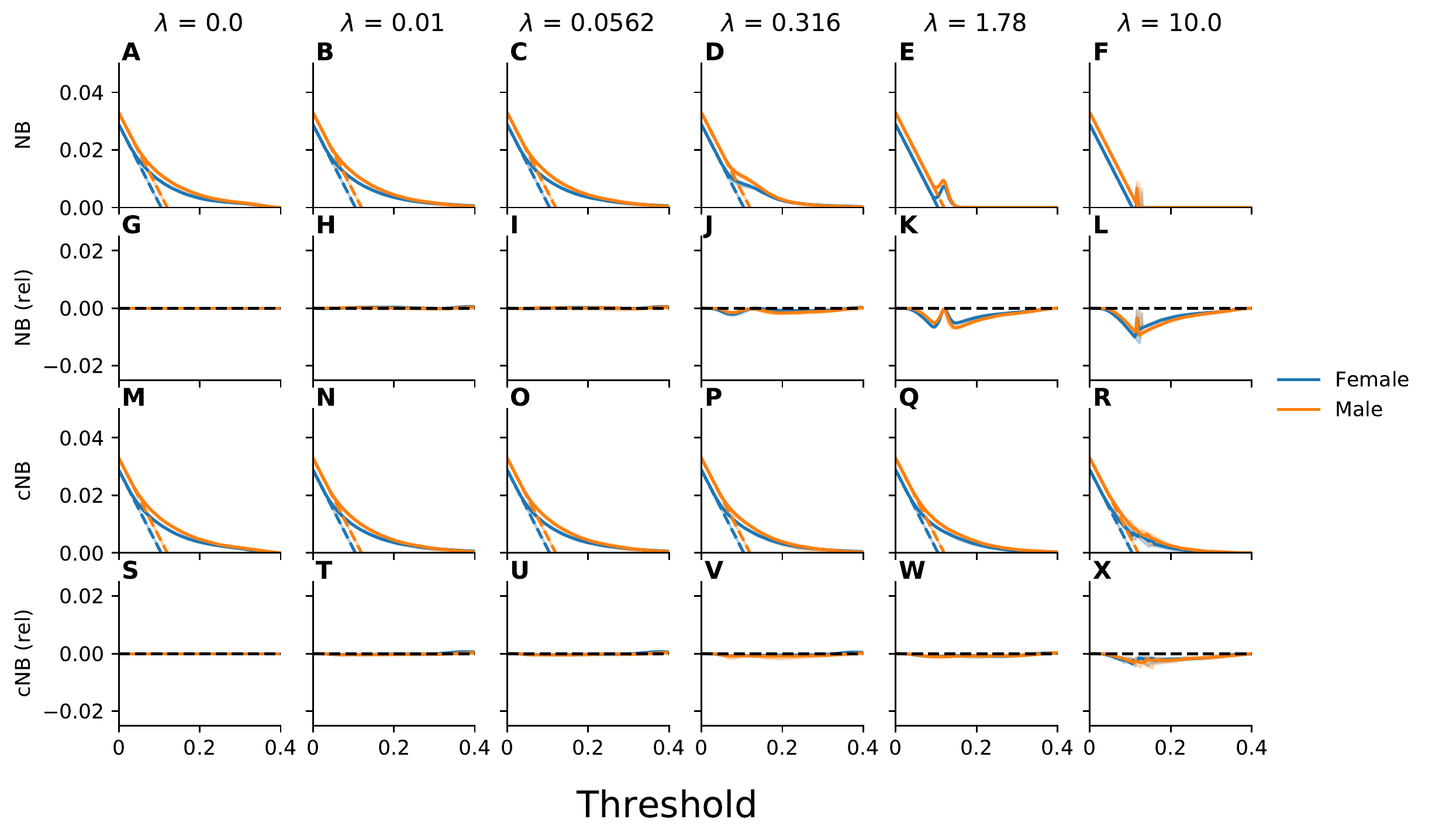}
	\caption{
	    Decision curve analysis to assess net benefit of models across subgroups defined by sex for models trained with an objective that penalizes violation of equalized odds across intersectional subgroups defined on the basis of race, ethnicity, and sex using a MMD-based penalty.
	    Plotted, for each subgroup and value of the regularization parameter $\lambda$, is the net benefit (NB) and calibrated net benefit (cNB) as a function of the decision threshold.
	    The net benefit of treating all patients is designated by dashed lines.
	    Results reported relative to the results for unconstrained empirical risk minimization are indicated by ``rel''.
	    Error bars indicate 95\% confidence intervals derived with the percentile bootstrap with 1,000 iterations.
	}
	\label{fig:supplement/eo_rr/sex/mmd/decision_curves}
\end{figure*}

%% Race threshold rate

\begin{figure*}[!t]
    \centering
    \includegraphics[width=0.95\linewidth]{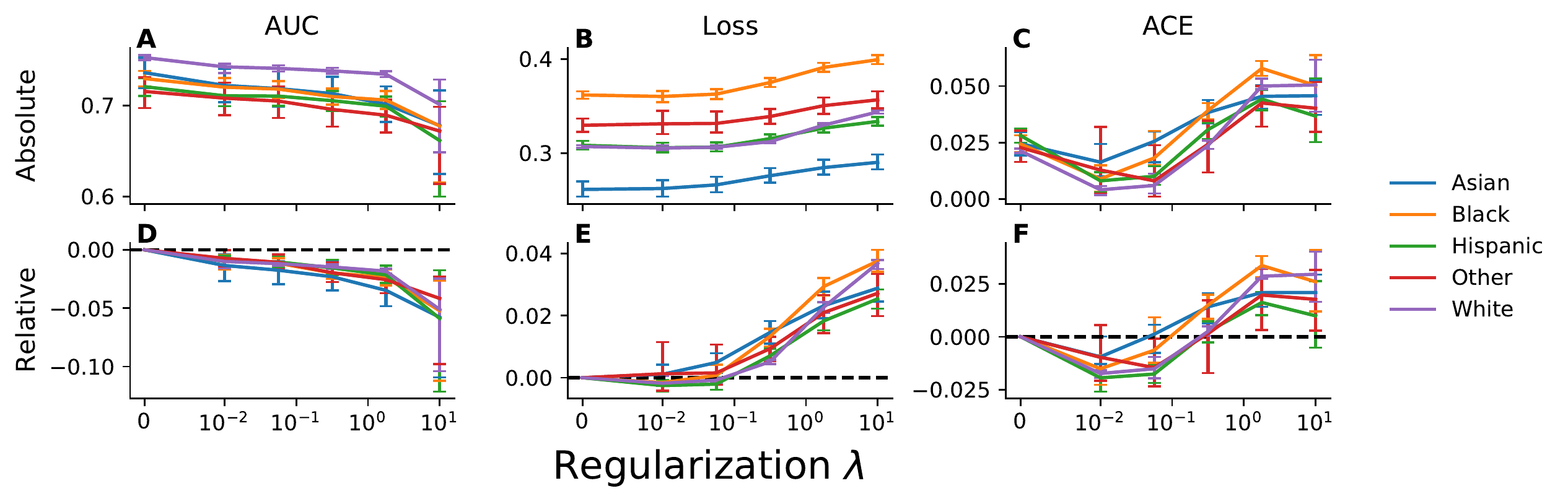}
    \caption{
        Model performance evaluated across racial and ethnic subgroups for models trained with an objective that penalizes violation of equalized odds across intersectional subgroups defined on the basis of race, ethnicity, and sex using a threshold-based penalty at 7.5\% and 20\%.
        Plotted, for each subgroup and value of the regularization parameter $\lambda$, is the area under the receiver operating characteristic curve (AUC), log-loss, and absolute calibration error (ACE).
        Relative results are reported relative to those attained for unconstrained empirical risk minimization. 
        Error bars indicate 95\% confidence intervals derived with the percentile bootstrap with 1,000 iterations.
    }
    \label{fig:supplement/eo_rr/race_eth/threshold_rate/eo_performance_lambda}
\end{figure*}

% Performance-EO grid
\begin{figure*}[!t]
    \centering
    \includegraphics[width=0.95\linewidth]{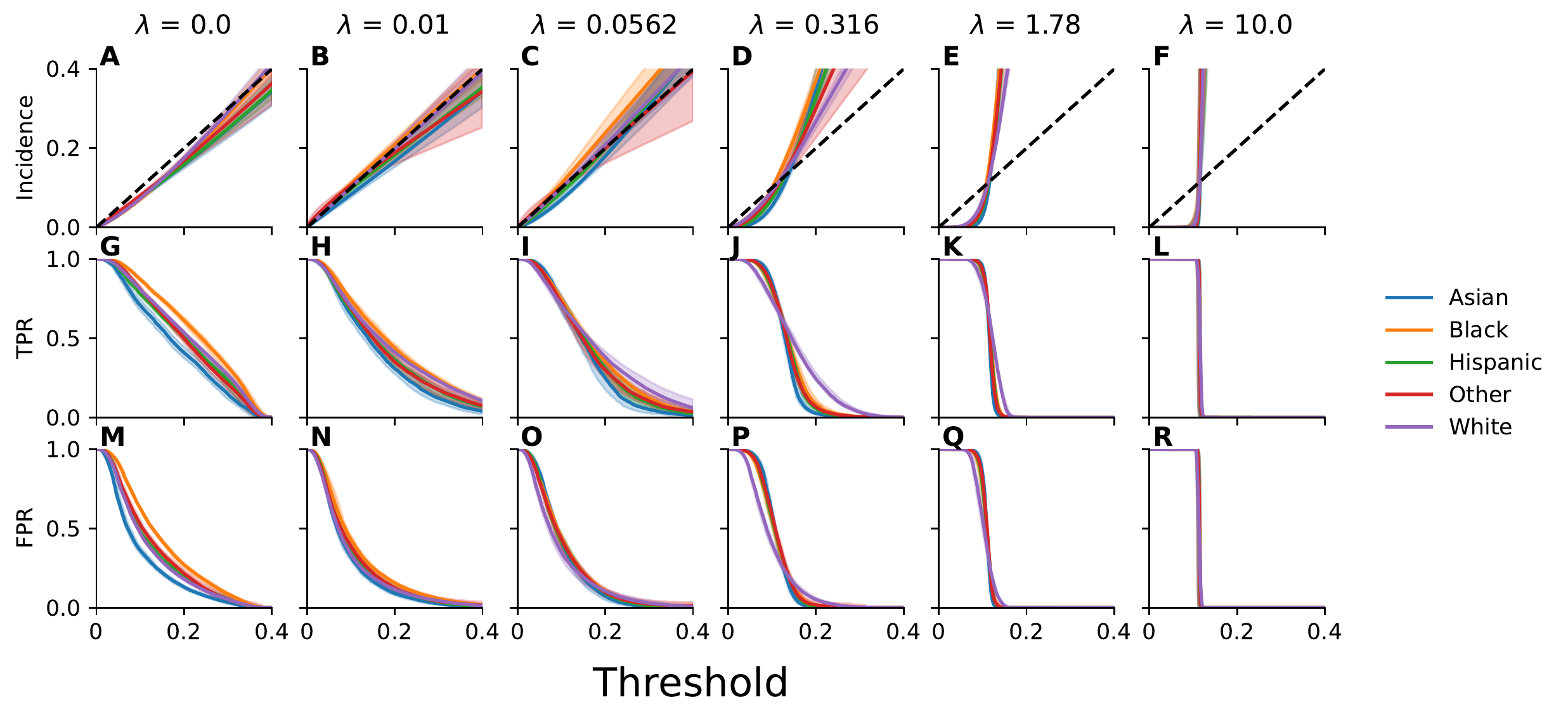}
    \caption{
        Calibration curves, true positive rates, and false positive rates evaluated for a range of thresholds across racial and ethnic subgroups for models trained with an objective that penalizes violation of equalized odds across intersectional subgroups defined on the basis of race, ethnicity, and sex using a threshold-based penalty at 7.5\% and 20\%.
        Plotted, for each subgroup and value of the regularization parameter $\lambda$, are the calibration curve (incidence), true positive rate (TPR), and false positive rate (FPR) as a function of the decision threshold.
        Error bars indicate 95\% confidence intervals derived with the percentile bootstrap with 1,000 iterations.
    }
    \label{fig:supplement/eo_rr/race_eth/threshold_rate/calibration_tpr_fpr}
\end{figure*}

% Net benefit metrics as a function of lambda
\begin{figure*}[!t]
    \centering
    \includegraphics[width=0.95\linewidth]{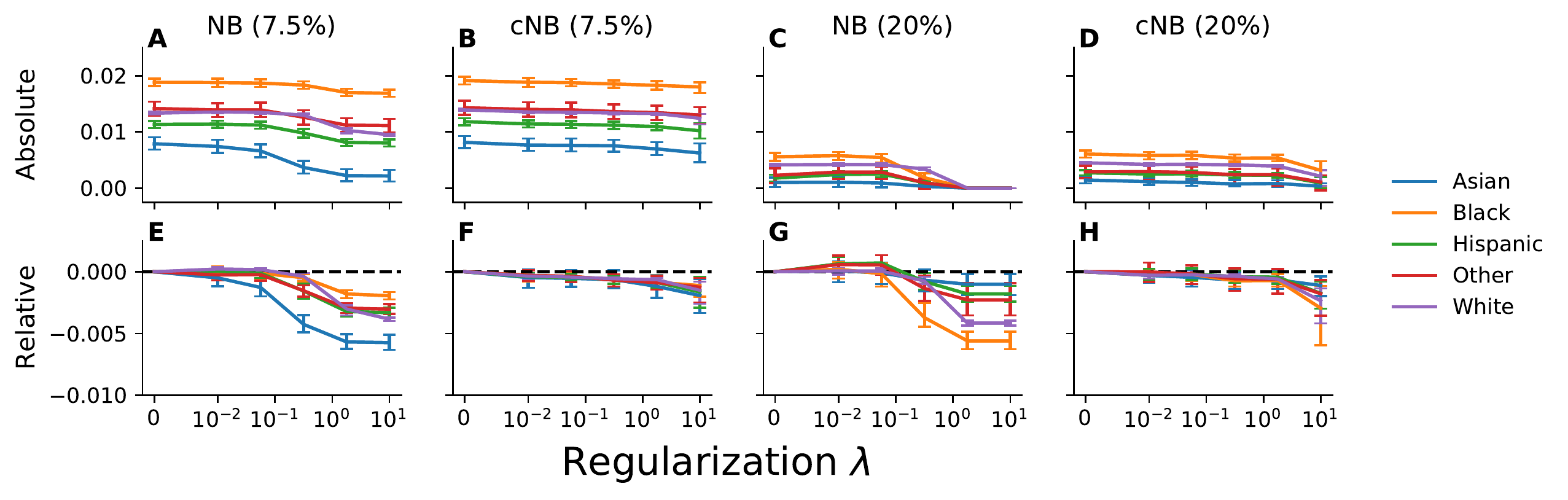}
    \caption{
        The net benefit evaluated across racial and ethnic subgroups, parameterized by the choice of a decision threshold of 7.5\% or 20\%, for models trained with an objective that penalizes violation of equalized odds across intersectional subgroups defined on the basis of race, ethnicity, and sex using a threshold-based penalty at 7.5\% and 20\%.
        Plotted, for each subgroup is the net benefit (NB) and calibrated net benefit (cNB) as a function of the value of the regularization parameter $\lambda$.
        Relative results are reported relative to those attained for unconstrained empirical risk minimization.
        Error bars indicate 95\% confidence intervals derived with the percentile bootstrap with 1,000 iterations.
    }
    \label{fig:supplement/eo_rr/race_eth/threshold_rate/eo_net_benefit_lambda}
\end{figure*}

% Equalized odds satisfaction
\begin{figure*}[!t]
    \centering
    \includegraphics[width=0.95\linewidth]{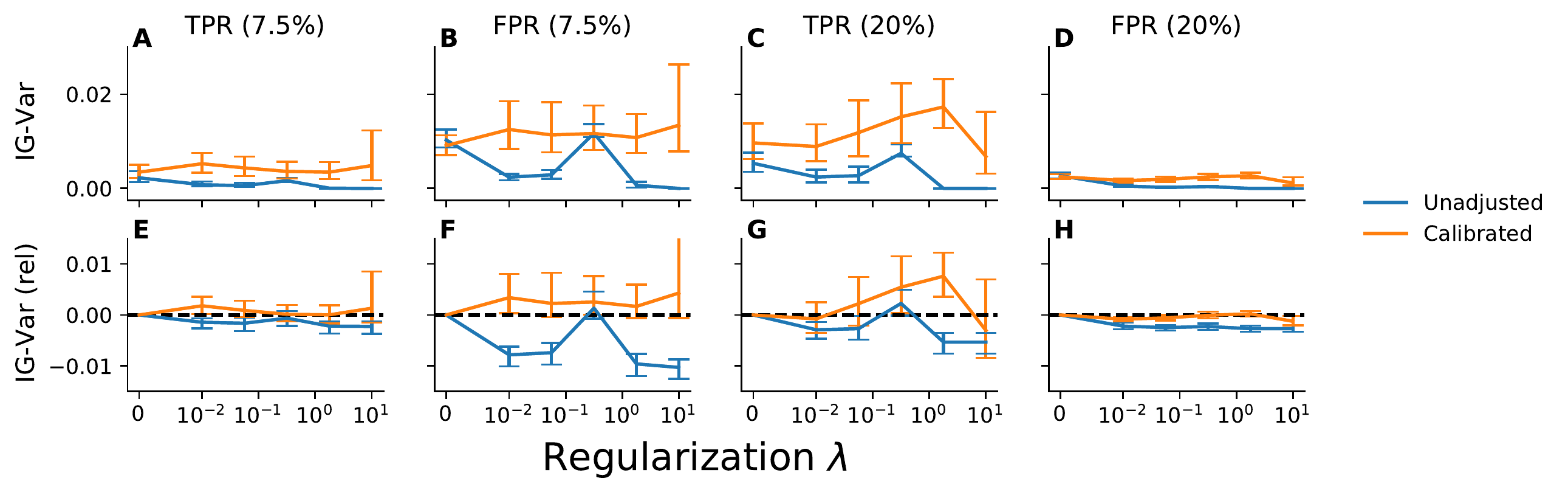}
    \caption{
        Satisfaction of equalized odds evaluated across racial and ethnic subgroups for models trained with an objective that penalizes violation of equalized odds across intersectional subgroups defined on the basis of race, ethnicity, and sex using a threshold-based penalty at 7.5\% and 20\%.
        Plotted is the intergroup variance (IG-Var) in the true positive and false positive rates at decision thresholds of 7.5\% and 20\%.
        Recalibrated results correspond to those attained for models for which the threshold has been adjusted to account for the observed miscalibration.
        Relative results are reported relative to those attained for unconstrained empirical risk minimization.
        Error bars indicate 95\% confidence intervals derived with the percentile bootstrap with 1,000 iterations.
    }
    \label{fig:supplement/eo_rr/race_eth/threshold_rate/eo_tpr_fpr_var_lambda}
\end{figure*}

% Standard decision curves

% Decision curves at t=0.075
\begin{figure*}[!ht]
    \centering
    \includegraphics[width=0.95\linewidth]{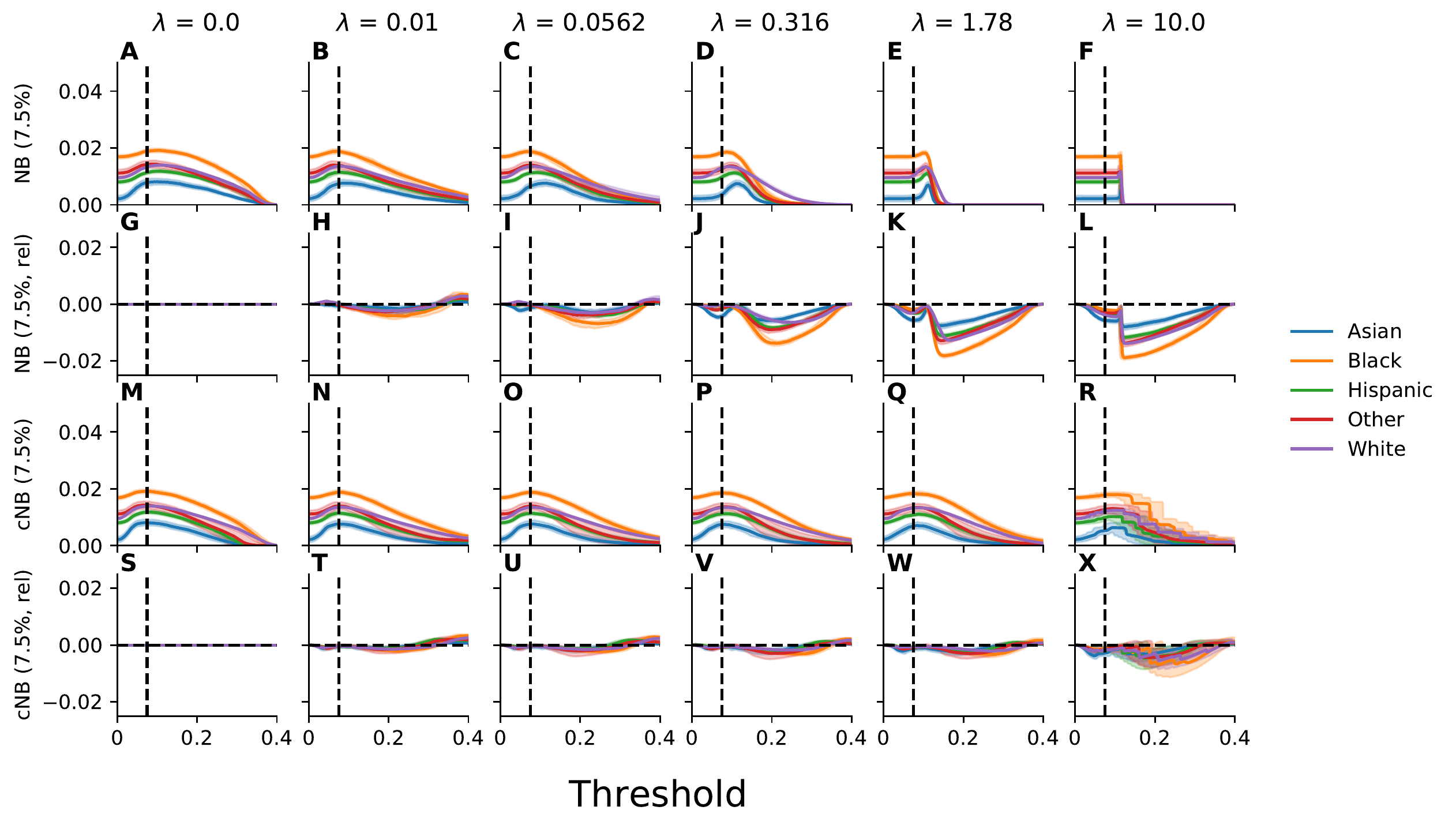}
    \caption{
        The net benefit evaluated for a range of thresholds across racial and ethnic subgroups, parameterized by the choice of a decision threshold of 7.5\%, for models trained with an objective that penalizes violation of equalized odds across intersectional subgroups defined on the basis of race, ethnicity, and sex using a threshold-based penalty at 7.5\% and 20\%.
        Plotted, for each subgroup and value of the regularization parameter $\lambda$, is the net benefit (NB) and calibrated net benefit (cNB) as a function of the decision threshold.
        Results reported relative to the results for unconstrained empirical risk minimization are indicated by ``rel''.
        Error bars indicate 95\% confidence intervals derived with the percentile bootstrap with 1,000 iterations.
    }
    \label{fig:supplement/eo_rr/race_eth/threshold_rate/decision_curves_075}
\end{figure*}

\begin{figure*}[!ht]
    
	\centering
	\includegraphics[width=0.95\linewidth]{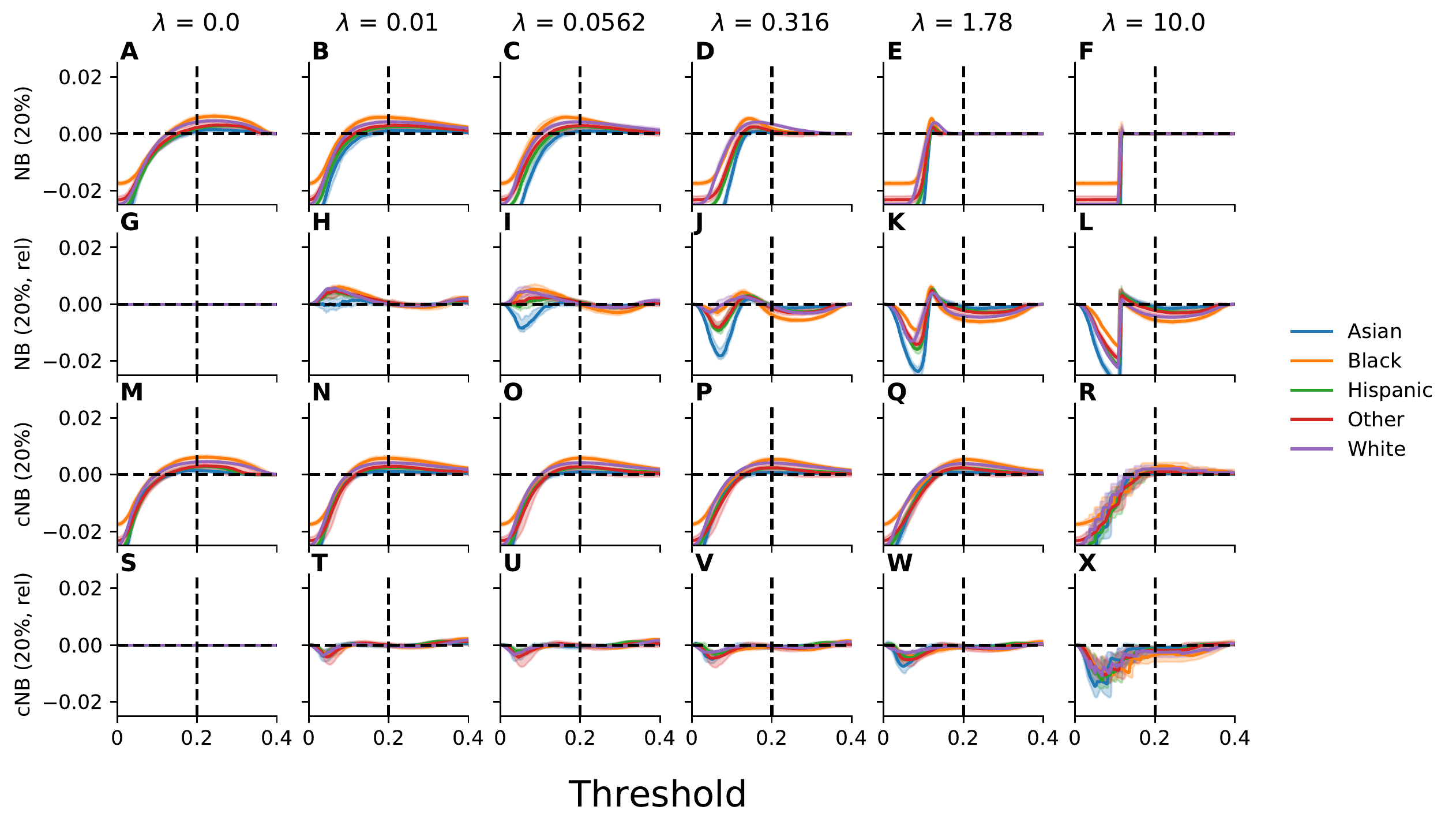}
	\caption{
	    The net benefit evaluated for a range of thresholds across racial and ethnic subgroups, parameterized by the choice of a decision threshold of 20\%, for models trained with an objective that penalizes violation of equalized odds across intersectional subgroups defined on the basis of race, ethnicity, and sex using a threshold-based penalty at 7.5\% and 20\%.
	    Plotted, for each subgroup and value of the regularization parameter $\lambda$, is the net benefit (NB) and calibrated net benefit (cNB) as a function of the decision threshold.
	    Results reported relative to the results for unconstrained empirical risk minimization are indicated by ``rel''.
	    Error bars indicate 95\% confidence intervals derived with the percentile bootstrap with 1,000 iterations.
	}
	\label{fig:supplement/eo_rr/race_eth/threshold_rate/decision_curves_20}
\end{figure*}

\begin{figure*}[!ht]
    
	\centering
	\includegraphics[width=0.95\linewidth]{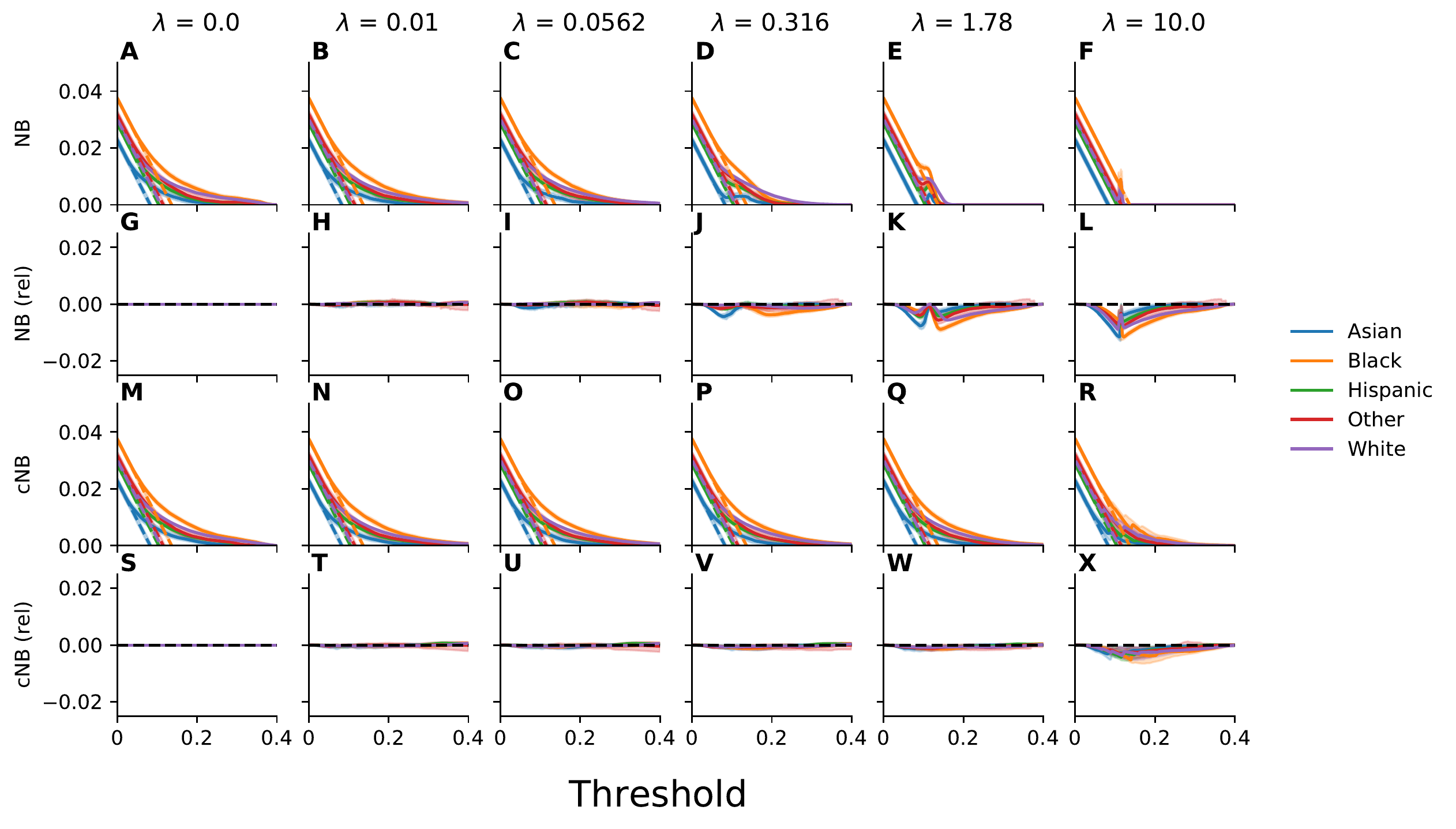}
	\caption{
	    Decision curve analysis to assess net benefit of models across racial and ethnic subgroups for models trained with an objective that penalizes violation of equalized odds across intersectional subgroups defined on the basis of race, ethnicity, and sex using a threshold-based penalty at 7.5\% and 20\%.
	    Plotted, for each subgroup and value of the regularization parameter $\lambda$, is the net benefit (NB) and calibrated net benefit (cNB) as a function of the decision threshold.
	    The net benefit of treating all patients is designated by dashed lines.
	    Results reported relative to the results for unconstrained empirical risk minimization are indicated by ``rel''.
	    Error bars indicate 95\% confidence intervals derived with the percentile bootstrap with 1,000 iterations.
	}
	\label{fig:supplement/eo_rr/race_eth/threshold_rate/decision_curves}
\end{figure*}

%% Intersectional threshold rate
\begin{figure*}[!ht]
	\centering
	\includegraphics[width=0.95\linewidth]{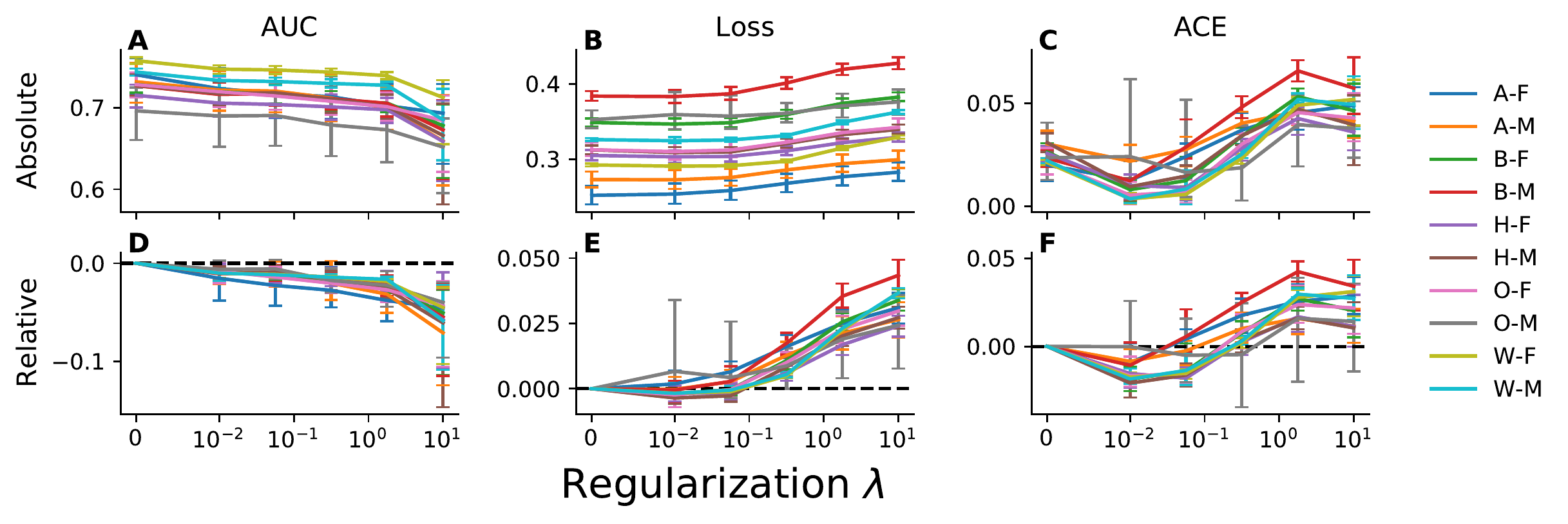}
	\caption{
	    The performance of models trained with an objective that penalizes violation of equalized odds across intersectional subgroups defined on the basis of race, ethnicity, and sex using a threshold-based penalty at 7.5\% and 20\%.
	    Plotted, for each subgroup and value of the regularization parameter $\lambda$, is the area under the receiver operating characteristic curve (AUC), log-loss, and absolute calibration error (ACE).
	    Relative results are reported relative to those attained for unconstrained empirical risk minimization. 
	    Labels correspond to Asian (A), Black (B), Hispanic (H), Other (O), White (W), Male (M), and Female (F) patients.
	    Error bars indicate 95\% confidence intervals derived with the percentile bootstrap with 1,000 iterations.
	}
	\label{fig:supplement/eo_rr/race_eth_sex/threshold_rate/eo_performance_lambda}
\end{figure*}

\begin{figure*}[!t]
    \centering
    \includegraphics[width=0.95\linewidth]{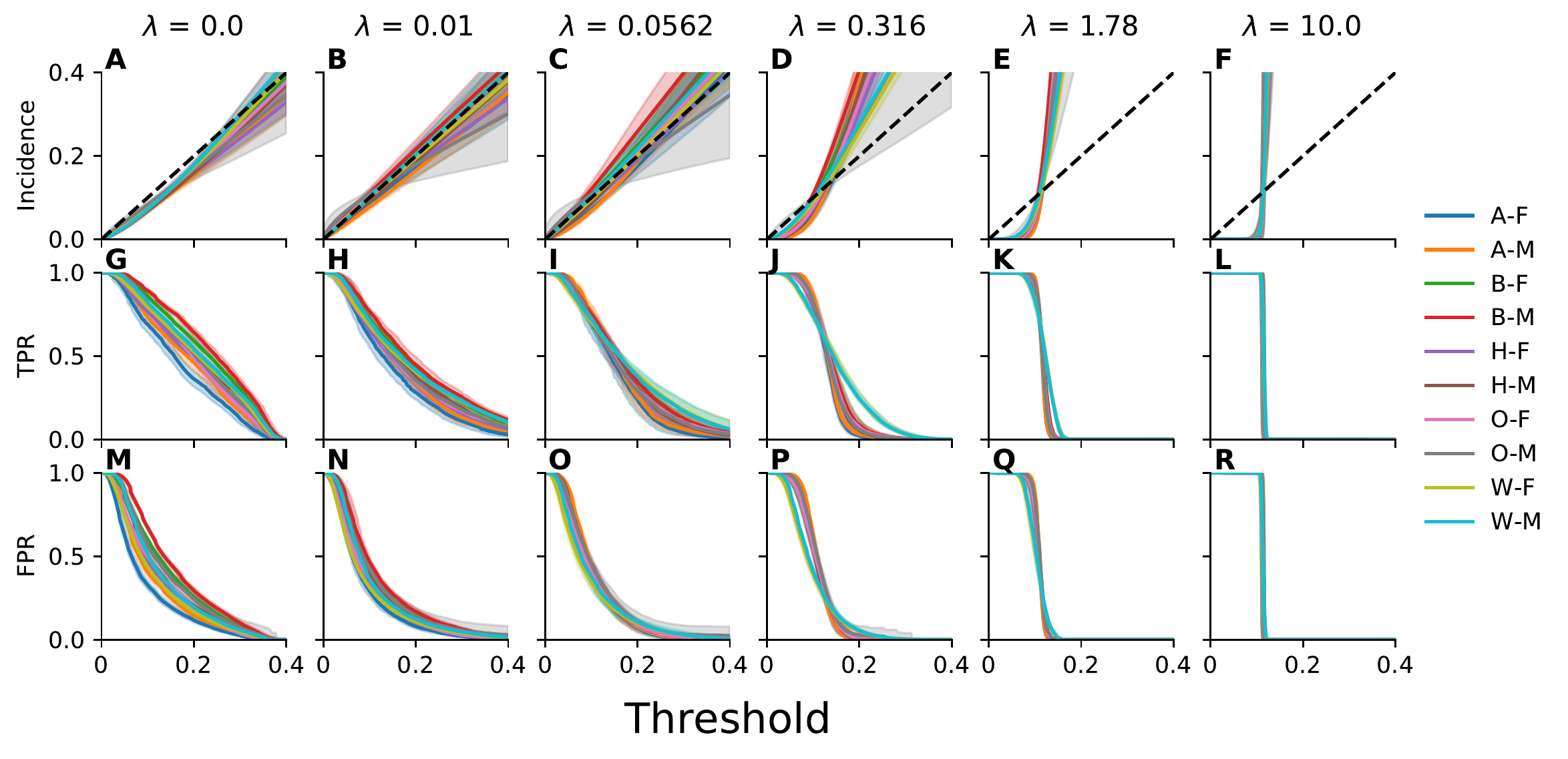}
    \caption{
        Calibration curves, true positive rates, and false positive rates evaluated for a range of thresholds for models trained with an objective that penalizes violation of equalized odds across intersectional subgroups defined on the basis of race, ethnicity, and sex using a threshold-based penalty at 7.5\% and 20\%.
        Plotted, for each subgroup and value of the regularization parameter $\lambda$, are the calibration curve (incidence), true positive rate (TPR), and false positive rate (FPR) as a function of the decision threshold.
        Error bars indicate 95\% confidence intervals derived with the percentile bootstrap with 1,000 iterations.
    }
    \label{fig:supplement/eo_rr/race_eth_sex/threshold_rate/calibration_tpr_fpr}
\end{figure*}

\begin{figure*}[!ht]
	\centering
	\includegraphics[width=0.95\linewidth]{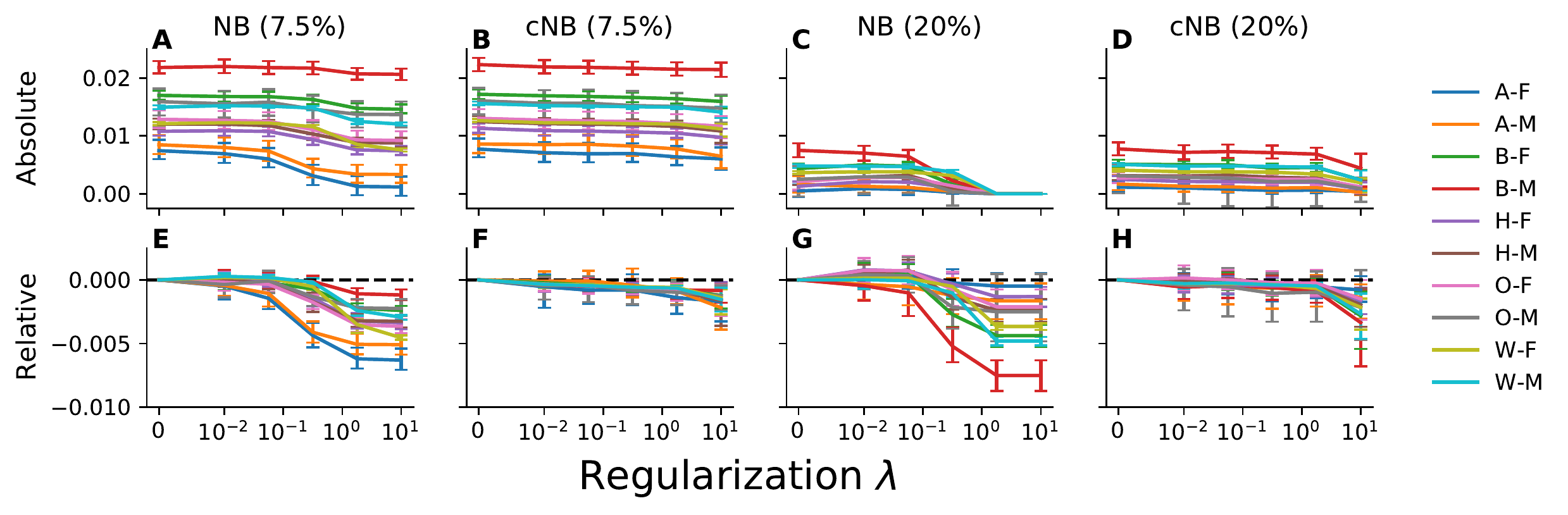}
	\caption{
	    The net benefit of models trained with an objective that penalizes violation of equalized odds across intersectional subgroups defined on the basis of race, ethnicity, and sex using a threshold-based penalty at 7.5\% and 20\%, parameterized by the choice of a decision threshold of 7.5\% or 20\%.
	    Plotted, for each subgroup is the net benefit (NB) and calibrated net benefit (rNB) as a function of the value of the regularization parameter $\lambda$, .
	    Relative results are reported relative to those attained for unconstrained empirical risk minimization.
	    Labels correspond to Asian (A), Black (B), Hispanic (H), Other (O), White (W), Male (M), and Female (F) patients.
	    Error bars indicate 95\% confidence intervals derived with the percentile bootstrap with 1,000 iterations.
	}
	\label{fig:supplement/eo_rr/race_eth_sex/threshold_rate/eo_net_benefit_lambda}
\end{figure*}

\begin{figure*}[!ht]
	\centering
	\includegraphics[width=0.95\linewidth]{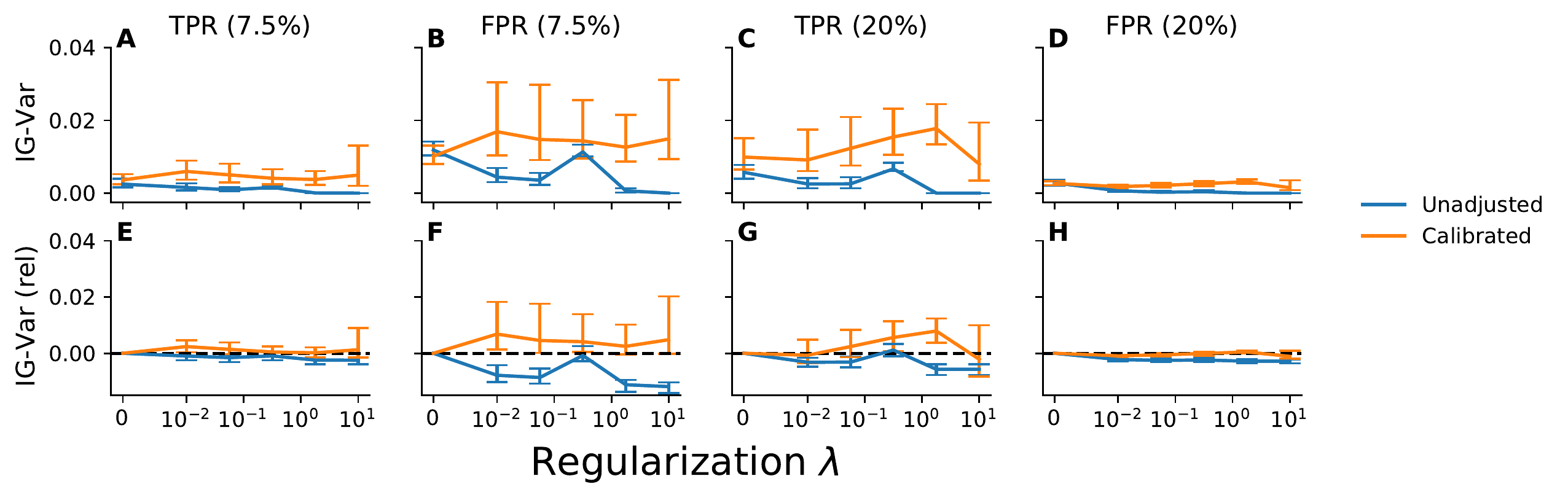}
	\caption{
        Satisfaction of equalized odds for models trained with an objective that penalizes violation of equalized odds across intersectional subgroups defined on the basis of race, ethnicity, and sex using a threshold-based penalty at 7.5\% and 20\%.
        Plotted is the intergroup variance (IG-Var) in the true positive and false positive rates at decision thresholds of 7.5\% and 20\%.
        Recalibrated results correspond to those attained for models for which the threshold has been adjusted to account for the observed miscalibration.
        Relative results are reported relative to those attained for unconstrained empirical risk minimization.
        Labels correspond to Asian (A), Black (B), Hispanic (H), Other (O), White (W), Male (M), and Female (F) patients.
	    Error bars indicate 95\% confidence intervals derived with the percentile bootstrap with 1,000 iterations.
	}
	\label{fig:supplement/eo_rr/race_eth_sex/threshold_rate/eo_tpr_fpr_var_lambda}
\end{figure*}

\begin{figure*}[!ht]
	\centering
	\includegraphics[width=0.95\linewidth]{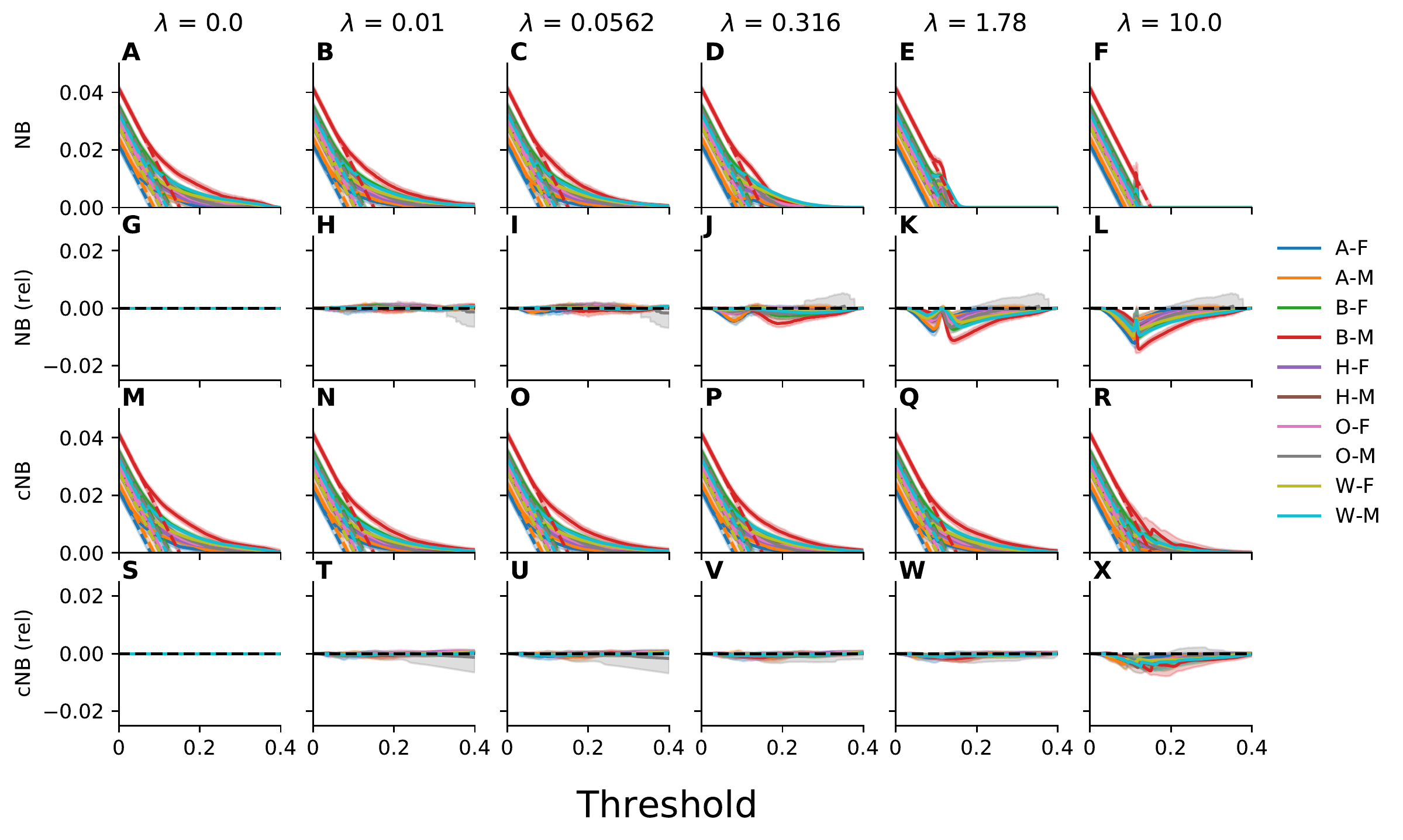}
	\caption{
	    Decision curve analysis to assess net benefit of models trained with an objective that penalizes violation of equalized odds across intersectional subgroups defined on the basis of race, ethnicity, and sex using a threshold-based penalty at 7.5\% and 20\%.
	    Plotted, for each subgroup and value of the regularization parameter $\lambda$, is the net benefit (NB) and calibrated net benefit (rNB) as a function of the decision threshold.
	    The net benefit of treating all patients is designated by dashed lines.
	    Results reported relative to the results for unconstrained empirical risk minimization are indicated by ``rel''.
	    Labels correspond to Asian (A), Black (B), Hispanic (H), Other (O), White (W), Male (M), and Female (F) patients.
	    Error bars indicate 95\% confidence intervals derived with the percentile bootstrap with 1,000 iterations.
	}
	\label{fig:supplement/eo_rr/race_eth_sex/threshold_rate/decision_curves}
\end{figure*}

\begin{figure*}[!ht]
	\centering
	\includegraphics[width=0.95\linewidth]{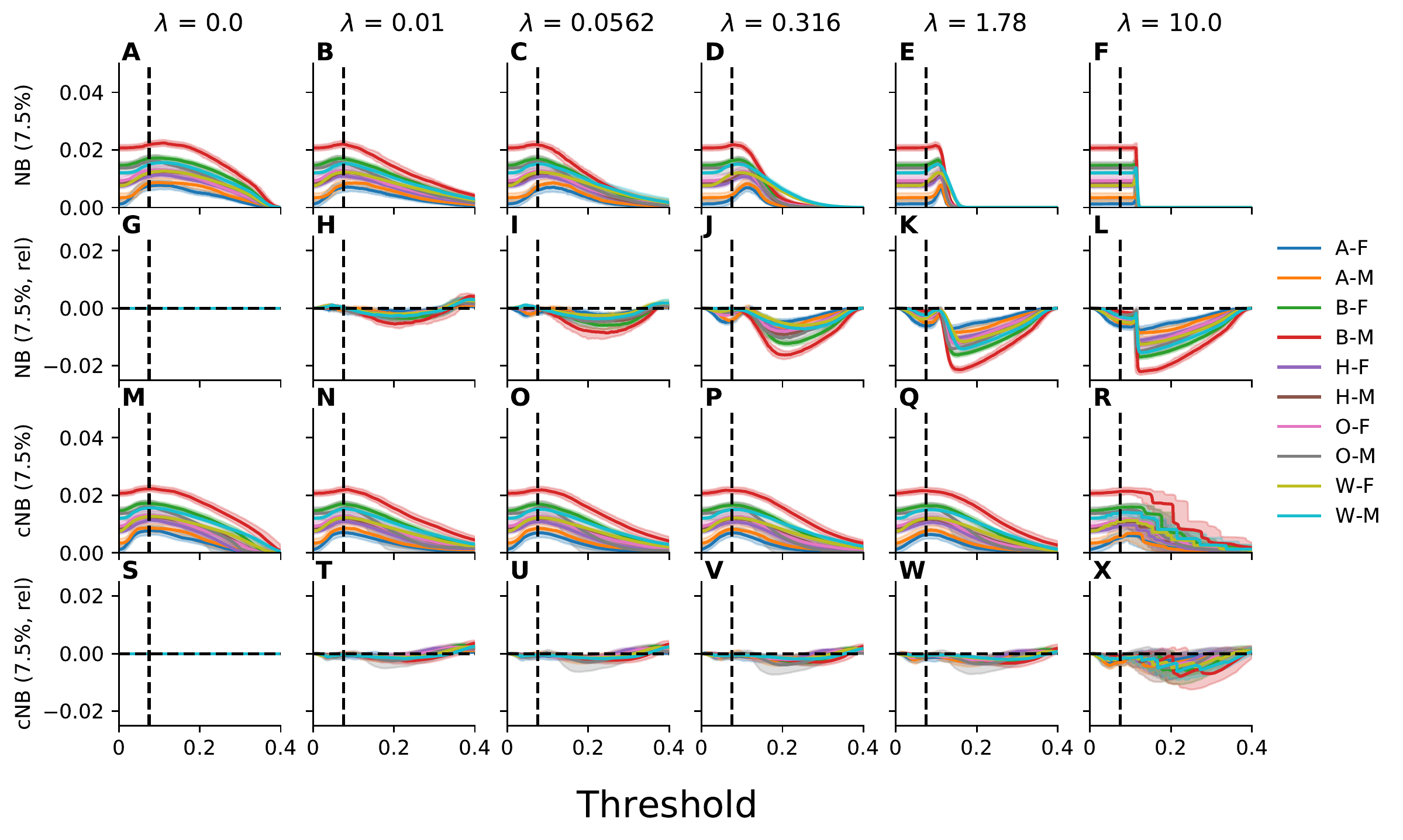}
	\caption{
	    The net benefit of models trained with an objective that penalizes violation of equalized odds across intersectional subgroups defined on the basis of race, ethnicity, and sex using a threshold-based penalty at 7.5\% and 20\%, parameterized by the choice of a decision threshold of 7.5\%.
	    Plotted, for each subgroup and value of the regularization parameter $\lambda$, is the net benefit (NB) and calibrated net benefit (rNB) as a function of the decision threshold.
	    The net benefit of treating all patients is designated by dashed lines.
	    Results reported relative to the results for unconstrained empirical risk minimization are indicated by ``rel''.
	    Labels correspond to Asian (A), Black (B), Hispanic (H), Other (O), White (W), Male (M), and Female (F) patients.
	    Error bars indicate 95\% confidence intervals derived with the percentile bootstrap with 1,000 iterations.
	}
	\label{fig:supplement/eo_rr/race_eth_sex/threshold_rate/decision_curves_075}
\end{figure*}

\begin{figure*}[!ht]
	\centering
	\includegraphics[width=0.95\linewidth]{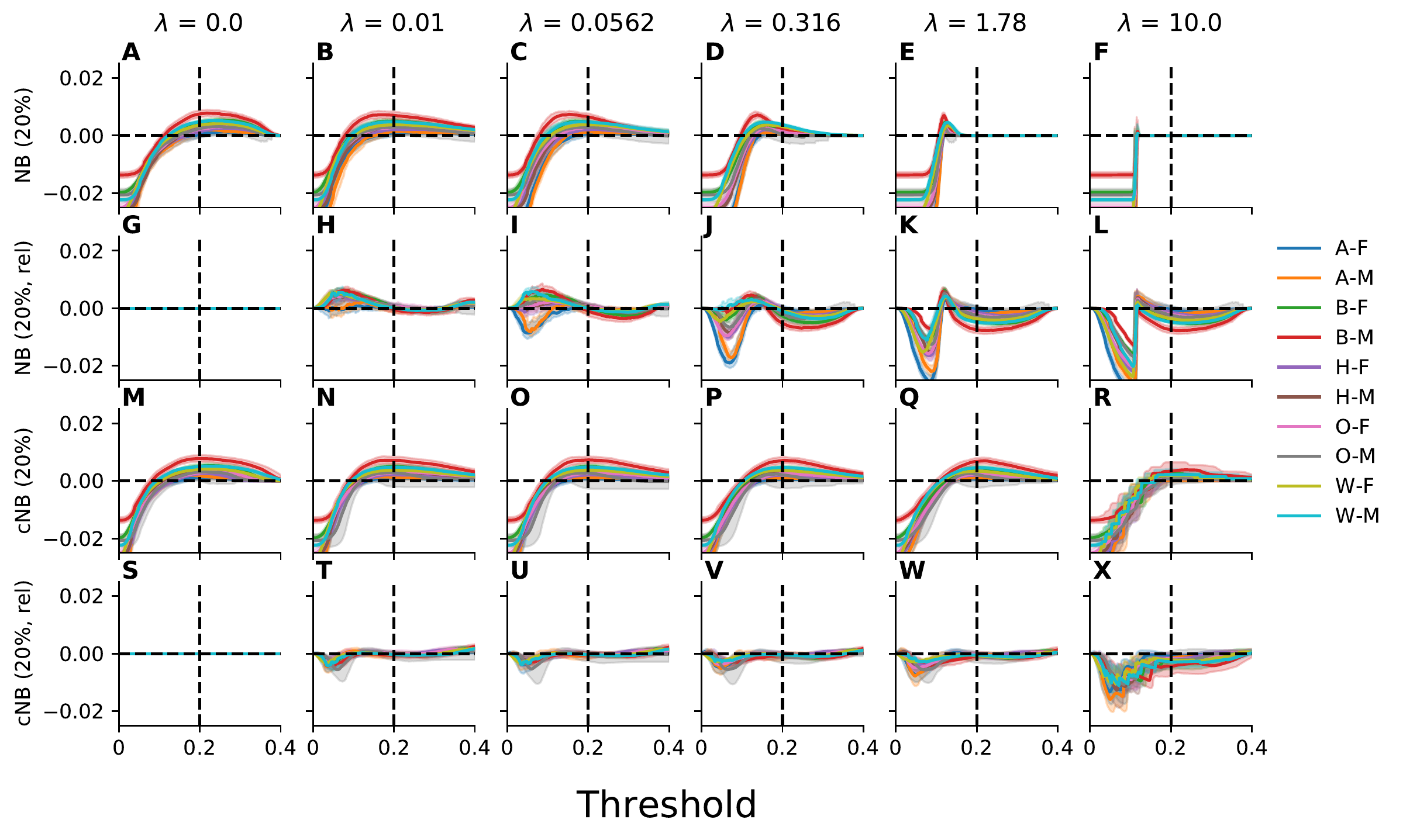}
	\caption{
	    The net benefit of models trained with an objective that penalizes violation of equalized odds across intersectional subgroups defined on the basis of race, ethnicity, and sex using a threshold-based penalty at 7.5\% and 20\%, parameterized by the choice of a decision threshold of 20\%.
	    Plotted, for each subgroup and value of the regularization parameter $\lambda$, is the net benefit (NB) and calibrated net benefit (rNB) as a function of the decision threshold.
	    The net benefit of treating all patients is designated by dashed lines.
	    Results reported relative to the results for unconstrained empirical risk minimization are indicated by ``rel''.
	    Labels correspond to Asian (A), Black (B), Hispanic (H), Other (O), White (W), Male (M), and Female (F) patients.
	    Error bars indicate 95\% confidence intervals derived with the percentile bootstrap with 1,000 iterations.
	}
	\label{fig:supplement/eo_rr/race_eth_sex/threshold_rate/decision_curves_20}
\end{figure*}

%% Sex - threshold-rate
\begin{figure*}[!t]
    \centering
    \includegraphics[width=0.95\linewidth]{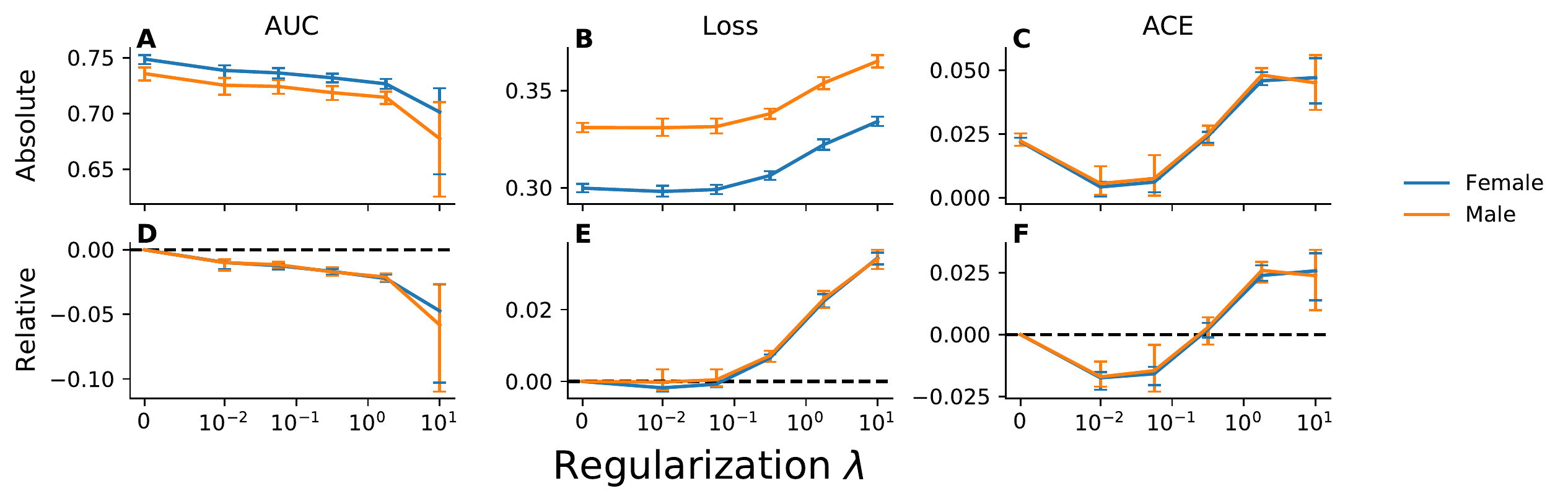}
    \caption{
        Model performance evaluated across subgroups defined by sex for models trained with an objective that penalizes violation of equalized odds across intersectional subgroups defined on the basis of race, ethnicity, and sex using a threshold-based penalty at 7.5\% and 20\%.
        Plotted, for each subgroup and value of the regularization parameter $\lambda$, is the area under the receiver operating characteristic curve (AUC), log-loss, and absolute calibration error (ACE).
        Relative results are reported relative to those attained for unconstrained empirical risk minimization. 
        Error bars indicate 95\% confidence intervals derived with the percentile bootstrap with 1,000 iterations.
    }
    \label{fig:supplement/eo_rr/sex/threshold_rate/eo_performance_lambda}
\end{figure*}

% Performance-EO grid
\begin{figure*}[!t]
    \centering
    \includegraphics[width=0.95\linewidth]{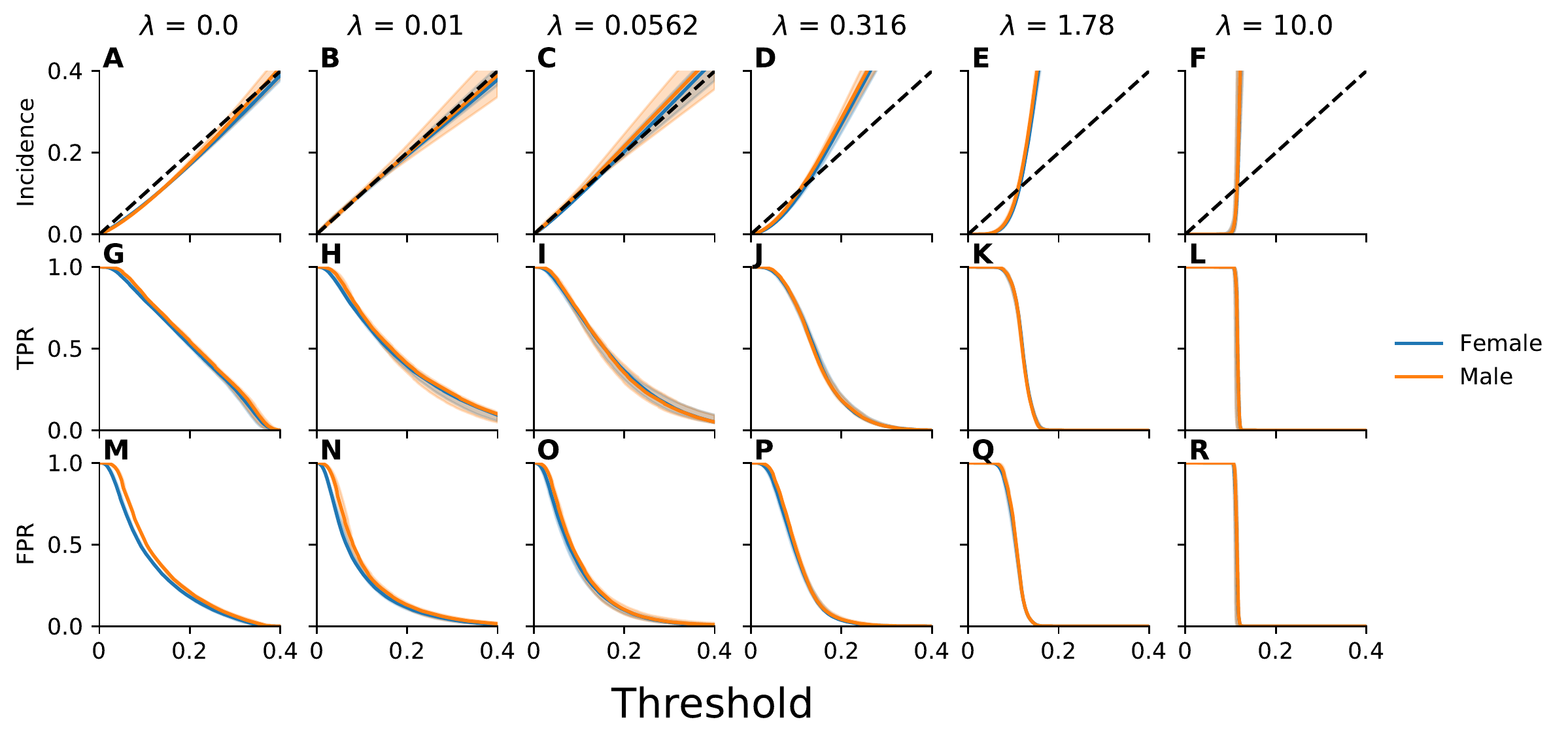}
    \caption{
        Calibration curves, true positive rates, and false positive rates evaluated for a range of thresholds across subgroups defined by sex for models trained with an objective that penalizes violation of equalized odds across intersectional subgroups defined on the basis of race, ethnicity, and sex using a threshold-based penalty at 7.5\% and 20\%.
        Plotted, for each subgroup and value of the regularization parameter $\lambda$, are the calibration curve (incidence), true positive rate (TPR), and false positive rate (FPR) as a function of the decision threshold.
        Error bars indicate 95\% confidence intervals derived with the percentile bootstrap with 1,000 iterations.
    }
    \label{fig:supplement/eo_rr/sex/threshold_rate/calibration_tpr_fpr}
\end{figure*}

% Net benefit metrics as a function of lambda
\begin{figure*}[!t]
    \centering
    \includegraphics[width=0.95\linewidth]{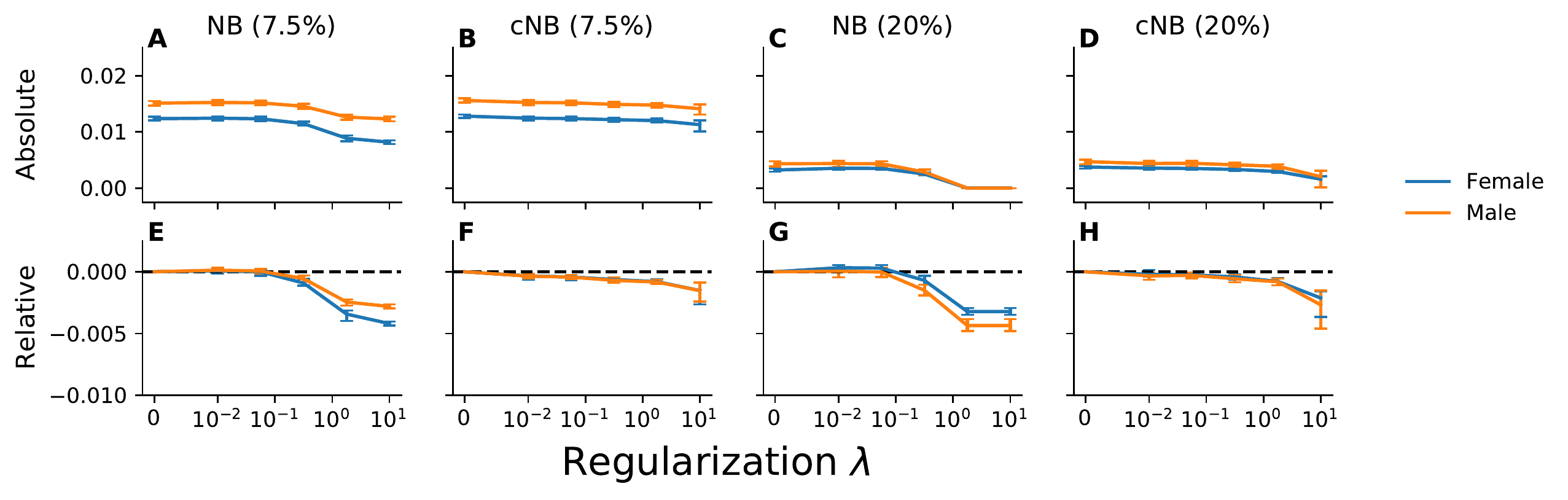}
    \caption{
        The net benefit evaluated across subgroups defined by sex, parameterized by the choice of a decision threshold of 7.5\% or 20\%, for models trained with an objective that penalizes violation of equalized odds across intersectional subgroups defined on the basis of race, ethnicity, and sex using a threshold-based penalty at 7.5\% and 20\%.
        Plotted, for each subgroup is the net benefit (NB) and calibrated net benefit (cNB) as a function of the value of the regularization parameter $\lambda$.
        Relative results are reported relative to those attained for unconstrained empirical risk minimization.
        Error bars indicate 95\% confidence intervals derived with the percentile bootstrap with 1,000 iterations.
    }
    \label{fig:supplement/eo_rr/sex/threshold_rate/eo_net_benefit_lambda}
\end{figure*}

% Equalized odds satisfaction
\begin{figure*}[!t]
    \centering
    \includegraphics[width=0.95\linewidth]{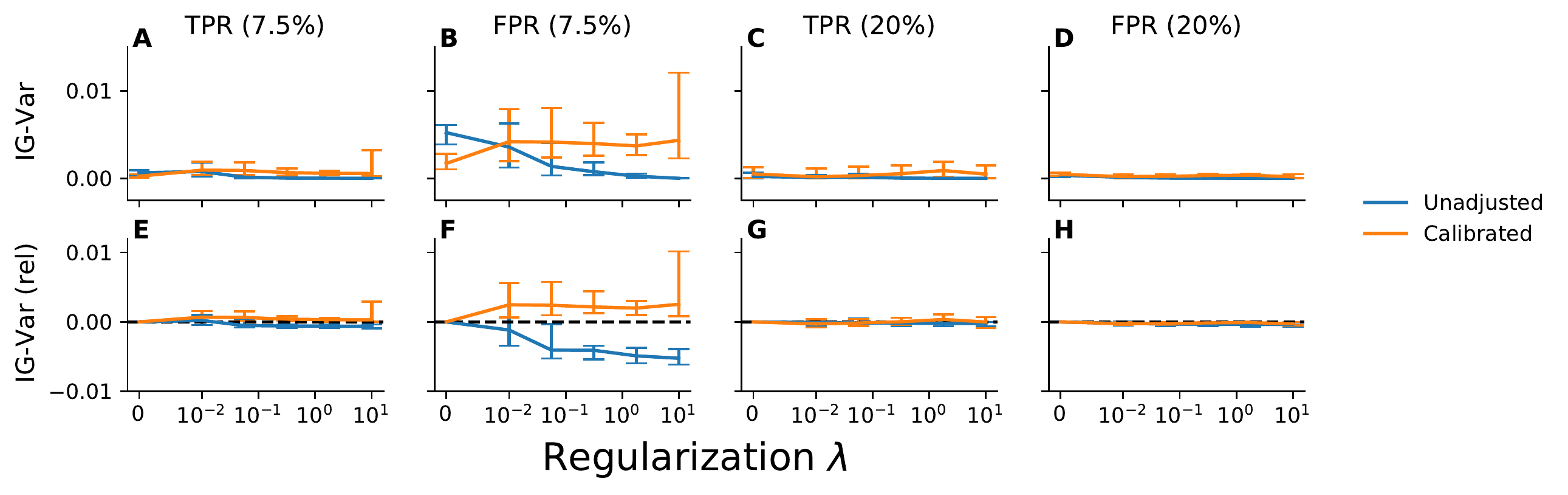}
    \caption{
        Satisfaction of equalized odds evaluated across subgroups defined by sex for models trained with an objective that penalizes violation of equalized odds across intersectional subgroups defined on the basis of race, ethnicity, and sex using a threshold-based penalty at 7.5\% and 20\%.
        Plotted is the intergroup variance (IG-Var) in the true positive and false positive rates at decision thresholds of 7.5\% and 20\%.
        Recalibrated results correspond to those attained for models for which the threshold has been adjusted to account for the observed miscalibration.
        Relative results are reported relative to those attained for unconstrained empirical risk minimization.
        Error bars indicate 95\% confidence intervals derived with the percentile bootstrap with 1,000 iterations.
    }
    \label{fig:supplement/eo_rr/sex/threshold_rate/eo_tpr_fpr_var_lambda}
\end{figure*}

% Decision curves at t=0.075
\begin{figure*}[!ht]
    \centering
    \includegraphics[width=0.95\linewidth]{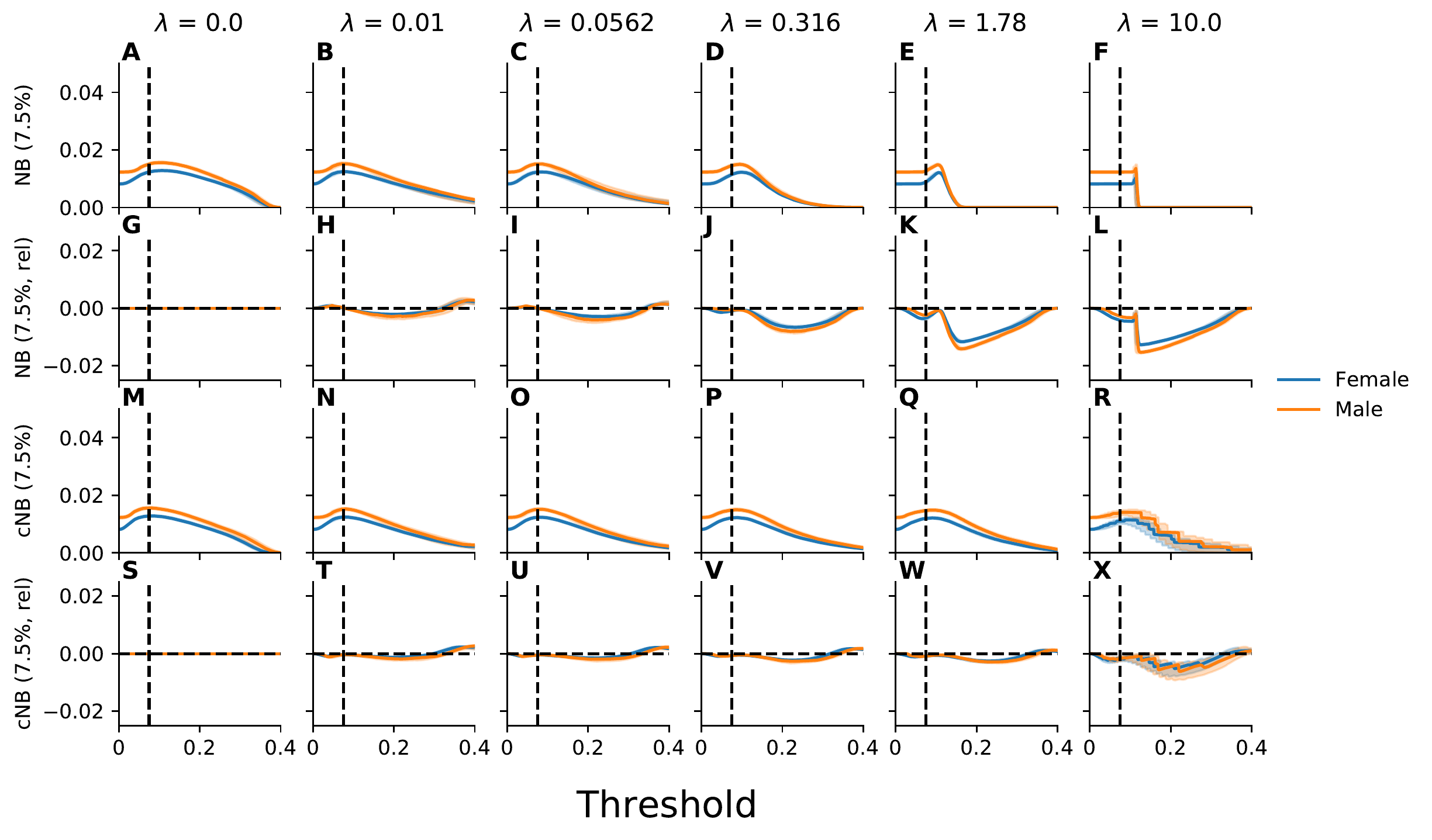}
    \caption{
        The net benefit evaluated for a range of thresholds across subgroups defined by sex, parameterized by the choice of a decision threshold of 7.5\%, for models trained with an objective that penalizes violation of equalized odds across intersectional subgroups defined on the basis of race, ethnicity, and sex using a threshold-based penalty at 7.5\% and 20\%.
        Plotted, for each subgroup and value of the regularization parameter $\lambda$, is the net benefit (NB) and calibrated net benefit (cNB) as a function of the decision threshold.
        Results reported relative to the results for unconstrained empirical risk minimization are indicated by ``rel''.
        Error bars indicate 95\% confidence intervals derived with the percentile bootstrap with 1,000 iterations.
    }
    \label{fig:supplement/eo_rr/sex/threshold_rate/decision_curves_075}
\end{figure*}

\begin{figure*}[!ht]
    
	\centering
	\includegraphics[width=0.95\linewidth]{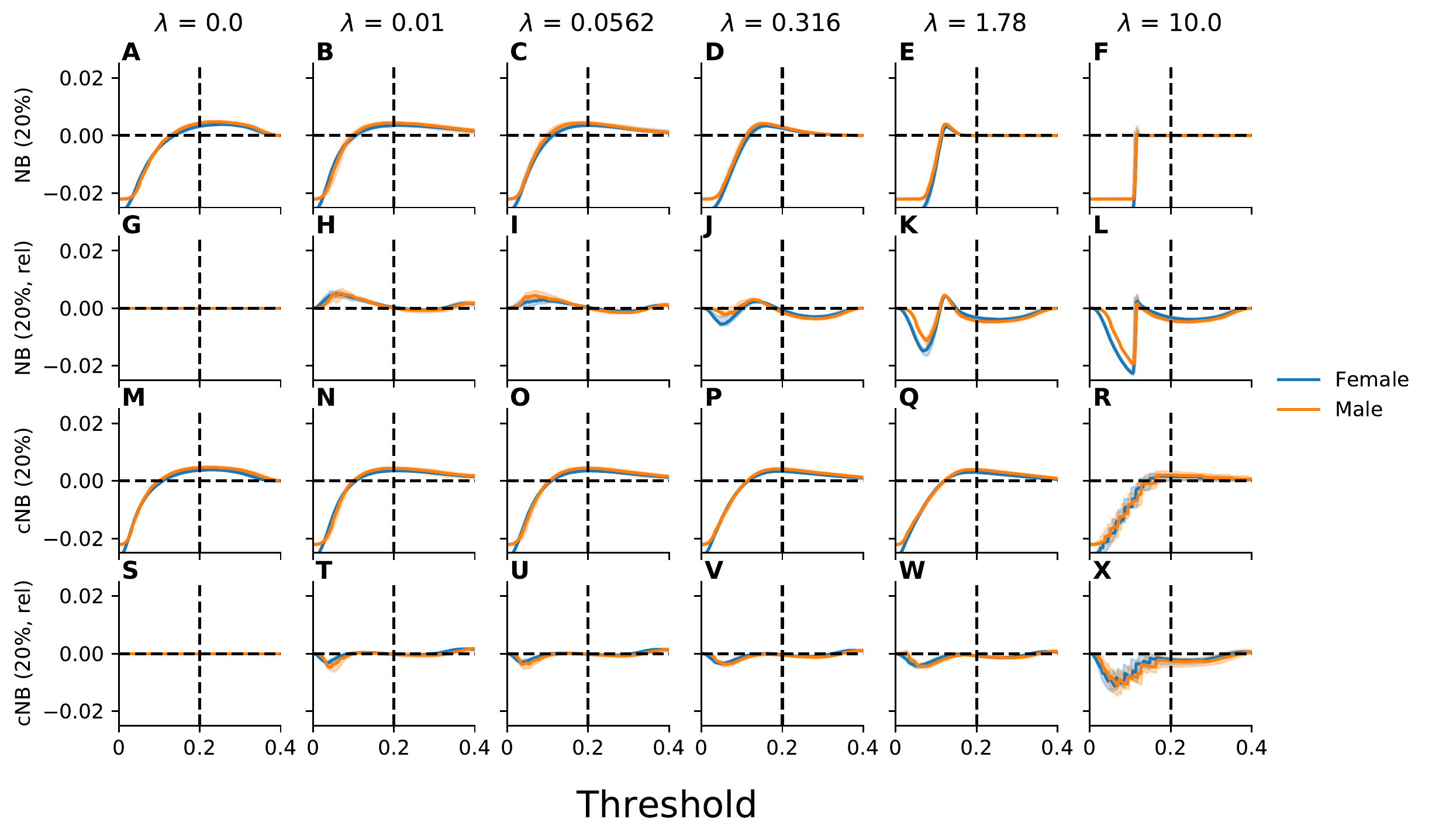}
	\caption{
	    The net benefit evaluated for a range of thresholds across subgroups defined by sex, parameterized by the choice of a decision threshold of 20\%, for models trained with an objective that penalizes violation of equalized odds across intersectional subgroups defined on the basis of race, ethnicity, and sex using a threshold-based penalty at 7.5\% and 20\%.
	    Plotted, for each subgroup and value of the regularization parameter $\lambda$, is the net benefit (NB) and calibrated net benefit (cNB) as a function of the decision threshold.
	    Results reported relative to the results for unconstrained empirical risk minimization are indicated by ``rel''.
	    Error bars indicate 95\% confidence intervals derived with the percentile bootstrap with 1,000 iterations.
	}
	\label{fig:supplement/eo_rr/sex/threshold_rate/decision_curves_20}
\end{figure*}

\begin{figure*}[!ht]
    
	\centering
	\includegraphics[width=0.95\linewidth]{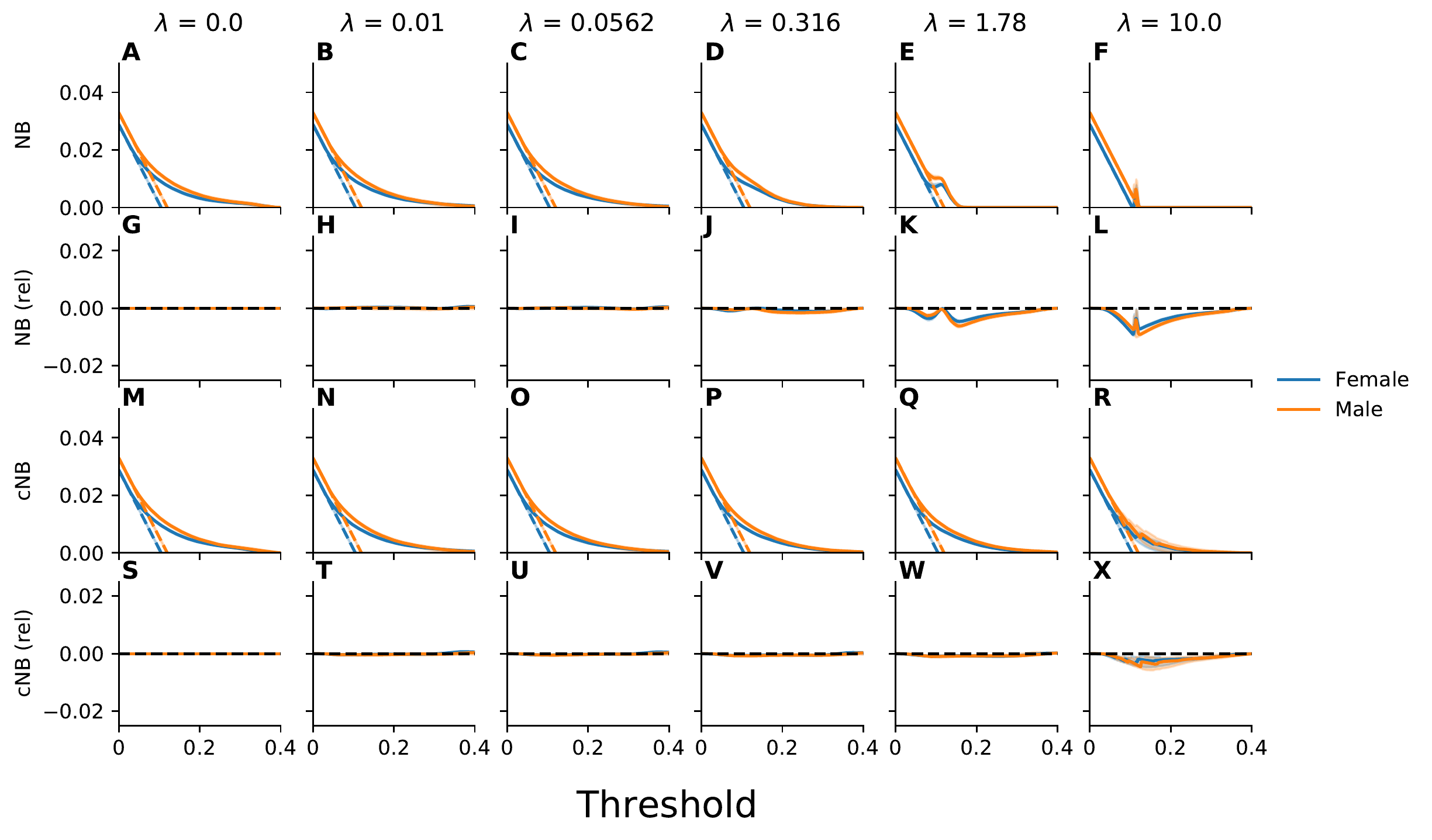}
	\caption{
	    Decision curve analysis to assess net benefit of models across subgroups defined by sex for models trained with an objective that penalizes violation of equalized odds across intersectional subgroups defined on the basis of race, ethnicity, and sex using a threshold-based penalty at 7.5\% and 20\%.
	    Plotted, for each subgroup and value of the regularization parameter $\lambda$, is the net benefit (NB) and calibrated net benefit (cNB) as a function of the decision threshold.
	    The net benefit of treating all patients is designated by dashed lines.
	    Results reported relative to the results for unconstrained empirical risk minimization are indicated by ``rel''.
	    Error bars indicate 95\% confidence intervals derived with the percentile bootstrap with 1,000 iterations.
	}
	\label{fig:supplement/eo_rr/sex/threshold_rate/decision_curves}
\end{figure*}

\end{document}